\documentclass{article}
\pdfoutput=1

\usepackage{fullpage}
\usepackage{natbib}
\usepackage{microtype}
\usepackage{graphicx}
\usepackage{listings}
\usepackage{enumitem}
\usepackage{mwe}
\usepackage{subfigure}
\usepackage{booktabs} % for professional tables
\usepackage{authblk}
\usepackage{mathtools}
\usepackage{amsfonts}
\usepackage[ruled,boxed]{algorithm2e}

\usepackage{amsmath,amsthm,amssymb}

\usepackage{graphicx}
\usepackage{subfigure}
\usepackage{tabularx}
\usepackage{colortbl}
\usepackage{xcolor}
\usepackage{enumerate}
\usepackage{caption}
\usepackage{multirow}
\usepackage{booktabs}
\usepackage{threeparttable}
\usepackage{dsfont}

\newcommand{\bR}{\mathbb{R}}

\newcommand{\cF}{\mathcal{F}}
\newcommand{\cX}{\mathcal{X}}
\newcommand{\cY}{\mathcal{Y}}
\newcommand{\bx}{{x}}
\newcommand{\bxprime}{{x}'}
\newcommand{\bxtidle}{\tilde{{x}}}
\newcommand{\epsball}{\mathcal{B}_\epsilon}
\newcommand{\xadv}{\tilde{{x}}}

\DeclareMathOperator{\sign}{sign}

\usepackage{hyperref}

\title{Understanding the Interaction of Adversarial Training\\ with Noisy Labels}

\author[1]{Jianing Zhu\thanks{Equal contributions.}}
\author[2]{Jingfeng Zhang$^*$}
\author[1,2]{Bo Han\thanks{Corresponding authors.}}
\author[3]{Tongliang Liu$^{\dagger}$}
\author[2]{Gang Niu$^{\dagger}$}
\author[4]{Hongxia Yang}
\author[5]{Mohan Kankanhalli}
\author[2,6]{Masashi Sugiyama}
\affil[1]{Hong Kong Baptist University}
\affil[2]{RIKEN Center for Advanced Intelligence Project}
\affil[3]{University of Sydney}
\affil[4]{Alibaba Group}
\affil[5]{National University of Singapore}
\affil[6]{University of Tokyo}
\date{}

\begin{document}

\maketitle

\begin{abstract}
\emph{Noisy labels}~(NL) and \emph{adversarial examples} both undermine trained models, but interestingly they have hitherto been studied \emph{independently}.
A recent \emph{adversarial training}~(AT) study showed that the number of \emph{projected gradient descent~(PGD) steps} to successfully attack a point (i.e., find an adversarial example in its proximity) is an effective measure of the robustness of this point.
Given that natural data are clean, this measure reveals an intrinsic geometric property---\emph{how far a point is from its class boundary}.
Based on this breakthrough, in this paper, we figure out how AT would interact with NL.
Firstly, we find if a point is \emph{too close} to its noisy-class boundary (e.g., \emph{one step} is enough to attack it), this point is likely to be \emph{mislabeled}, which suggests to adopt the number of PGD steps as a new criterion for \emph{sample selection} for correcting NL.
Secondly, we confirm AT with strong smoothing effects \emph{suffers less from NL} (without NL corrections) than \emph{standard training} (ST), which suggests AT itself is an NL correction.
Hence, AT with NL is helpful for improving even the natural accuracy, which again illustrates the superiority of AT as a \emph{general-purpose} robust learning criterion.
\end{abstract}

\section{Introduction}
\label{sec:intro}
In practice, the process of data labeling is usually noisy. Thus, it seems inevitable to learn with noisy labels~\citep{natarajan2013learning}. To combat noisy labels, researchers have designed robust label-noise learning methods, such as sample selection~\citep{jiang2018mentornet} and loss/label correction~\citep{patrini2017making,nguyen2019self}. Meanwhile, safety-critical areas (e.g., medicine and finance) require deep neural networks to be robust against adversarial examples~\citep{szegedy,Nguyen_2015_CVPR}. To combat adversarial examples, adversarial training methods empirically generate adversarial data on the fly for updating the model~\citep{Madry_adversarial_training,Zhang_trades}. 

%While the research community has extensively explored the \textit{adversarial training}~\citep{Madry_adversarial_training} and \textit{learning with noisy labels}~\citep{natarajan2013learning} independently, few studies have explored them jointly. 
%To combat adversarially corrupted input data, adversarial training methods generate adversarial data on the fly for updating the model [cite AT, and Fast AT, FAT]. 
%To combat corrupted labels, noisy-label learning methods select samples \citep{jiang2017mentornet,han2018co}[cite \footnote{some sample selection papers}] and correct labels [cite \footnote{some label corrections papers}]. Nevertheless, few has yet studied the learning with corrupted input data and corrupted labels jointly. 

% Two paragraph. 
An interesting fact is that, the research community is  exploring label-noise learning and adversarial training \textit{independently}. For example, \citet{ding2020mma} and \citet{zhang2020geometryaware} demonstrated that \textit{the non-robust data are close to the class boundary that are easy to be attacked}: their adversarial variants easily cross over the decision boundary. To fine-tune the decision boundary for adversarial robustness, \citet{ding2020mma} adaptively optimized small margins for the non-robust data; while \citet{zhang2020geometryaware} gave more weights on the non-robust data. However, both methods in adversarial training explore the adversarial robustness with an implicit assumption that \textit{data have clean labels}. Obviously, it is not realistic in practice. To this end, we figure out the interaction of adversarial training with noisy labels. 

%Those adversarial training methods implicitly assumes
%This inspires us to ask an interesting question: What if incorrect labels present?
\begin{figure}[tp!]
\vspace{0mm}
    \centering
    \includegraphics[scale=0.38]{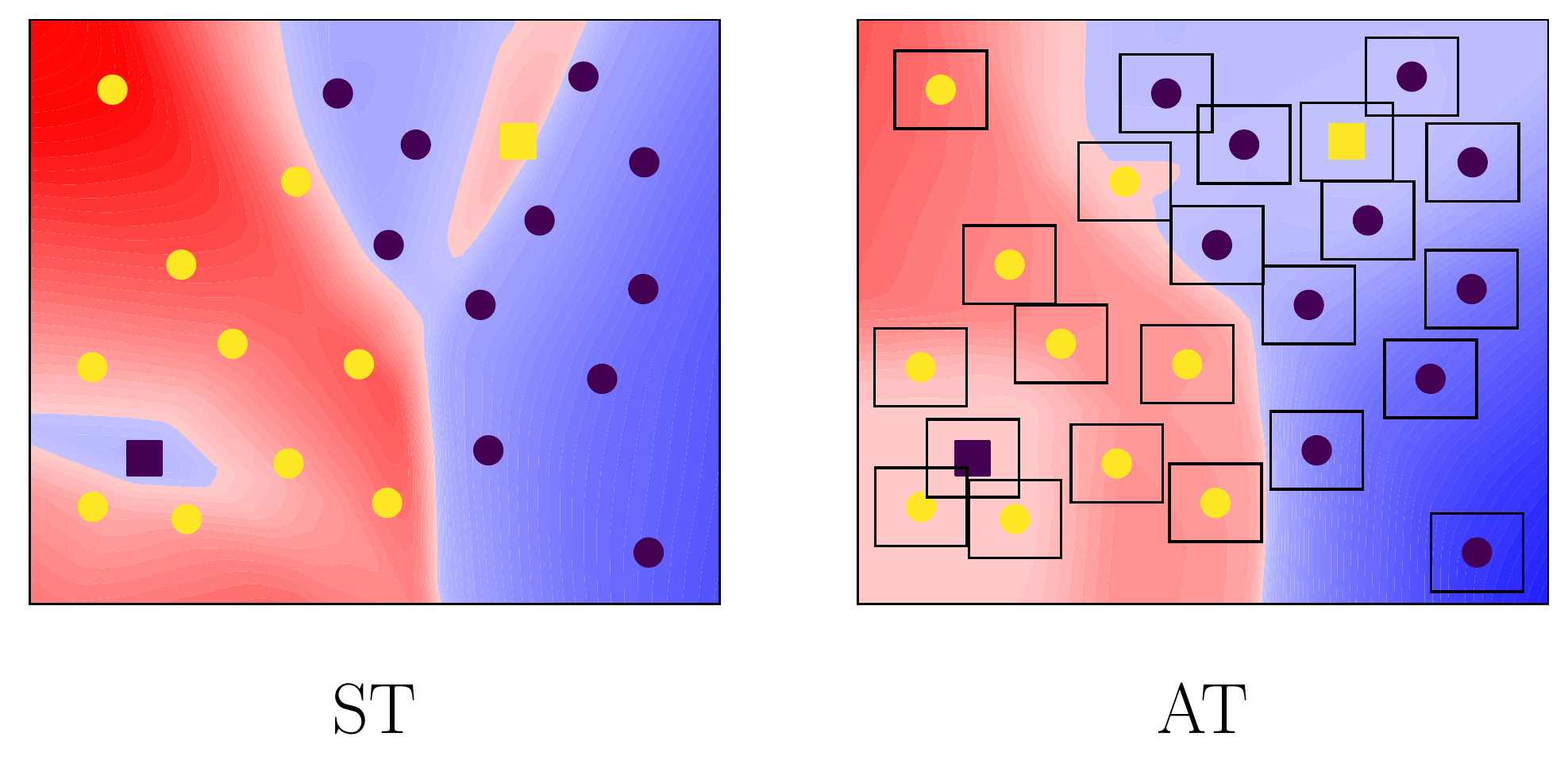}
    \vspace{-2mm}
    \caption{The results of ST and AT on a binary dataset with noisy labels. Dots denote correct data; squares denote incorrect data. The color gradient represents the prediction confidence: the deeper color represents higher prediction confidence. \textbf{Left panel}: deep network shapes two small clusters (red and blue ones in cross-over areas) around two incorrect data due to memorization effects in ST. \textbf{Right panel}: these clusters have been smoothed out in AT. Boxes represent the norm ball of AT.}
    \vspace{-3mm}
    \label{fig:illustration}
\end{figure}

We discover that when noisy labels occur in adversarial training (the right panel of Figure~\ref{fig:illustration}), \textit{incorrect data (square points) are more likely to be the non-robust data} (i.e., the predicted labels of their adversarial variants disagree with the given labels). Specifically, Figure~\ref{fig:illustration} compares the difference between standard training (ST, \citep{zhang_chiyuan_2017_rethink}) and adversarial training (AT, \citep{Madry_adversarial_training}) with noisy labels.
Commonly, a small number of incorrect data (square points) are surrounded by a large number of correct data (round points). In ST, deep network shapes two small clusters (the left panel of Figure~\ref{fig:illustration}) around the two incorrect data due to memorization effects~\citep{zhang_chiyuan_2017_rethink}. In contrast, AT has strong \textit{smoothing effects}, i.e., smoothing out the small clusters around incorrect data and letting incorrect data alone (the right panel of Figure~\ref{fig:illustration}).

To explain the above phenomenon in AT, we believe that the adversarial counterparts generated by (majority) correct data can help to smooth the local neighborhoods of correct data, which encourages deep networks to be locally constant within the neighborhood~\citep{papernot2016towards,goodfellow2016deep}. Therefore, in AT, it becomes difficult for deep networks to form small but separated clusters around each incorrect data. Consequently, these incorrect data are non-robust, which echos the parallel findings that robust training avoids memorization of label noise~\citep{sanyal2020benign}.

\begin{figure*}[t!]
\vspace{1mm}
    \centering
    \subfigure[CIFAR-10]{
    \includegraphics[scale=0.195]{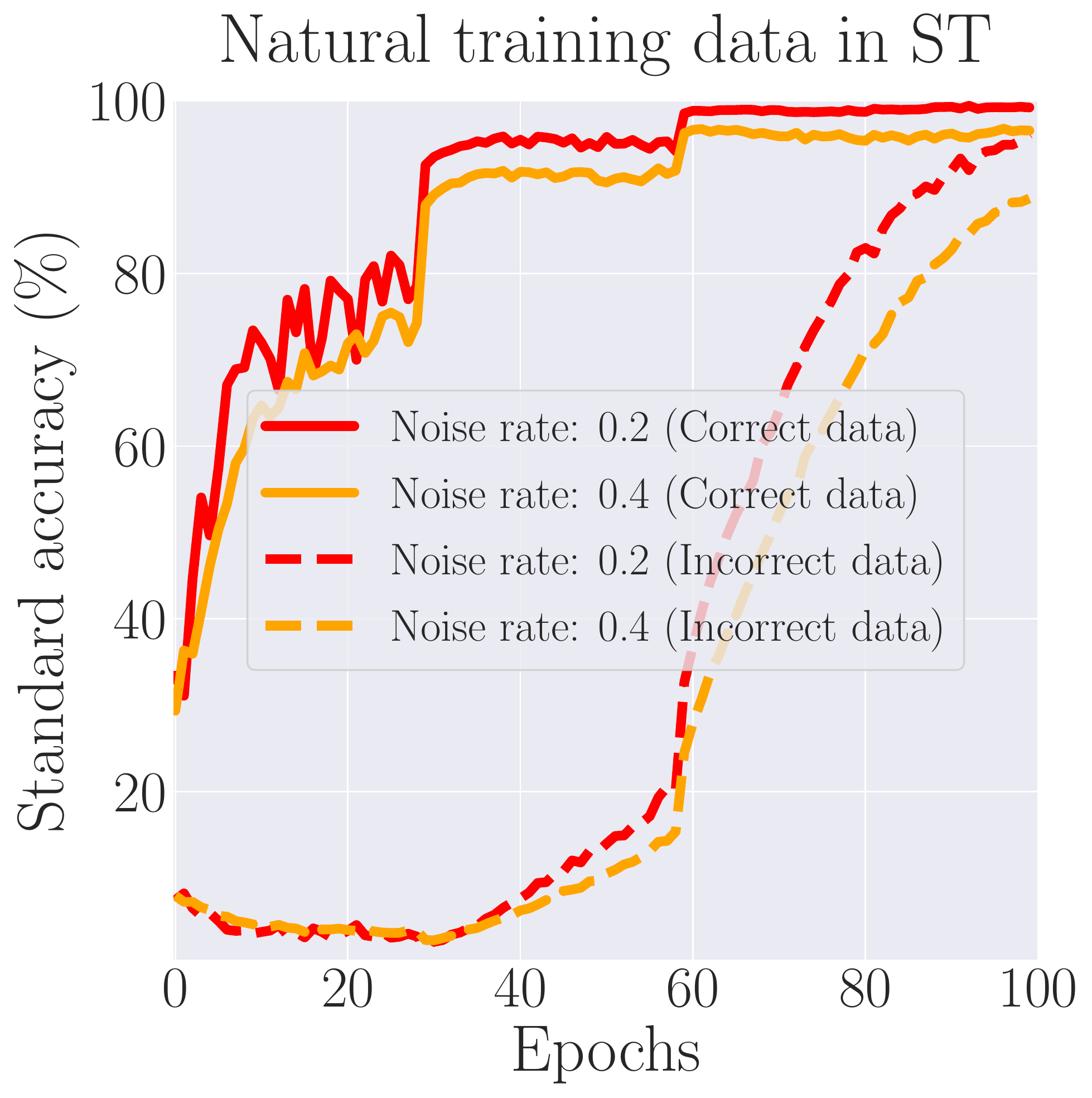}
    \includegraphics[scale=0.195]{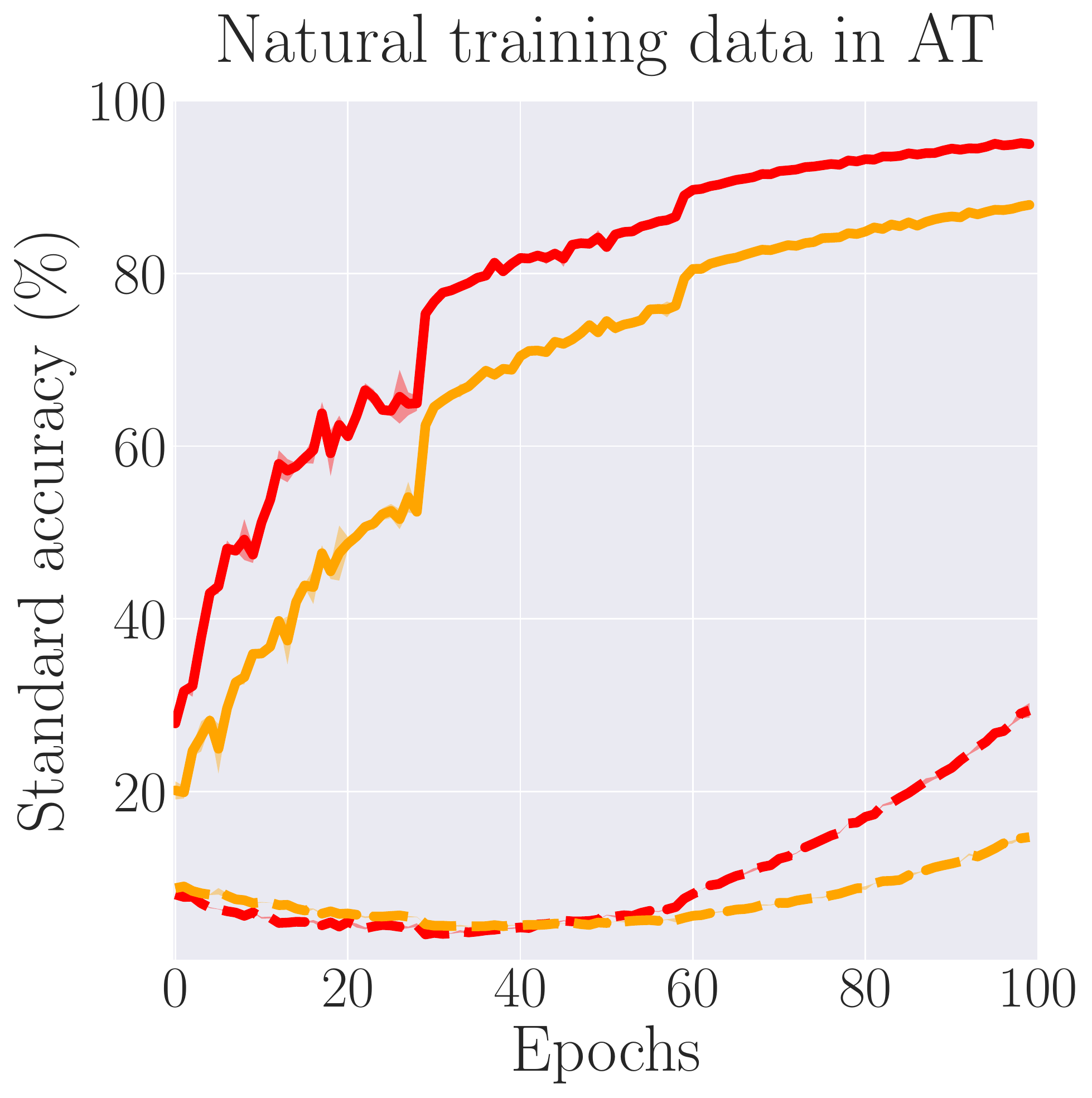}
    }
    \subfigure[MNIST]{
    \includegraphics[scale=0.195]{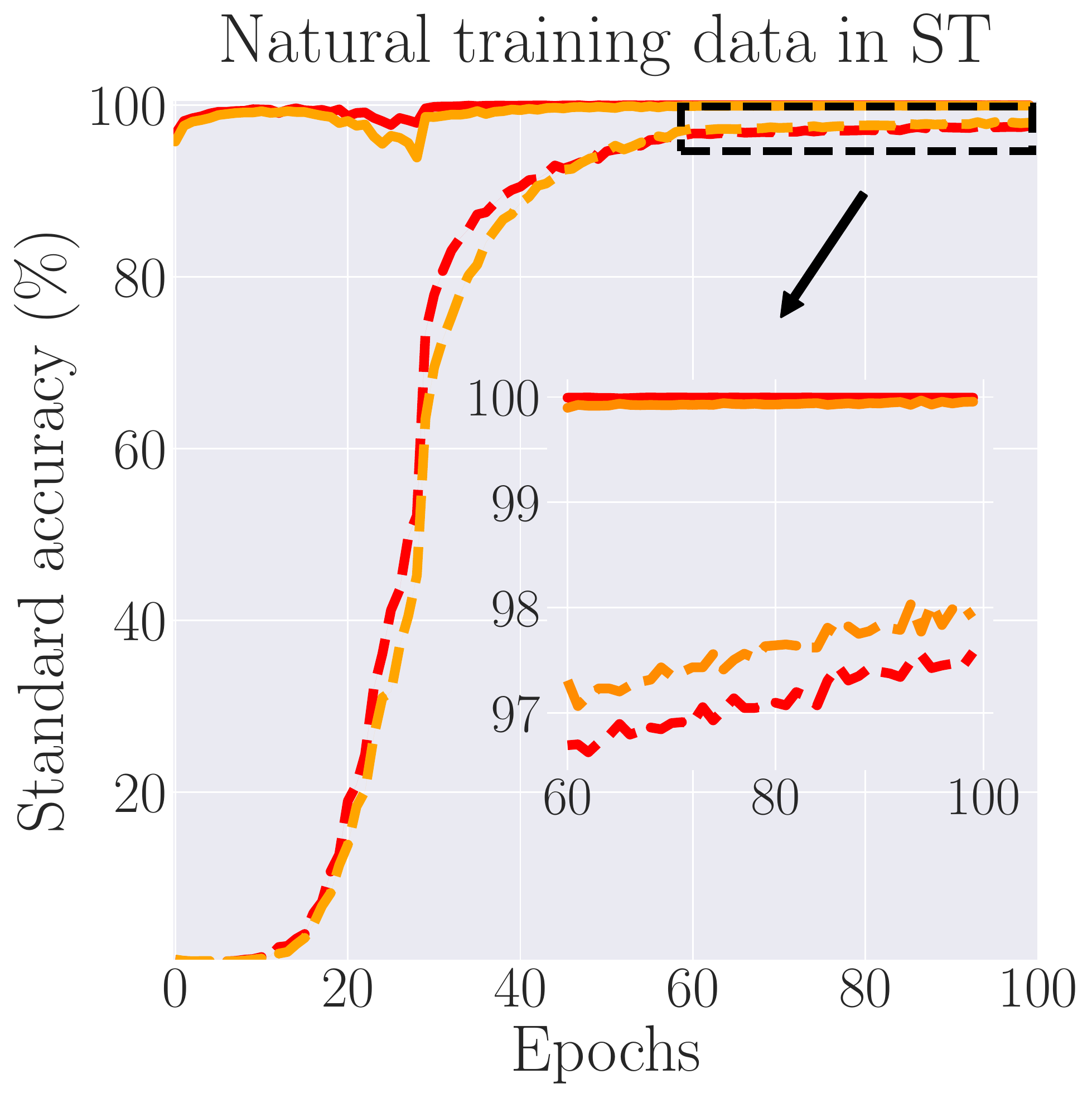}
    \includegraphics[scale=0.195]{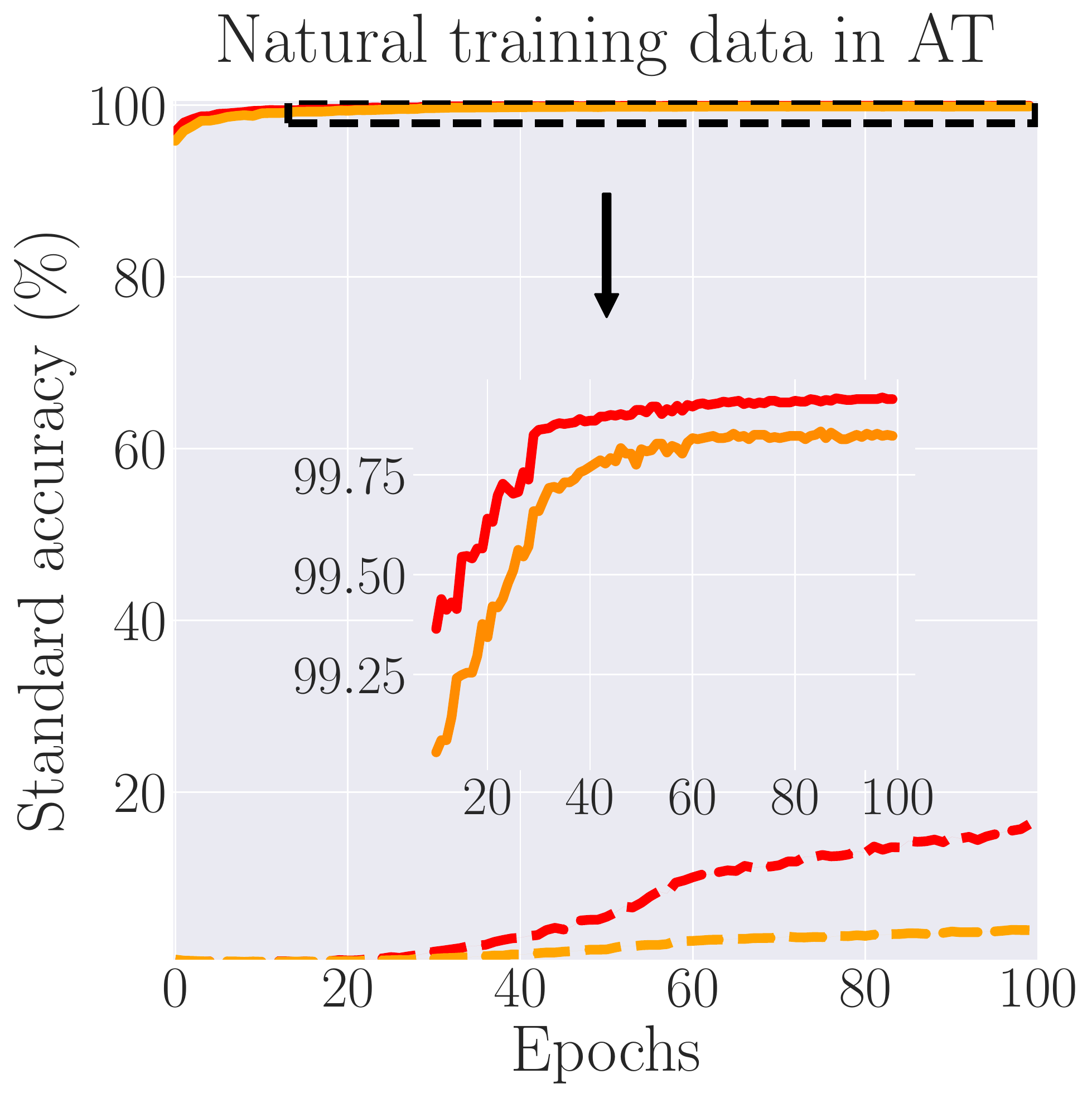} 
    }
    \vspace{-2mm}
    \caption{The standard accuracy of ST and AT on correct/incorrect training data using \textit{CIFAR-10} and \textit{MNIST} with symmetric-flipping noise. Solid lines denote the accuracy of correct training data, while dashed lines correspond to that of incorrect training data. Compared with ST, there is a large performance gap in the standard accuracy of correct/incorrect training data in AT.}
    \vspace{-2mm}
    \label{fig:part1_diff_st_at_label_noise_train_acc}
\end{figure*}

Furthermore, we make quantitative comparisons between ST and AT in the presence of label noise. From Figure~\ref{fig:part1_diff_st_at_label_noise_train_acc}, it can be seen that the training accuracy of deep networks on incorrect data is obviously lower than that on correct data in AT. Nonetheless, the performance gap totally disappeared in ST. Therefore, compared to ST, AT can always distinguish correct data and incorrect data. Observing Figure~\ref{fig:part1_diff_st_at_label_noise_test_acc}, the test accuracy of deep networks first increases then decreases in ST. Nonetheless, such a trend has been largely alleviated or totally eliminated in AT. Therefore, AT can mitigate negative effects of noisy labels, since the \textit{smoothing effects} of AT can prevent memorizing such incorrect data. Moreover, ~\citet{zhang_chiyuan_2017_rethink} showed that ST indeed overfits noisy labels, which definitely degrades the generalization performance of deep networks.

Under noisy labels, we realize that AT provides a new measure---\textit{how difficult it is to attack data to generate adversarial variants whose predictive labels are different from the given labels}---which can distinguish correct/incorrect data (Figures~\ref{fig:dis_pgd_loss_sym_02} and \ref{fig:dis_pgd_loss_asym_04}) and typical/rare data (Figure~\ref{fig:rare_classic_pgd_steps}) well. This new measure can be approximately realized by the number of projected gradient descent (PGD) steps~\citep{Madry_adversarial_training}, i.e., how many PGD iterations we need to generate misclassified adversarial variants. Compared with the commonly used measure, i.e., the \textit{loss value}~\citep{jiang2018mentornet,han2018co}, we find that the \textit{number of PGD steps} could be an alternative or even better measure in AT (Figures~\ref{fig:dis_pgd_loss_sym_02} and \ref{fig:dis_pgd_loss_asym_04}). Moreover, we discover that this new measure can pick up rare (atypical) data among typical data (Figure~\ref{fig:rare_classic_pgd_steps}), where modern datasets often follow long-tailed distributions~\citep{feldman2020does,feldman2020neural}.

\paragraph{Main contributions.} To sum up, our contributions can be summarized as three aspects as follows.
%\vspace{-3mm}
\begin{itemize}
    \item 1) We explore the in-depth interaction of adversarial training with noisy labels. Namely, we take a closer look at the smoothing effects of AT under label noise (Section~\ref{sec:geo_smoothing}). Subsequently, we conduct quantitative comparisons. Compared with ST, AT can always distinguish correct and incorrect data and mitigate negative effects of label noise (Section~\ref{sec:diff_st_at}).
    % \vspace{-1.5mm}
    \item 2) We realize that AT naturally provides a new measure called \textit{the number of PGD steps}, i.e., how many PGD iterations are needed to generate misclassified adversarial examples. Such a new measure can clearly differentiate the correct/incorrect data and typical/rare data (Section~\ref{sec:geo_kappa}).
    \item 3) We provide two simple examples of the applications of our new measure: a) we develop a robust annotator, which can robustly annotate unlabeled (U) data considering that U data could be adversarially perturbed (Section~\ref{sec:app_robust_annotator}); b) our new measure could be an alternative to the predictive probability for providing the confidence of annotated labels (Section~\ref{sec:confidence_score}).
\end{itemize}
\vspace{-3mm}

\section{Background}
\label{sec:background}

%(Review AT and review GAIRAT. (AT, PGD, FAST and FAT, GAIRAT))

%In this section, we briefly overviewed the adversarial training and label-noise learning as follows.

In this section, we give a brief overview of adversarial training and label-noise learning.

%1 AT formulation
%2 PGD

\paragraph{Adversarial training.} 
As one of the primary defenses against adversarial examples~\citep{Goodfellow14_Adversarial_examples,Carlini017_CW,Athalye_ICML_18_Obfuscated_Gradients}, adversarial training (AT) has been widely studied to improve the adversarial robustness of deep neural networks (DNNs)~\citep{Cai_CAT,carmon2019unlabeled,wang2020improving_MART,jiang2020robust,bai2021improving,chen2021robust}. The key objective of AT is to minimize the training loss on the adversarial variants of training data. We formally review the details of AT~\citep{Madry_adversarial_training} used in this paper:

Let $(\cX,d_\infty)$ denote the input feature space $\cX$ with the infinity distance metric $d_{\inf}(\bx,\bxprime)=\|\bx-\bxprime\|_\infty$, and $\epsball[\bx] = \{\bxprime \in \cX \mid d_{\inf}(\bx,\bx')\le\epsilon\}$
be the closed ball of radius $\epsilon>0$ centered at $\bx$ in $\cX$.
$S = \{ ({x}_i, y_i)\}^n_{i=1}$ is a dataset, where ${x}_i \in \cX$ and $y_i \in \cY =  \{0, 1, \ldots, C-1\}$. The objective function of AT is 
\begin{align}
\label{eq:adv-obj}
\min_{f_{\theta}\in\cF} \frac{1}{n}\sum_{i=1}^n \ell(f_{\theta}(\xadv_i),y_i),
\end{align}
where $\bxtidle$ is an adversarial variant of input data within the $\epsilon$-ball centered at ${x}$, $f_{\theta}(\cdot):\cX\to\bR^C$ is a score function, and the loss function $\ell:\bR^C\times\cY\to\bR$ is a composition of a base loss $\ell_\textrm{B}:\Delta^{C-1}\times\cY\to\bR$ (e.g., the cross-entropy loss) and an inverse link function $\ell_\textrm{L}:\bR^C\to\Delta^{C-1}$ (e.g., the soft-max activation), in which $\Delta^{C-1}$ is the corresponding probability simplex---in other words, $\ell(f_{\theta}(\cdot),y)=\ell_\textrm{B}(\ell_\textrm{L}(f_{\theta}(\cdot)),y)$. 

%AT employs the most adversarial data generated according to Eq.~(\ref{eq:madry_obj}) for updating the current model.

To generate the adversarial variants, AT employs the projected gradient descent (PGD) method~\citep{Madry_adversarial_training}. Given a starting point ${x}^{(0)} \in \cX$ and step size $\alpha > 0$, PGD works as follows:
\begin{normalsize}
\begin{equation}
    \label{PGD-k}
    {x}^{(t+1)} = \Pi_{\mathcal{B}[{x}^{(0)}]} \big( {x}^{(t)} +\alpha \sign (\nabla_{{x}^{(t)}} \ell(f_{\theta}({x}^{(t)}), y )  )  \big ) , 
\end{equation}
\end{normalsize}
until a certain stopping criterion is satisfied. In the above equation, $t \in \mathbb{N}$, $\ell$ is the loss function, ${x}^{(0)}$ refers to natural data or natural data perturbed by a small Gaussian or uniform random noise, $y$ is the corresponding label for natural data, ${x}^{(t)}$ is adversarial data at step $t$, and $\Pi_{\epsball[{x}_0]}(\cdot)$ is the projection function that projects the adversarial data back into the $\epsilon$-ball centered at ${x}^{(0)}$ if necessary. 

It is common to use PGD to generate adversarial variants $\tilde{x}$ in adversarial training methods~\citep{wong2020fast_zico_kolter,zhang2020fat}. Recently, \citet{zhang2020geometryaware} explored adversarial robustness by giving more weights on the non-robust data with the assumption that all labels are correct. Specifically, the non-robust data are geometrically close to the class boundary, which can easily go across the class boundary by a small perturbation. To approximate the distance between the data and the class boundary, they proposed the geometry-aware projected gradient descent (GA-PGD) to calculate the \textit{geometry value $\kappa$}, which is the least number of iterations that PGD needs to find the misclassified adversarial variants of input data. In this paper, we utilize the geometry value $\kappa$ to represent
our proposed measure (i.e., the \textit{number of PGD steps}); we further explore its applications such as selecting correct/incorrect and typical/rare data (Section~\ref{sec:geo_kappa}), assisting to develop robust annotator (Section~\ref{sec:app_robust_annotator}) and providing the annotation confidence (Section~\ref{sec:confidence_score}).

%After each PGD iteration, we will check whether the model prediction is equal to the given label of the data. The geometry value $\kappa$ will be incremented until the misclassified adversarial example is found. 

\paragraph{Label-noise learning.}
We consider a training set with $\cX = (x_1,\ldots,x_N)$ drawn i.i.d.~from some unknown distribution and its associated labels are $\cY = (y_1,\ldots,y_N)$, where $y_i \in \cY$ is the one-hot label for the instance $x_i$. In the setting of label noise, we observe noisy labels $\widetilde{\cY} = (\tilde{y}_1,\ldots,\tilde{y}_N)$ where $\tilde{y}_i \in \widetilde{\cY}$ might be different from the corresponding ground-truth label $y_i \in \cY$. In this paper, we mainly focus on typical class-conditional noise: 1) \textit{symmetric-flipping noise}~\citep{van2015learning}, where noisy labels are corrupted at random with the uniform distribution; 2) \textit{pair-flipping noise}~\citep{han2018co}, where noisy labels are corrupted between adjacent classes that are prone to be mislabeled. Note that pair-flipping noise is an extremely hard case of asymmetric-flipping noise~\citep{patrini2017making}.

To combat noisy labels, researchers have designed robust label-noise learning methods, such as sample selection~\citep{malach2017decoupling,jiang2020beyond,han2020sigua}, loss correction~\citep{han2020training,liu2020peer} and label correction~\citep{wang2018iterative}. Among them, sample selection is emerging due to its simplicity. The key idea of sample selection is to back-propagate clean samples (regarded as correct data) during training. Since DNNs learn simple patterns first~\citep{zhang_chiyuan_2017_rethink,arpit2017closer}, the \textit{loss value} is used as a general criterion for selecting clean samples~\citep{han2018co,yao2020searching}. Specifically, the data with small-loss values are considered as clean samples, which are used to update the model. In contrast, the data with large-loss values are considered as noisy samples, which should be discarded or utilized in another way~\citep{han2020sigua}.

\paragraph{Discussion.}
Previously, few studies focused on corrupted features and labels jointly. For example, \citet{teng1999correcting} corrupted features randomly with symmetric label noise. However, they considered neither adversarial corruption nor asymmetric label noise. \citet{huang2020self} designed a self-adaptive method to learn with noisy labels or adversarial examples robustly. Nonetheless, they did not consider the co-existence of adversarial examples and noisy labels. \citet{sanyal2020benign} identified label noise as one of the causes for adversarial vulnerability, but they only justified symmetric label noise. In this paper, instead of designing new algorithms along two different directions, our focus is to \textit{figure out} the interaction of adversarial training with (generalized) noisy labels. By studying the performance of AT with correct/incorrect data, we discover some interesting findings (Sections~\ref{sec:geo_smoothing} and~\ref{sec:diff_st_at}) and provide a new measure (Section~\ref{sec:geo_kappa}) and its applications (Section~\ref{sec:app_kappa}).

% 2-3 pages

%Compared with adversarial training using clean data, noisy-labeled data in adversarial training cause degradations of both standard accuracy and robust accuracy using PGD-20 attack. (Figure~\ref{fig:pre_effect})

%Show analysis about the noisy-labeled data in adversarial training. (Visualization by PCA or TSNE: most of the noisy-labeled data are surrounded by clean data. (similar to Figure~\ref{fig:illustration}))

\begin{figure}[tp!]
\vspace{1mm}
    \centering
    \includegraphics[scale=0.28]{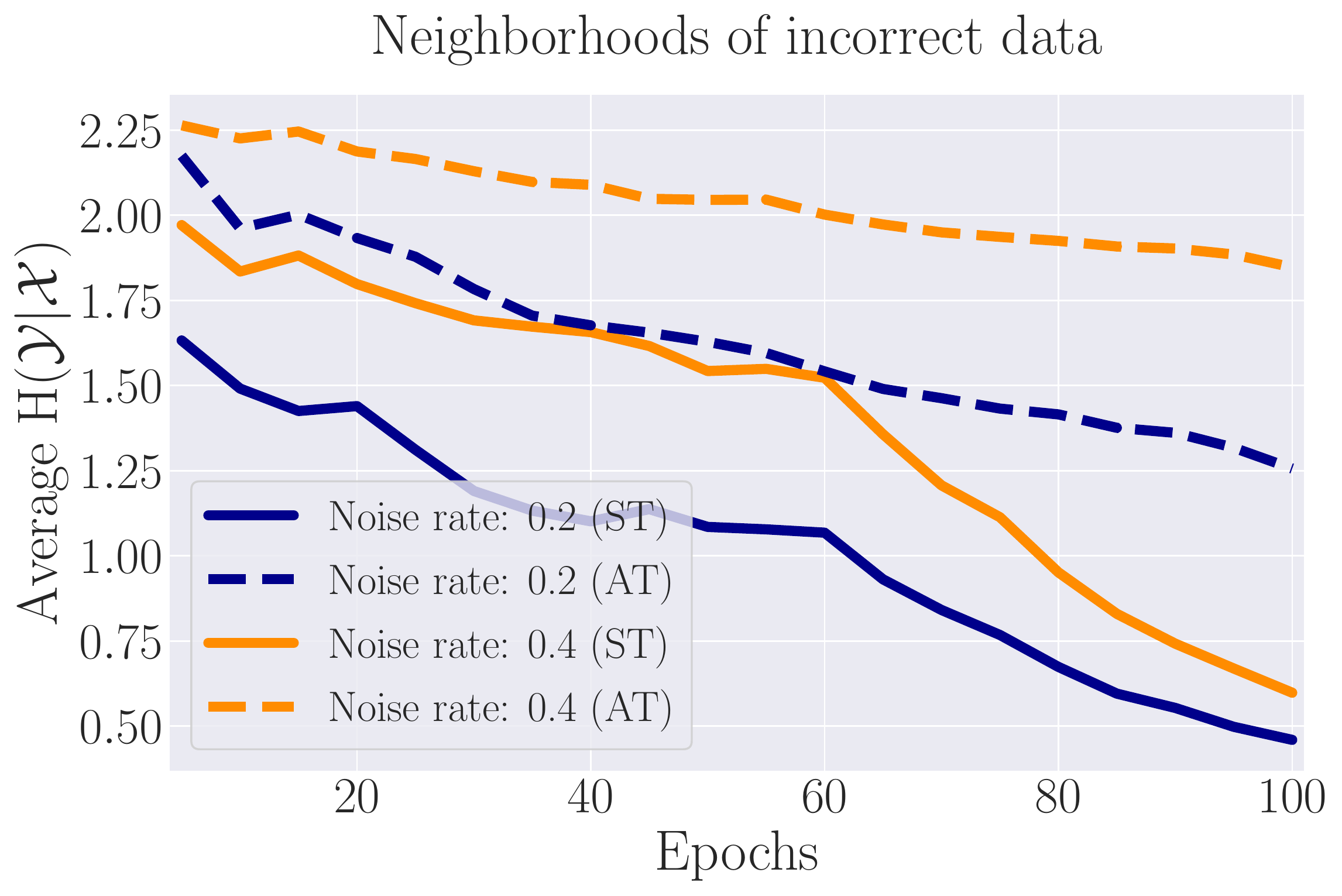}
    \vspace{-2mm}
    \caption{The average entropy of models trained by ST and AT. This value is calculated on $100$ points in each neighborhood of incorrect data, using \textit{CIFAR-10} with symmetric-flipping noise. Both \textit{solid} and \textit{dashed} lines represent ST and AT, respectively. Note that ST learns incorrect data more deterministically than AT.}
    \vspace{-3mm}
    \label{fig:entropy_plot}
\end{figure}

\section{Smoothing Effects of Adversarial Training}
\label{sec:geo_smoothing}

%(\textbf{What new} points have been discovered; \textbf{How to verify}; a \textbf{reasonable explanation} for the new findings; discussion)

In this section, we take a closer look at the smoothing effects of AT with noisy labels. At a high level, we conduct experiments on a synthetic dataset with incorrect labels, which explicitly show the smoothing effects of AT (Figure~\ref{fig:illustration}). We then use a real-world dataset, \textit{CIFAR-10}~\citep{krizhevsky2009learning_cifar10}, with incorrect labels, which further validates the smoothing effects of AT (Figure~\ref{fig:entropy_plot}). As a key result, we find that AT can smooth out the small clusters around incorrect data (the right panel of Figure~\ref{fig:illustration}), which leads to incorrect data being non-robust in AT, i.e., easily attacked to flip labels. The setup can be found in Appendix~\ref{appendix:smoothing_effect}.

In detail, we empirically confirm that DNNs can memorize random noise in standard training (ST, the left panel of Figure~\ref{fig:illustration}), which has been found in previous works~\citep{zhang_chiyuan_2017_rethink,arpit2017closer}. However, a recent study~\citep{sanyal2020benign} claimed that AT can avoid the memorization of incorrect data through \textit{analyzing model predictions}. Going beyond their analysis, we further investigate AT with noisy labels and provide an in-depth explanation, namely \textit{smoothing effects}. Specifically, AT prevents incorrect data from forming small clusters during training, which should be the primary reason for avoiding the memorization of incorrect data.

%Mainpint: Incorrect data can not generate its own cluster for classification.

%(Cite ICLR 2021, the previous study has found that robust training can avoid memorization of incorrect data, here we take a closer look at the reason behind the phenomena.)

%This section aims to study the smoothing effect of adversarial training, which can help DNNs to avoid memorization of incorrect data. The incorrect data are hard to smooth their neighborhood due to a stronger effect of superposition from the surrounding correct data.

%The adversarial training methods seek to train a robust deep model whose predictions are locally invariant to a small neighborhood~\citep{Madry_adversarial_training,Zhang_trades}. Generally, many methods~\citep{carmon2019unlabeled,zhang2020fat,zhang2020geometryaware} employ the adversarial variants of input data to smooth their neighborhood. As for the incorrect data, which are geometrically surrounded by the correct data, 

To justify our smoothing effect, we perform a series of comparison experiments using ST and AT on a synthetic dataset with incorrect labels. In Figure~\ref{fig:illustration}, the model trained by ST can overfit the incorrect data (yellow and black squares), and thus have incorrect predictions (red and blue clusters in cross-over areas) around incorrect data. While in AT (with smoothing effects), such clusters have obviously disappeared. The reason is due to the smoothing effects from the adversarial variants generated from correct data. Namely, the number of correct data is larger than that of incorrect data. Thus, it is difficult for incorrect data to smooth their neighborhood. More results can be found in Appendix~\ref{appendix:smoothing_effect}.

Further, we calculate the entropy values of the model predictions on the \textit{CIFAR-10} dataset, which aims to validate the smoothing effect in practice. Specifically, we randomly select $100$ points in each neighborhood (within a small $\epsilon$-ball) of the incorrect data and calculate their average entropy values in training (Figure~\ref{fig:entropy_plot}). As a measure of uncertainty~\citep{dai2012entropy}, the entropy value is calculated by the following formula:
\begin{align}
\vspace{-6mm}
\label{eq:entropy}
H(\cY|\cX) = -\sum_{x\in \cX}\sum_{y\in \cY} p(x,y)\cdot\log{p(y|x)}.
\vspace{-6mm}
\end{align}
The smaller value represents the higher certainty of model prediction (and vice versa), which indicates that the model learns the data more deterministically. Thus, the higher certainty leads to the higher possibility of incorrect data forming small clusters in their neighborhoods. 

We compare the entropy values of ST and AT. During the training process, under the same noise rate, the entropy value of AT is always higher than that of ST. After epoch~$60$, the entropy value of ST drops very fast, while that of AT remains high. It clearly shows that smoothing effects in AT prevent the model from learning incorrect data with their neighborhoods deterministically, which further confirms that it is harder for incorrect data to form the small clusters. By observing Figures~\ref{fig:illustration} and~\ref{fig:entropy_plot}, we confirm that it is difficult for incorrect data in AT to form small clusters due to the smoothing effects from the adversarial variants of correct data. 

%Besides smoothing effects, we displayed positive knock-on effects on adversarial training with noisy labels as follows.

\begin{figure*}[t!]
%\vspace{1mm}
    \centering
    \subfigure[CIFAR-10]{
    \includegraphics[scale=0.195]{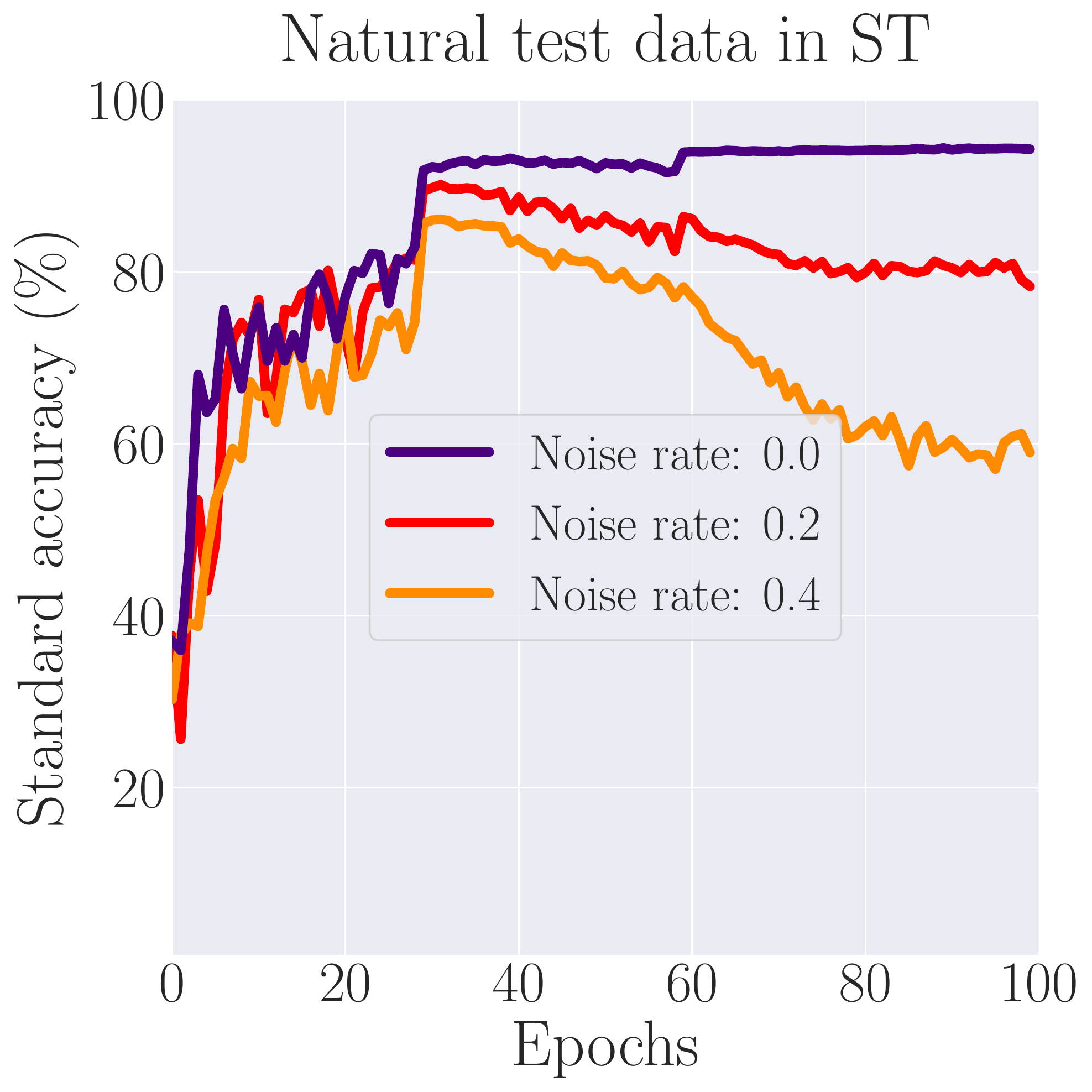}
    \includegraphics[scale=0.195]{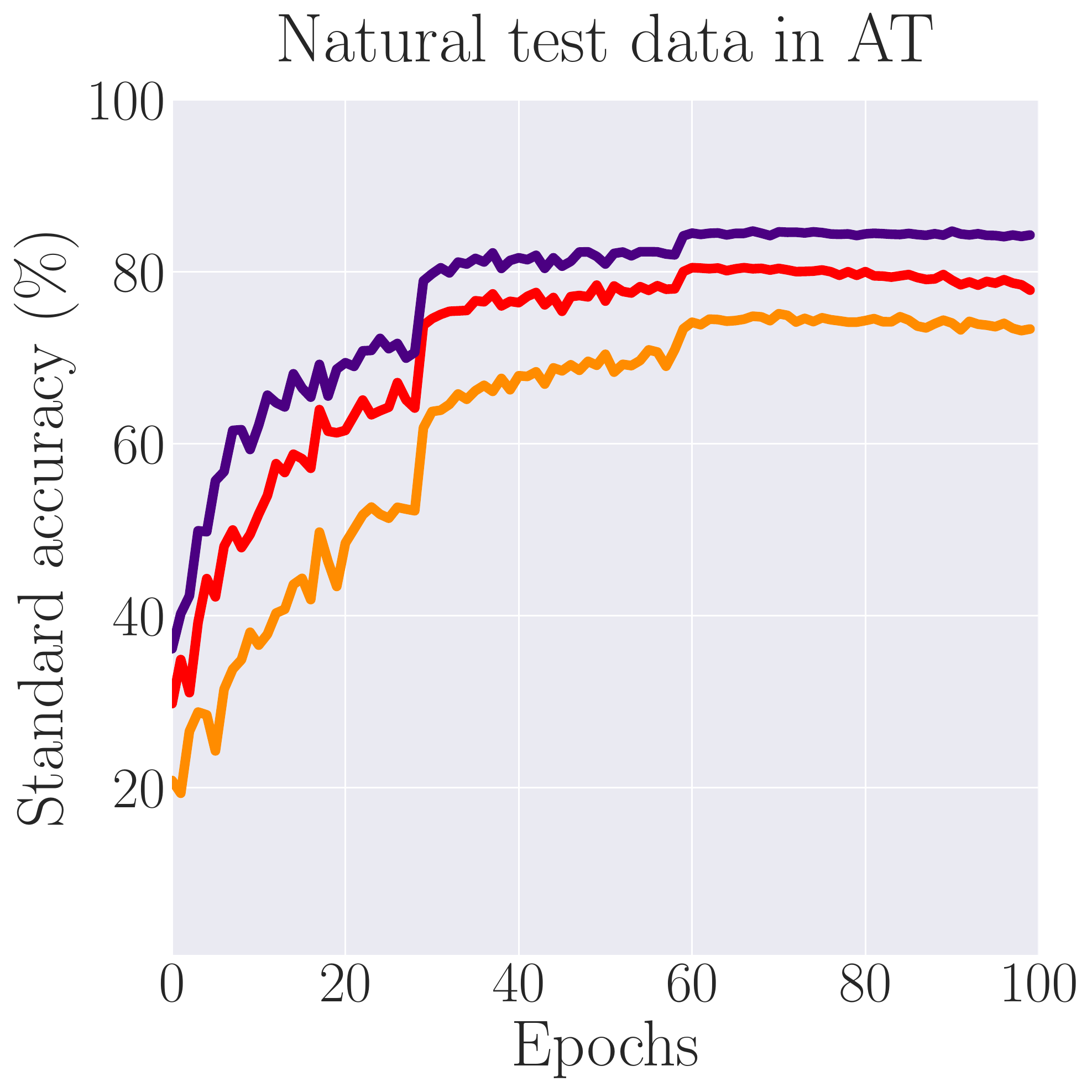}
    }
    \subfigure[MNIST]{
    \includegraphics[scale=0.195]{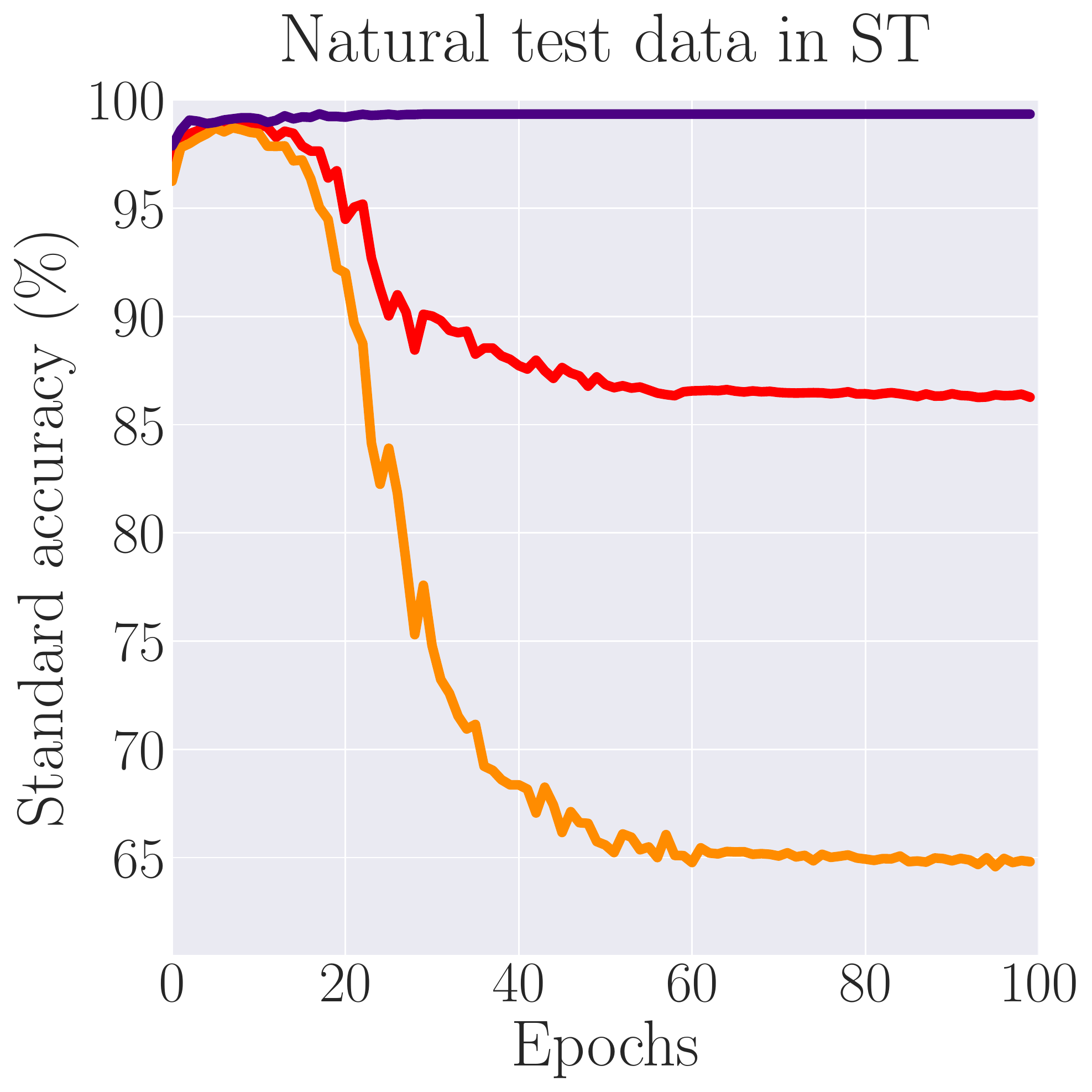}
    \includegraphics[scale=0.195]{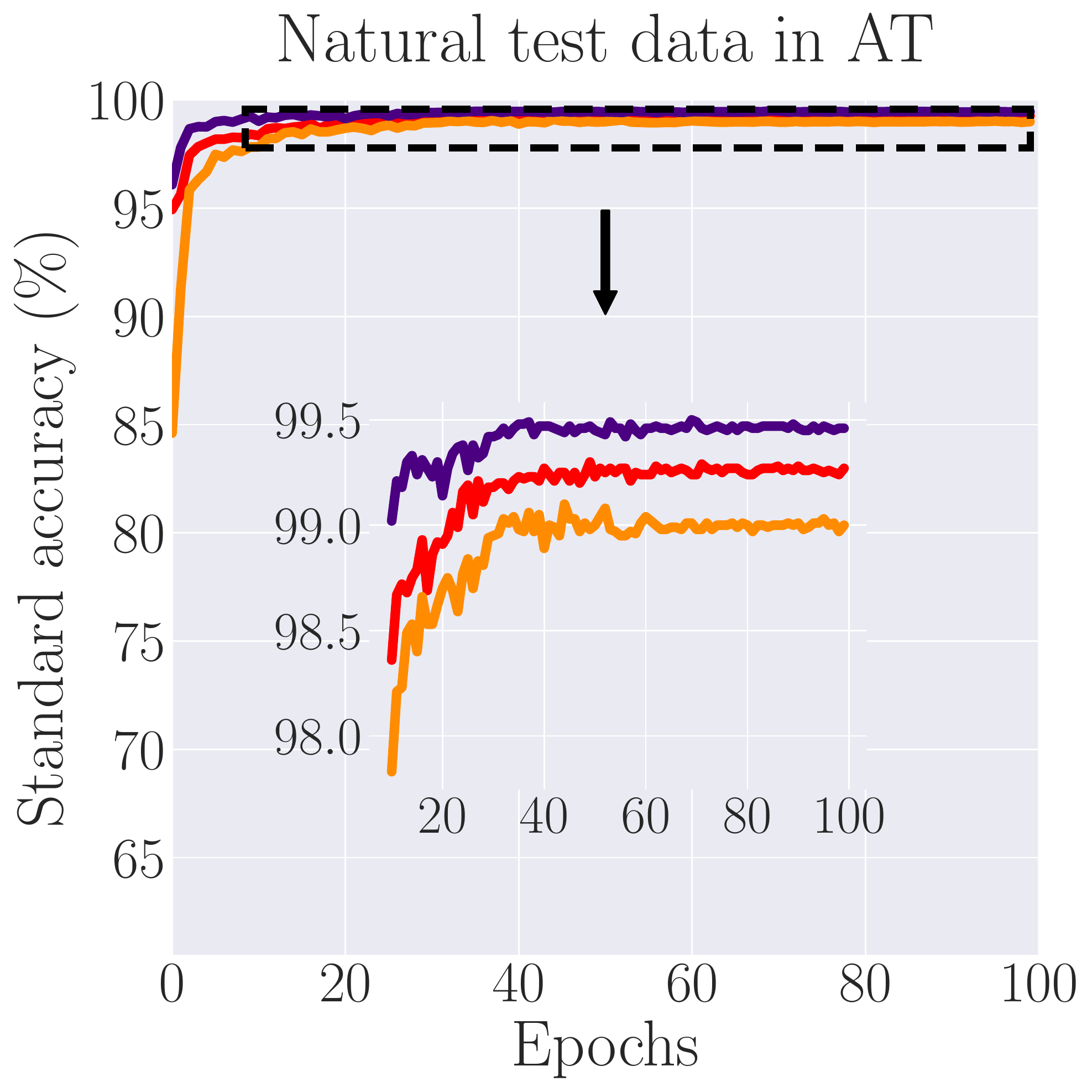}
    }
    \vspace{-2mm}
    \caption{The standard accuracy of ST and AT on natural test data using \textit{CIFAR-10} and \textit{MNIST} with symmetric-flipping noise for training. Note that the larger noise rate causes the test accuracy of ST dropping more seriously due to memorization effects in deep learning, while AT alleviates such negative effects.}
    \vspace{-3mm}
    \label{fig:part1_diff_st_at_label_noise_test_acc}
\end{figure*}

\section{Knock-on Effects of Adversarial Training}
\label{sec:diff_st_at}

%(\textbf{What new} points have been discovered; \textbf{How to verify}; a \textbf{reasonable explanation} for the new findings; discussion)
%Mainpoint: incorrect data and correct data are more distinguishable in AT than ST.
%(Mainpoint: Due to the smoothing effect, model avoids memorization of incorrect data, thus the incorrect data and correct data are more distinguishable, and the model generalization becomes better)

In this section, we explore knock-on effects of adversarial training comprehensively. We show the quantitative differences between ST and AT with noisy labels. First, in terms of training accuracy, we show that correct/incorrect data can be always distinguishable in AT (Figure~\ref{fig:part1_diff_st_at_label_noise_train_acc}). Second, in terms of test accuracy, we demonstrate that AT alleviates negative effects of incorrect data and then improves the model generalization (Figure~\ref{fig:part1_diff_st_at_label_noise_test_acc}). Note that we display the experimental results on the \textit{CIFAR-10} and \textit{MNIST} datasets~\citep{lecun1998gradient} with symmetric-flipping noise in this section. More results (e.g., pair-flipping noise, different networks, the loss value) can be found in Appendix~\ref{appendix:diff_st_at}.

%This section aims to explore the differences between the ST and AT using the training data with different noise ratios and noise types. In particular, we use the training/test accuracy, and the visualization of loss landscape to characterize this difference, which empirically show the smoothing effect of AT analyzed in Section~\ref{sec:geo_smoothing} and comprehensively demonstrate that correct data and incorrect data are more distinguishable in the AT than ST.

\subsection{Distinguishable Correct/Incorrect Data}

In Figure~\ref{fig:part1_diff_st_at_label_noise_train_acc}, we plot the standard accuracy of natural \textit{training} data in ST and AT. In the early stage of training, there is a clear performance gap between the standard accuracy of ST on correct/incorrect training data. However, after 60 epochs, the standard accuracy of ST on incorrect training data increases rapidly, while that of AT on incorrect training data rises relatively slowly. When the training comes to epoch $100$, the standard accuracy of ST on correct/incorrect training data are merged together. Nonetheless, there is still a large performance gap in AT. Compared to the results on \textit{CIFAR-10}, such a gap is more obvious on \textit{MNIST}. 

\begin{figure}[t!]
\vspace{-2mm}
    \centering
    \subfigure[ST]{
    \begin{minipage}[b]{0.2\textwidth}
        \includegraphics[scale=0.2]{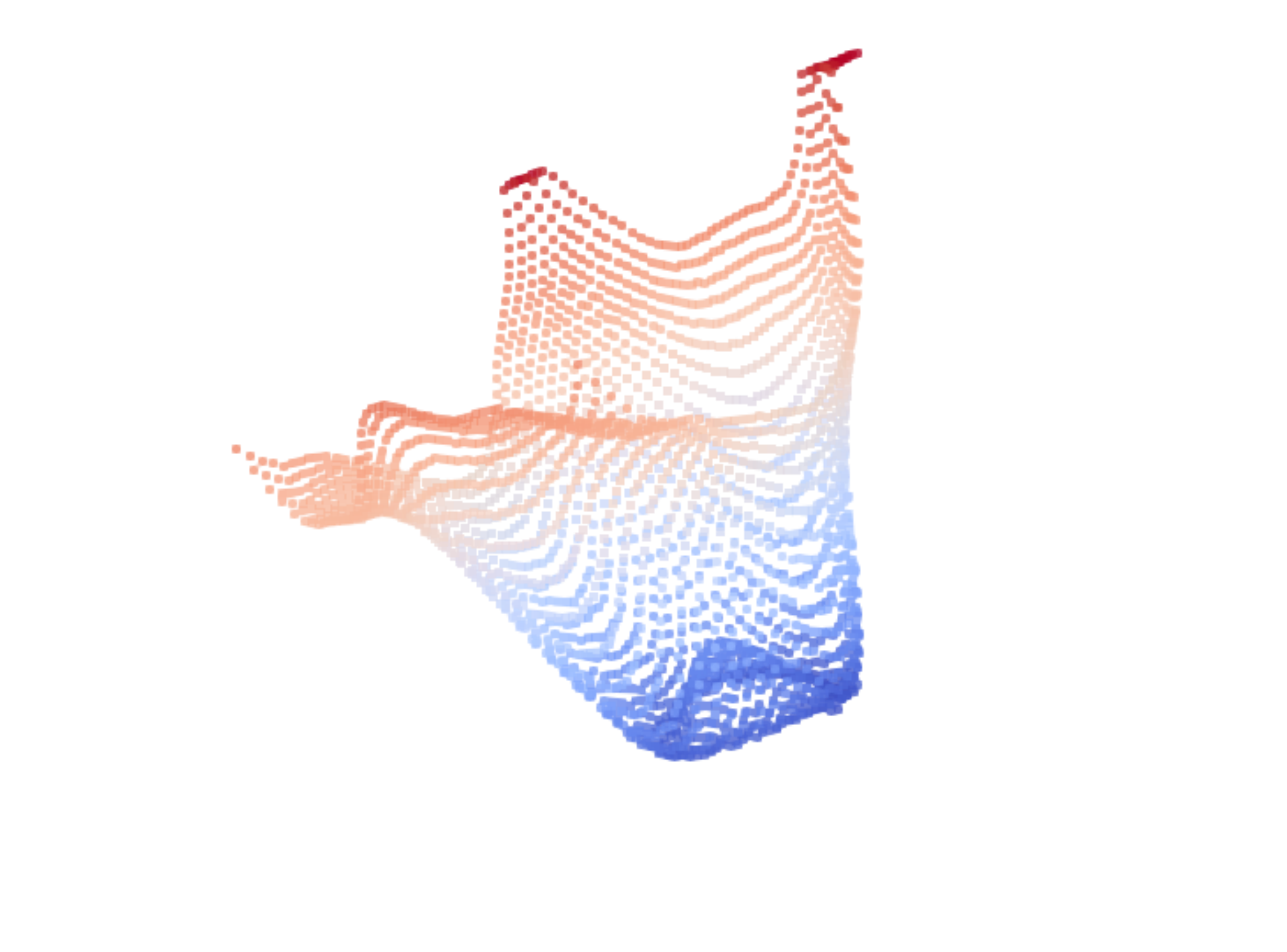}
    \end{minipage}
    }
    \subfigure[AT]{
    \begin{minipage}[b]{0.2\textwidth}
        \includegraphics[scale=0.2]{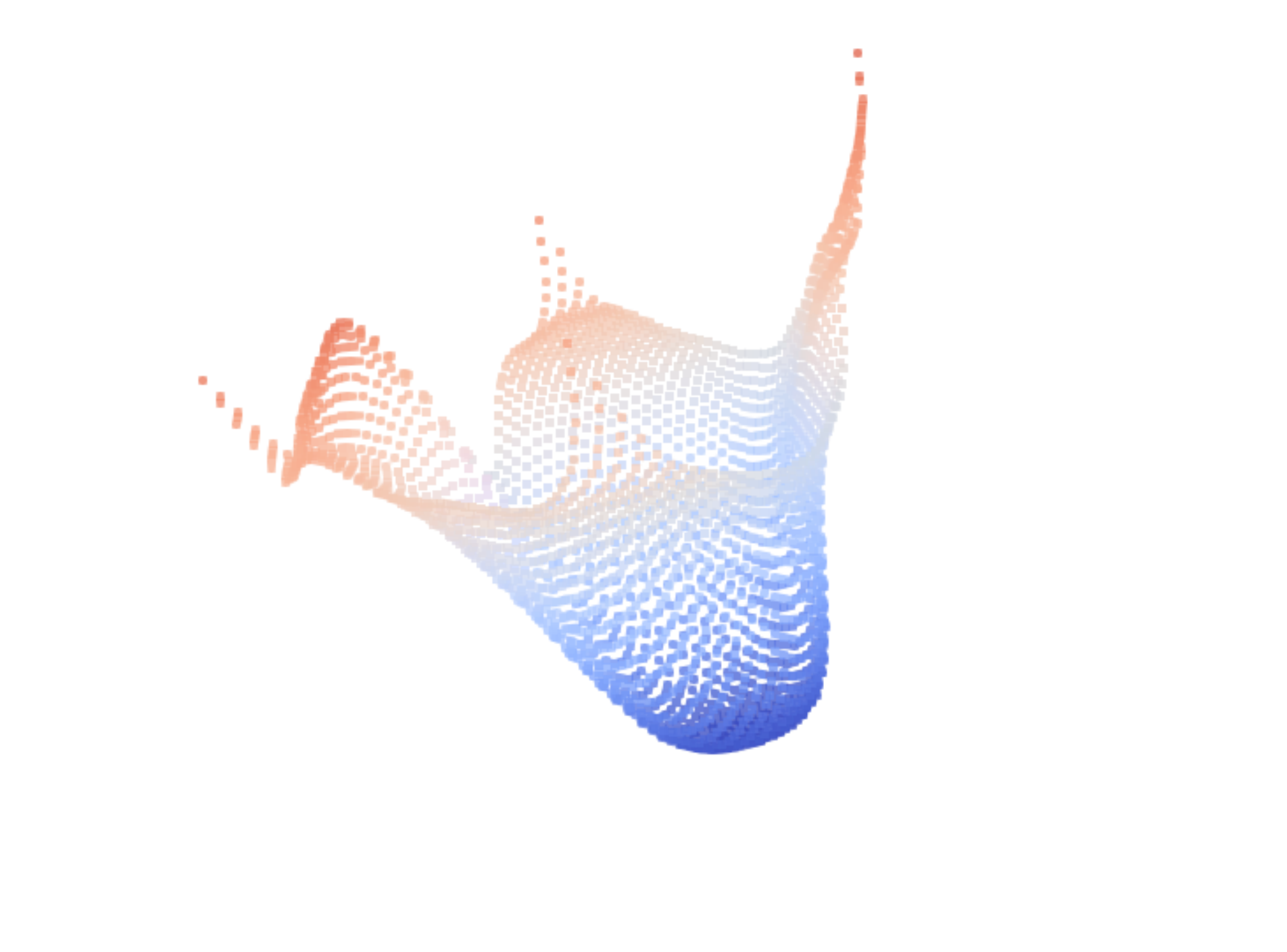}
        %\\
        \vspace{-4mm}
    \end{minipage}
    }
    \vspace{-2mm}
    \caption{The loss landscape w.r.t weight space of models trained by ST and AT using \textit{CIFAR-10} with $20\%$ symmetric-flipping noise. The red/blue colors denote  large/small values, which reflect the relative position in the loss landscape. Note that the loss landscape of a model trained by AT is smoother and flatter than that by ST, which reflects the better model generalization by AT.}
    \vspace{-4mm}
    \label{fig:loss_landscape}
\end{figure}

\subsection{Alleviation of Memorization Effects}

In Figure~\ref{fig:part1_diff_st_at_label_noise_test_acc}, we plot the standard accuracy of natural \textit{test} data in ST and AT. We show that AT can alleviate the negative effects of label noise and then improve the model generalization. Specifically, the larger noise rate causes the test accuracy of ST to drop more seriously, i.e., memorization effects~\citep{arpit2017closer}. However, AT reduces such negative effects. By checking the standard accuracy of natural test data, we find that there is no obvious overfitting phenomenon in AT. In Figure~\ref{fig:loss_landscape}, we visualize the loss landscape~\citep{li2018visualizing} of models trained by ST and AT. Such visualization can further substantiate that AT mitigates negative effects of label noise via the lens of the model generalization. Namely, the loss landscape w.r.t. weight space of an adversarially trained model (i.e., AT) is smoother and flatter than that of a model using ST.

% After these experiments in previous sections, we confirm that the smoothing effect of AT can prevent incorrect data from forming small clusters and help model to avoid the memorization of incorrect data. 
% As a result, the standard accuracy of incorrect training data is reduced, and the generalization performance of the model is improved. 

It is worthwhile to observe the results on \textit{MNIST}: \textit{simply using AT can make the model obtain a performance similar to noise-free training}. However, on more complex \textit{CIFAR-10}, incorrect data still have a certain negative impact on the model trained by AT. To reduce such an impact, a simple yet effective method is to use sample selection to filter correct/incorrect data for training~\citep{jiang2018mentornet,cheng2021learning}. Therefore, it is critical to have a measure which can provide the stratification for correct/incorrect data. Normally, the \textit{loss value} can be a good candidate in ST. However, in AT, we can find a better measure such as the \textit{number of PGD steps} (i.e., geometry value $\kappa$). Since the smoothing effects in AT can make incorrect data be non-robust, the geometry value $\kappa$---how difficult it is to attack data to let them go across the 
decision boundary---could be naturally used as a measure for this task.

\begin{figure}[h!]
\vspace{1mm}
    \centering
    %\includegraphics[scale=0.2]{resnet18_sym_01_loss.pdf}
    %\hspace{1mm}
    \hspace{-3mm}
    \includegraphics[scale=0.195]{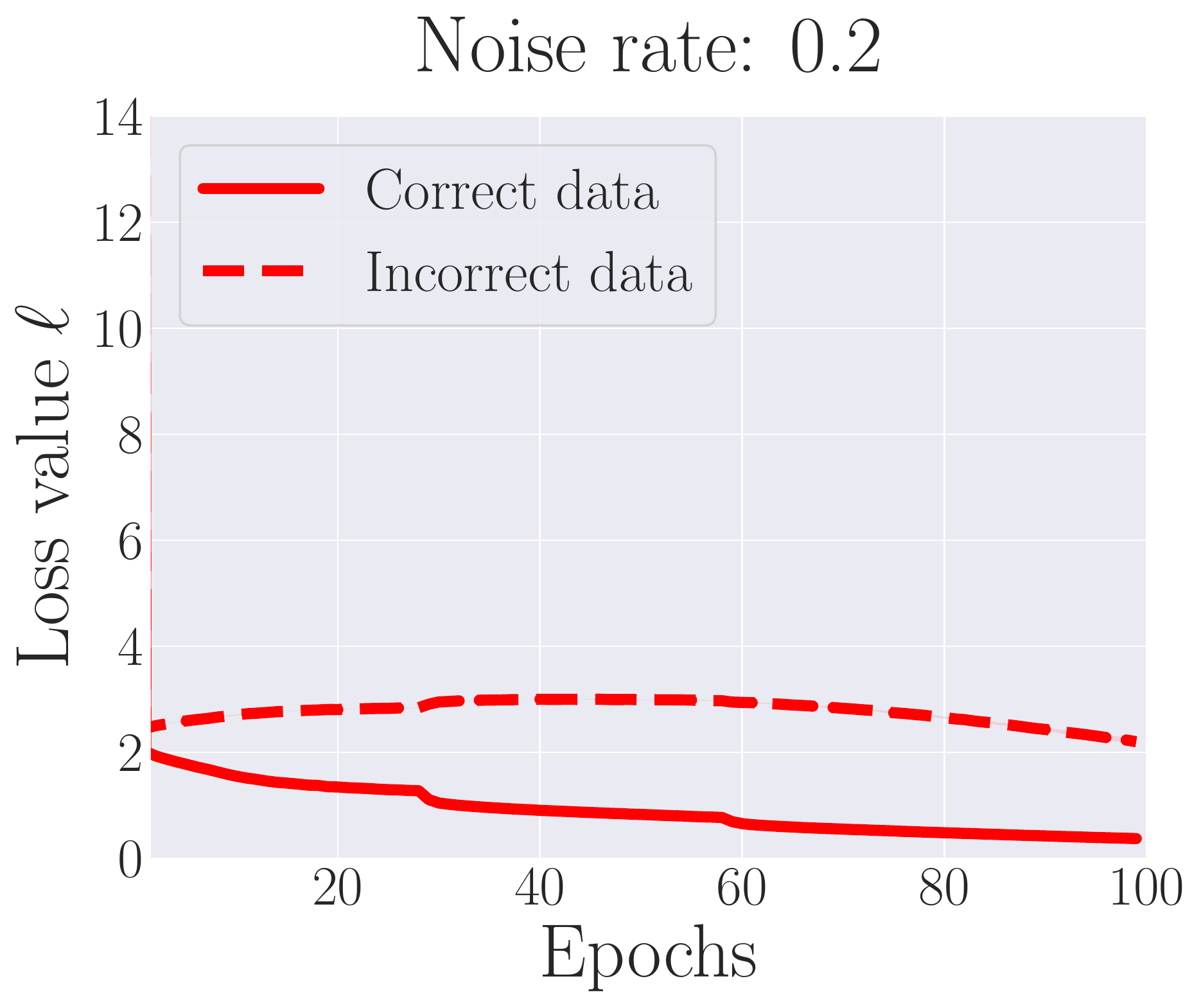}
    %\hspace{1mm}
    %\includegraphics[scale=0.2]{resnet18_sym_03_loss.pdf}
    %\hspace{1mm}
    \includegraphics[scale=0.195]{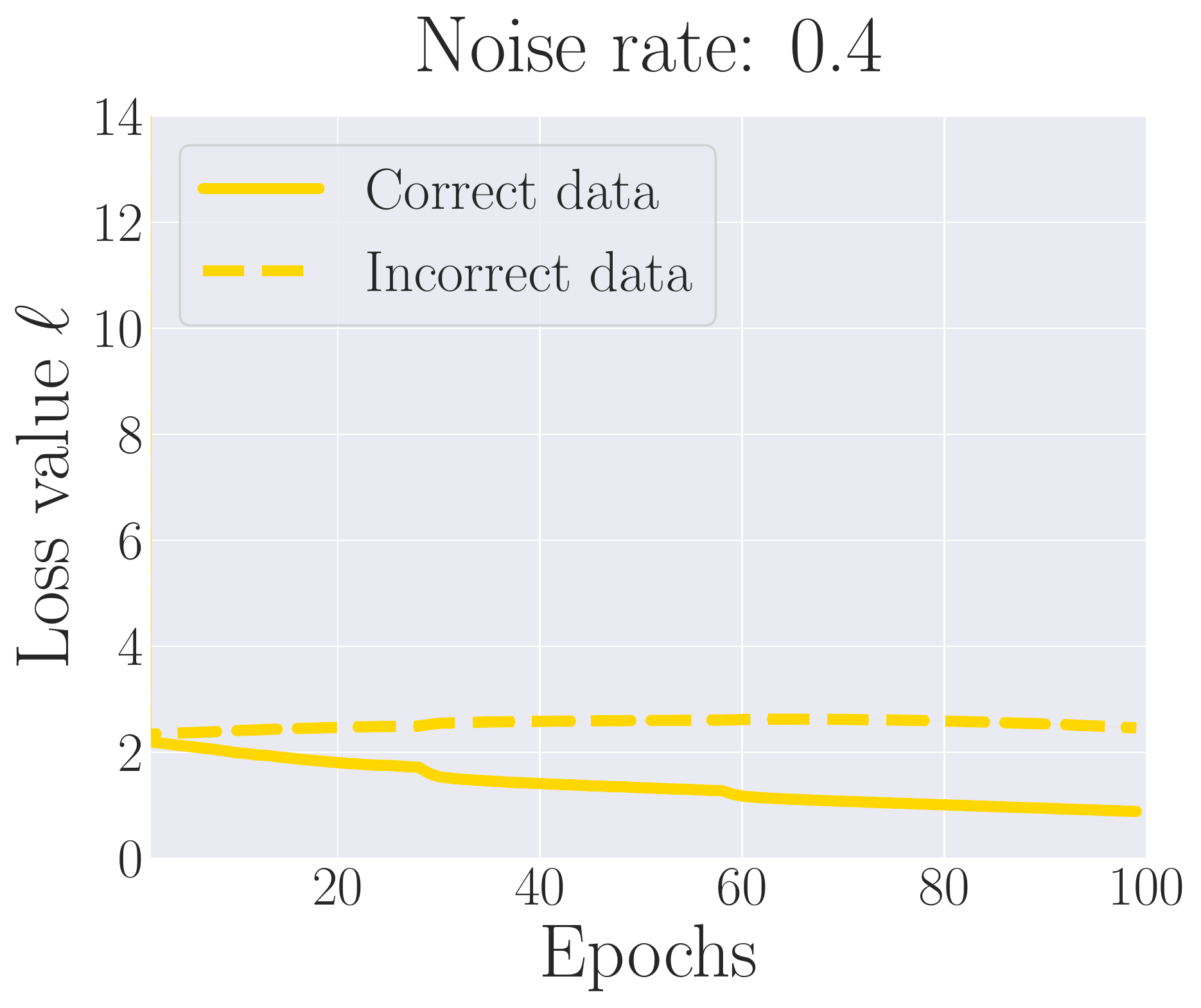}\\
    %\includegraphics[scale=0.2]{resnet18_sym_01_kappa.pdf}
    %\hspace{1mm}
    \includegraphics[scale=0.195]{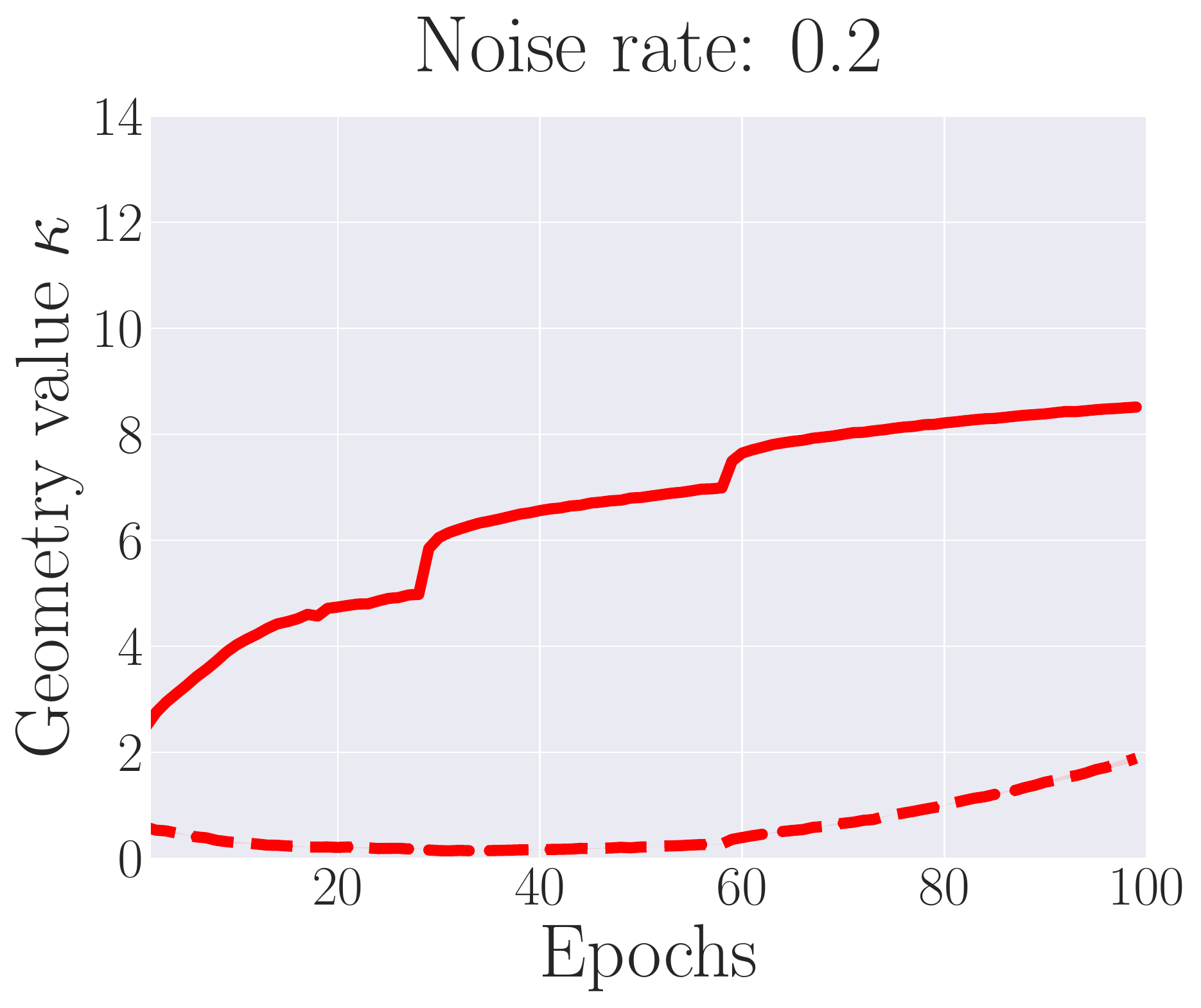}
    %\hspace{1mm}
    %\includegraphics[scale=0.2]{resnet18_sym_03_kappa.pdf}
    %\hspace{1mm}
    \includegraphics[scale=0.195]{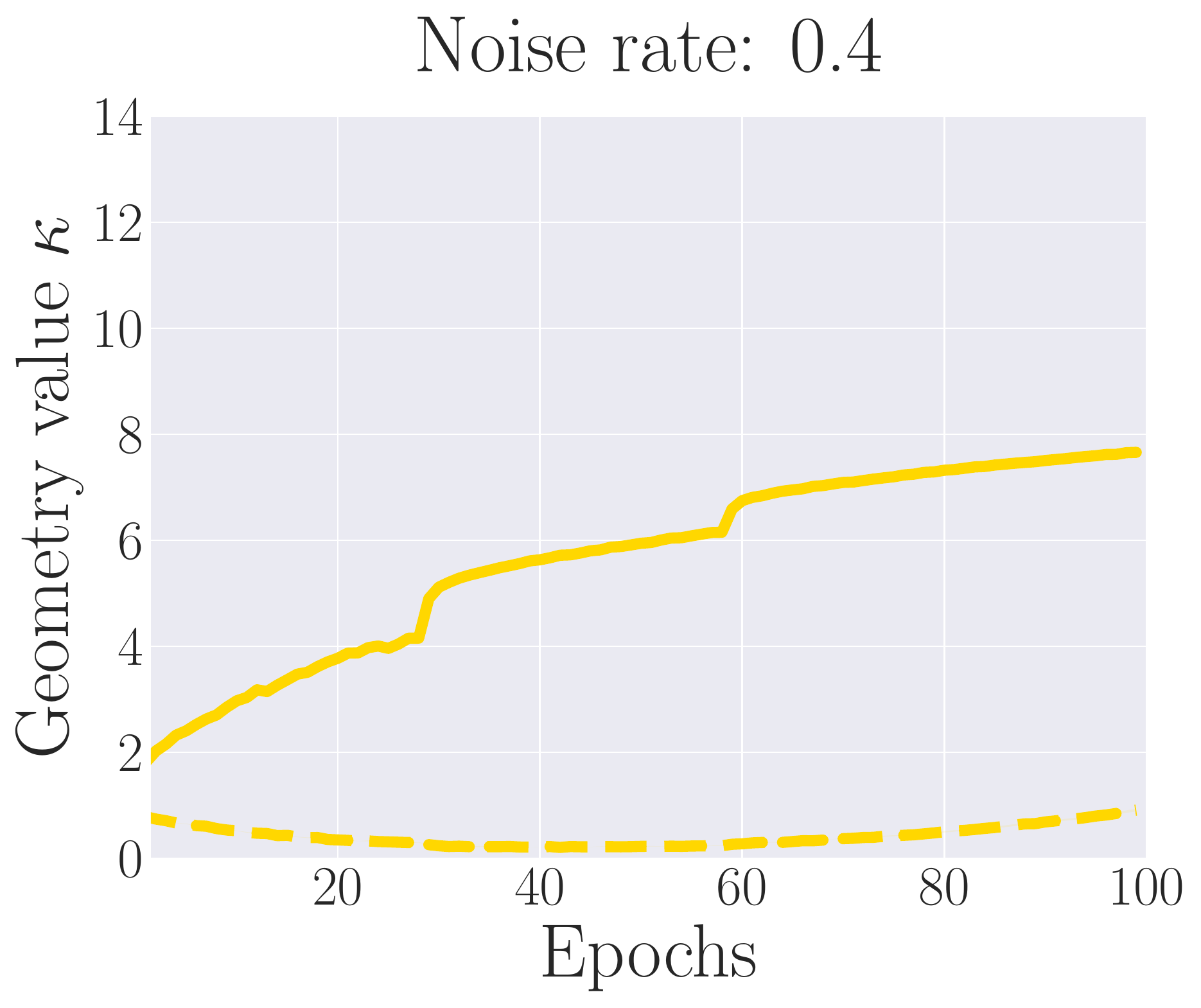}
    \vspace{-2mm}
    \caption{Comparisons of correct/incorrect data in terms of the loss value (top panel) and the geometry value $\kappa$ (bottom panel) on \textit{CIFAR-10} with symmetric-flipping noise in AT. We calculate the mean values in each epoch. We clearly demonstrate that the value $\kappa$ has a similar trend as loss value in AT; both can be used for differentiating correct/incorrect data in AT.}
    \vspace{-5mm}
    \label{fig:loss_value_pgd_steps_AT}
\end{figure}

\section{New Measure: Geometry Value $\kappa$}
\label{sec:geo_kappa}

In this section, we show the geometry value $\kappa$ could be a new measure for the data stratification. First, the geometry value $\kappa$ can differentiate correct/incorrect data in AT (Figures~\ref{fig:loss_value_pgd_steps_AT},~\ref{fig:dis_pgd_loss_sym_02}  and~\ref{fig:dis_pgd_loss_asym_04}). Compared with the loss value, which has been wildly used in sample selection~\citep{jiang2018mentornet,han2018co,yu2019does}, we show that the geometry value $\kappa$ can have a better performance to filter incorrect data with different noise types. Second, we demonstrate that the geometry value $\kappa$ can provide a finer stratification on typical/rare data (Figures~\ref{fig:resnet18_AT_30_PGD_loss} and \ref{fig:rare_classic_pgd_steps}).

%Specifically, we utilize the geometry value $\kappa$ in~\citep{zhang2020geometryaware} to realize the measurement. By carefully check the data with different geometry value $\kappa$, we find that the new measurement can stratify the incorrect data, rare (correct) data and classic (correct) data. Through further investigation on the value $\kappa$, we comprehensively analyze the similarities and differences between it and the loss value during the training.

%(Mainpoint: $\kappa$ can help distinguish incorrect and correct data. In some case, it is better than the loss value.)

%(First, further explore the relationship of the value $\kappa$ with loss value. we find the similarity between the loss value and $\kappa$)
\subsection{Geometry Value vs. Loss Value}
\label{sec:incorrect_correct}

\begin{figure}[h!]
%\vspace{2mm}
    \centering
    \includegraphics[scale=0.19]{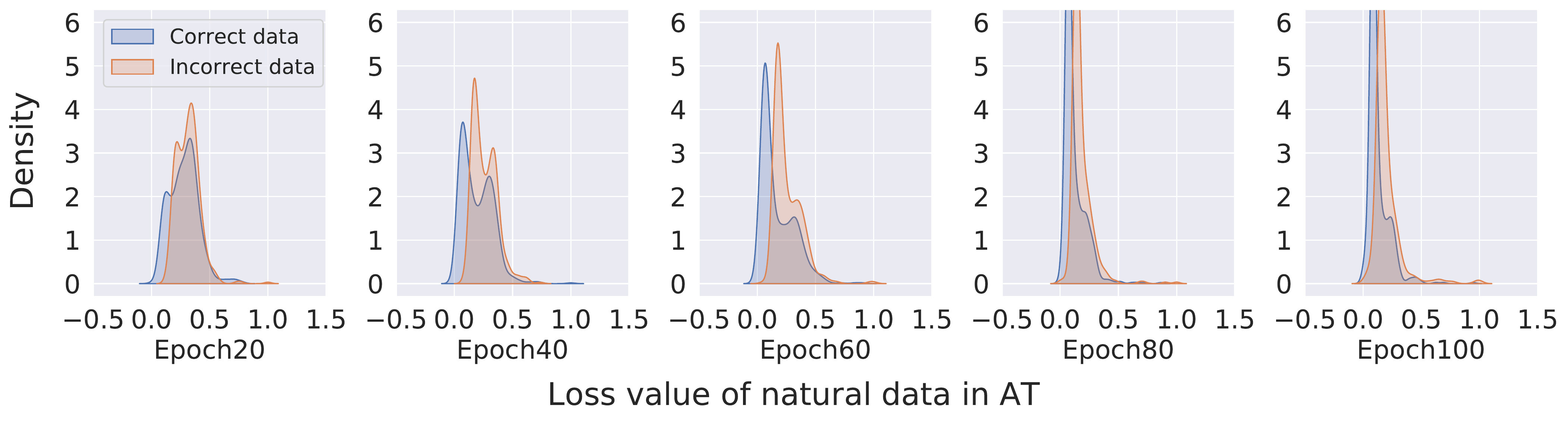}\\
    \includegraphics[scale=0.19]{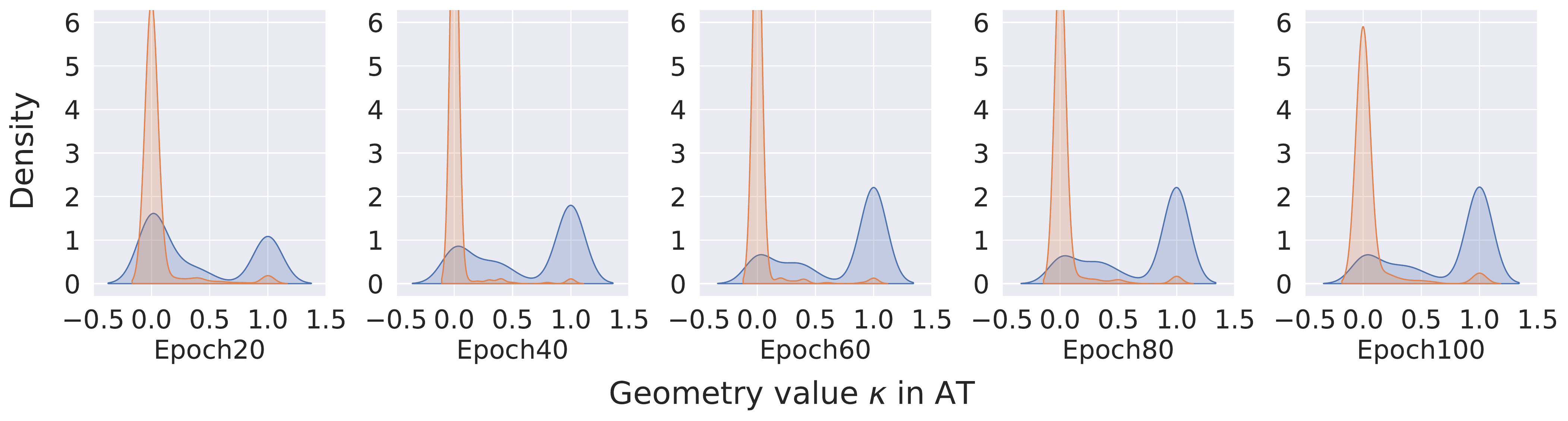}\\
    \vspace{-3mm}
    \caption{The density of AT on correct/incorrect data using \textit{CIFAR-10} with $40\%$ pair-flipping noise. \textit{Top panels}: the loss value in AT. \textit{Bottom panels}: the geometry value $\kappa$ in AT. Note that the geometry value $\kappa$ has a better distinction on correct/incorrect data.}
    \vspace{-1mm}
    \label{fig:dis_pgd_loss_asym_04}
\end{figure}

To combat noisy labels, sample selection methods are very effective. As a common measure in sample selection, the loss value is used to filter incorrect data. For example, small-loss data can be regarded as ``correct'' data. However, there are two limitations in using the loss value as a measure. First, we need to adjust different thresholds to obtain a better selection effect, when the dataset has different noise rates and types~\citep{yao2020searching}. Second, for pair-flipping noise, the loss value cannot distinguish correct/incorrect data well (the top panel of Figure~\ref{fig:dis_pgd_loss_asym_04}).

\begin{figure}[h!]
\vspace{2mm}
    \centering
    \includegraphics[scale=0.19]{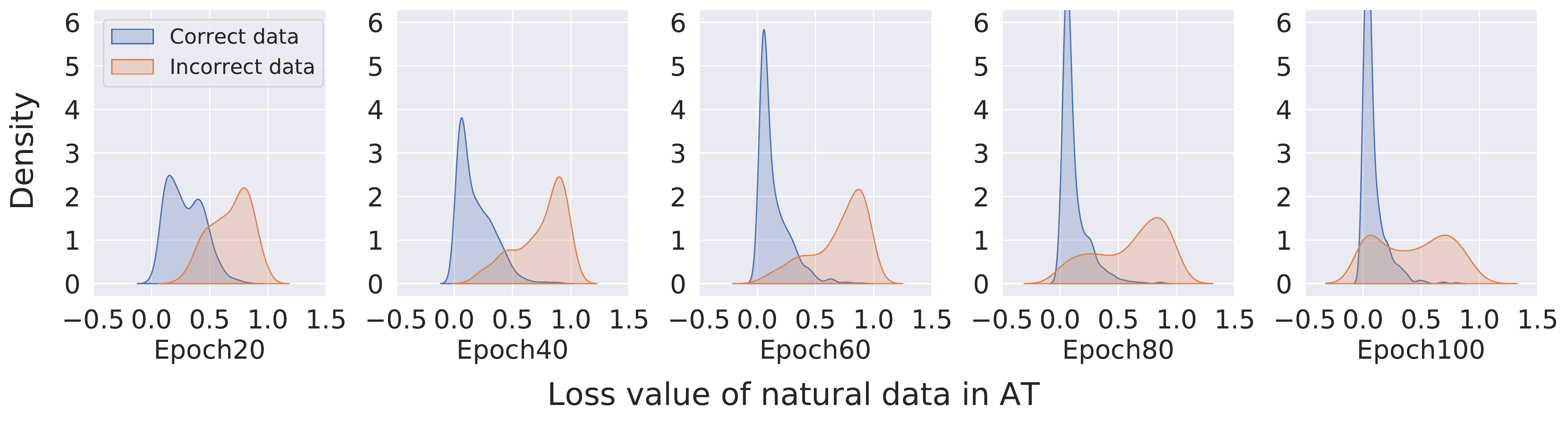}\\
    \includegraphics[scale=0.19]{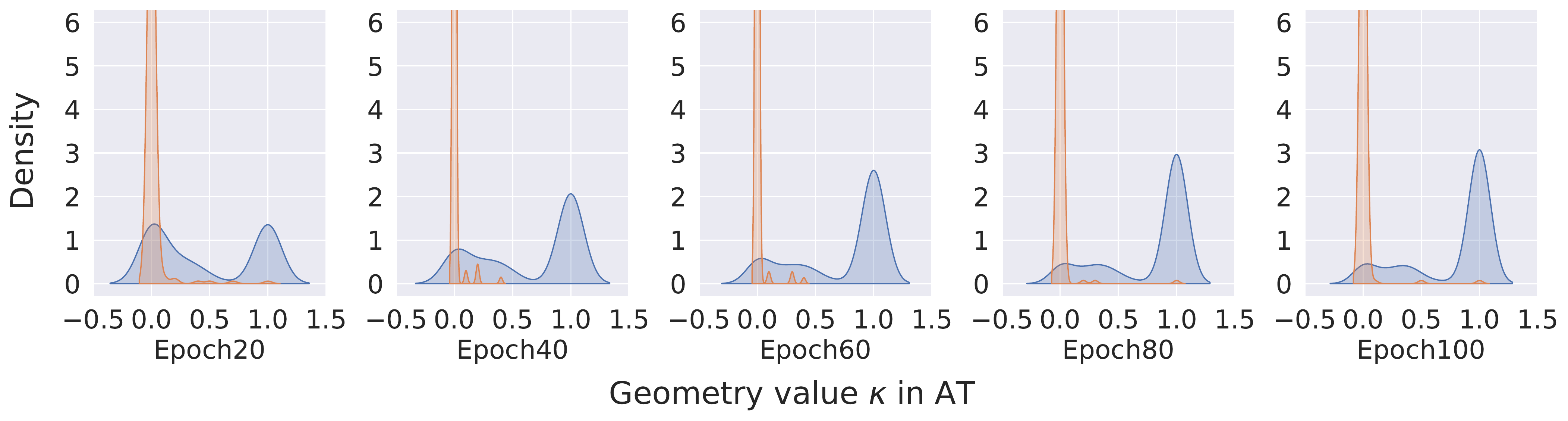}\\
    \vspace{-3mm}
    \caption{The density of AT on correct/incorrect data using \textit{CIFAR-10} with $20\%$ symmetric-flipping noise. \textit{Top panels}: the loss value in AT. \textit{Bottom panels}: the geometry value $\kappa$ in AT. Note that the geometry value $\kappa$ has a better distinction on correct/incorrect data.}
    %\vspace{-5mm}
    \label{fig:dis_pgd_loss_sym_02}
\end{figure}

In Figure~\ref{fig:loss_value_pgd_steps_AT}, we compare the geometry value $\kappa$ and loss value of correct/incorrect data in the training process. We find that the value $\kappa$ can be used to differentiate incorrect data from correct data, since it has a similar trend as the loss value in AT. To further compare two measures in distinguishing correct/incorrect data, we plot the density maps of two measures on the \textit{CIFAR-10} dataset with different noise types in Figures~\ref{fig:dis_pgd_loss_sym_02} and~\ref{fig:dis_pgd_loss_asym_04}. To compare the two measures in a meaningful way, we perform the min-max normalization~\citep{tax2000feature} on both the loss value and geometric value $\kappa$, which scales the range of values in $[0,1]$.

For symmetric-flipping noise (Figure~\ref{fig:dis_pgd_loss_sym_02}), although the loss value can distinguish correct data from incorrect data during the training process, the geometric value $\kappa$ has a better distinction between correct and incorrect data. Specifically, the top panels of Figure~\ref{fig:dis_pgd_loss_sym_02} show that there are a large number of correct/incorrect data with the same loss value, which requires a carefully designed threshold to select the correct data from incorrect data. In contrast, correct/incorrect data can be well divided using the value $\kappa$ in the bottom panels of Figure~\ref{fig:dis_pgd_loss_sym_02}. We can easily select correct/incorrect data with high purity. More obviously, for pair-flipping noise, the loss value of correct/incorrect data overlaps in the top panels of
Figure~\ref{fig:dis_pgd_loss_asym_04}. However, the value $\kappa$ in the bottom panels of Figure~\ref{fig:dis_pgd_loss_asym_04} still provides a good discrimination on correct/incorrect data.

\begin{figure}[h!]
\vspace{2mm}
    \centering
    \includegraphics[scale=0.37]{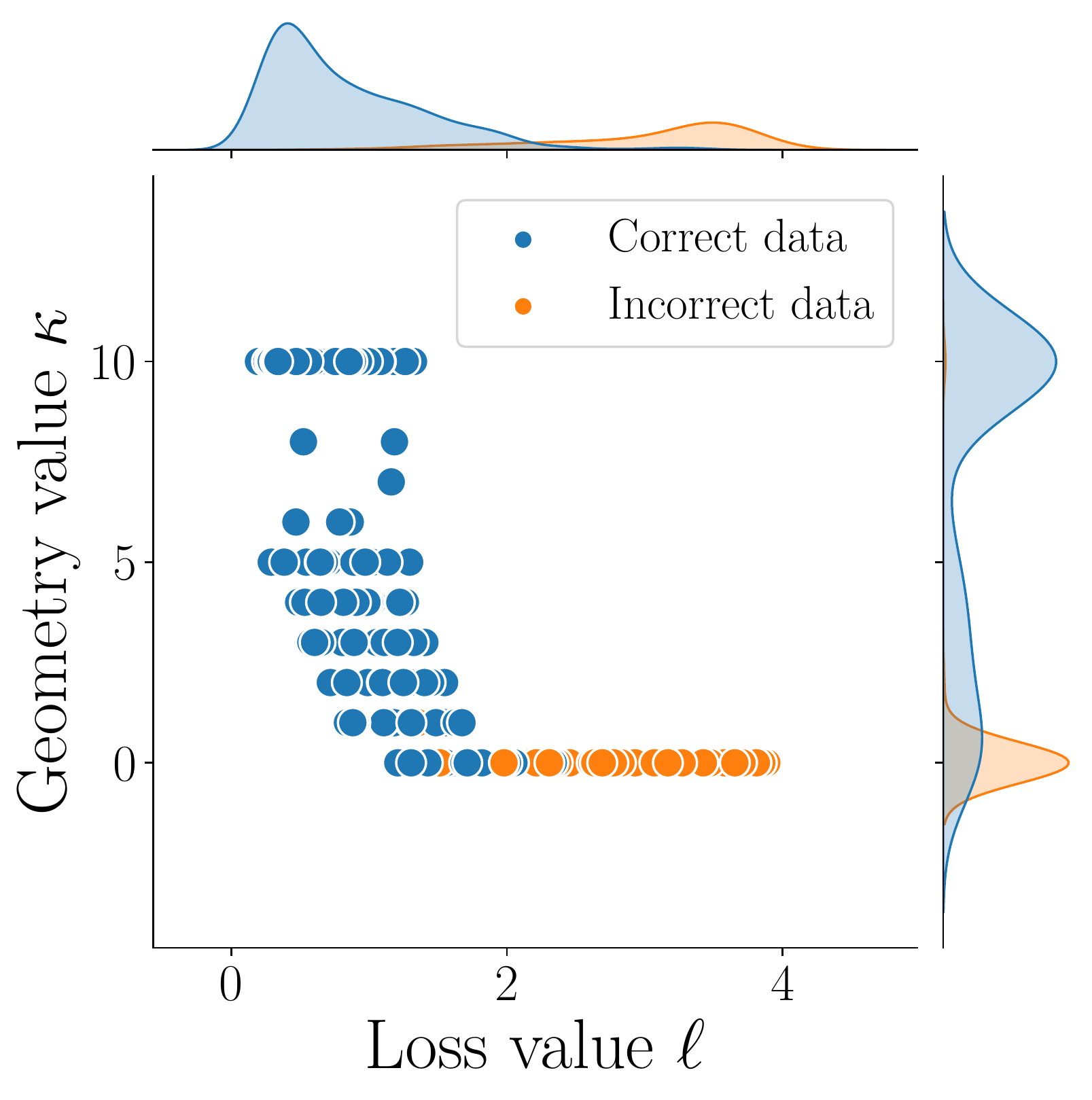}
    \vspace{-3mm}
    \caption{We choose the model trained by AT using \textit{CIFAR-10} with $20\%$ symmetric-flipping noise. We jointly analyze the geometry value $\kappa$ and the loss value, which shows that the value $\kappa$ can provide a fine stratification on typical (i.e., larger $\kappa$)/rare (i.e., smaller $\kappa$) data.}
    \vspace{0mm}
    \label{fig:resnet18_AT_30_PGD_loss}
\end{figure}

\begin{figure*}[h!]
%\vspace{2mm}
    \centering
    \hspace{-2mm}
    \subfigure[CIFAR-10]{
    \begin{minipage}[b]{0.48\linewidth}
        \includegraphics[scale=0.43]{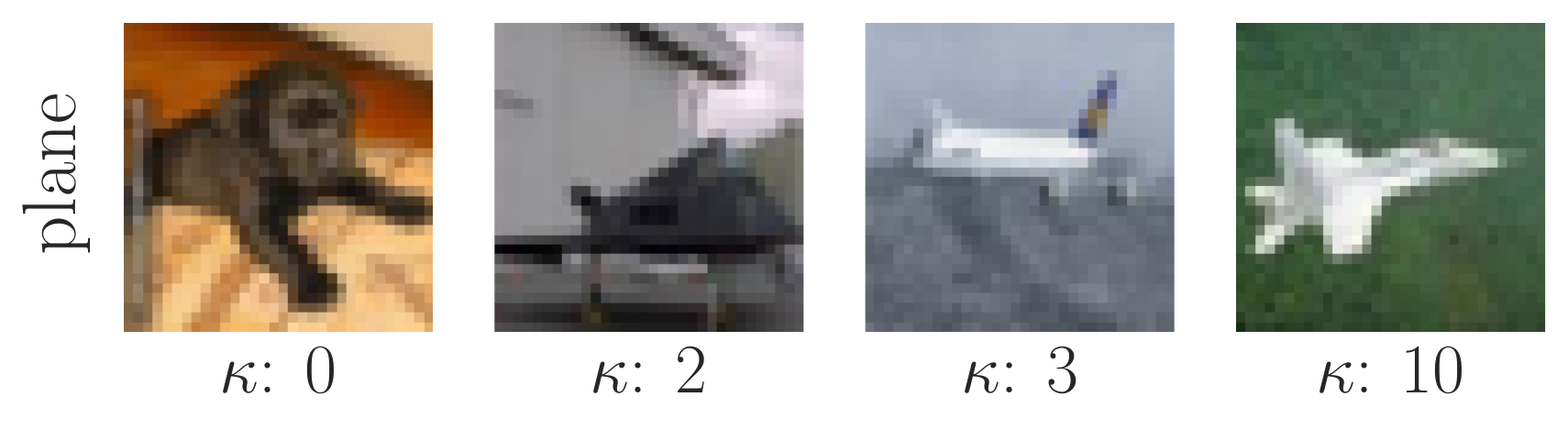}\\
        \includegraphics[scale=0.43]{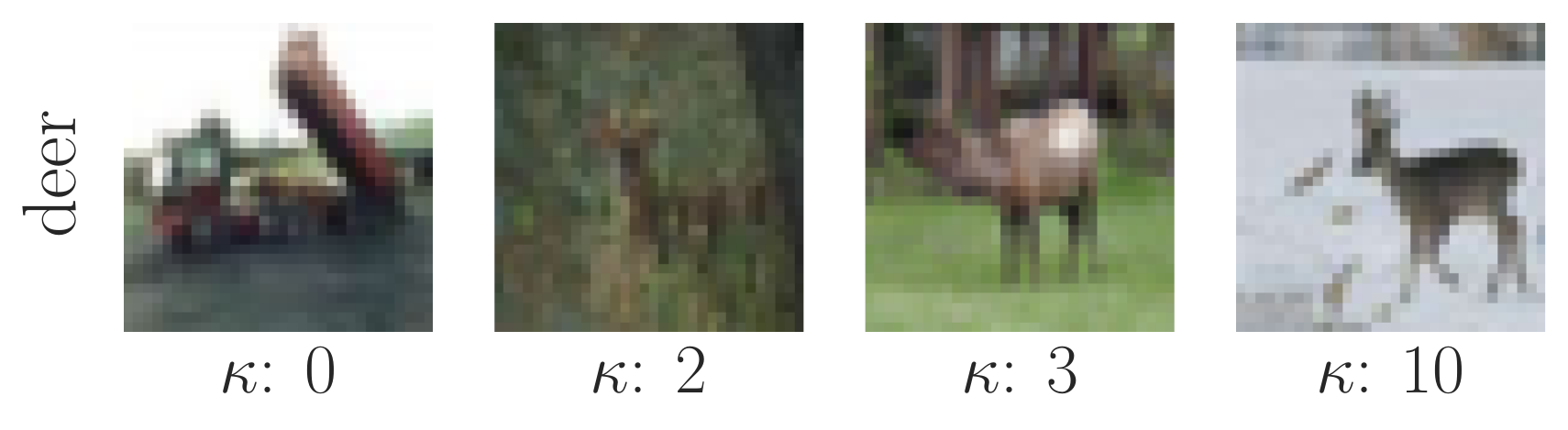}\\
        \label{fig:cifar10_rare_pgd_steps}
        \vspace{-5mm}
    \end{minipage}
    }
    %\hspace{2mm}
    \subfigure[MNIST]{
    \begin{minipage}[b]{0.44\linewidth}
        \includegraphics[scale=0.43]{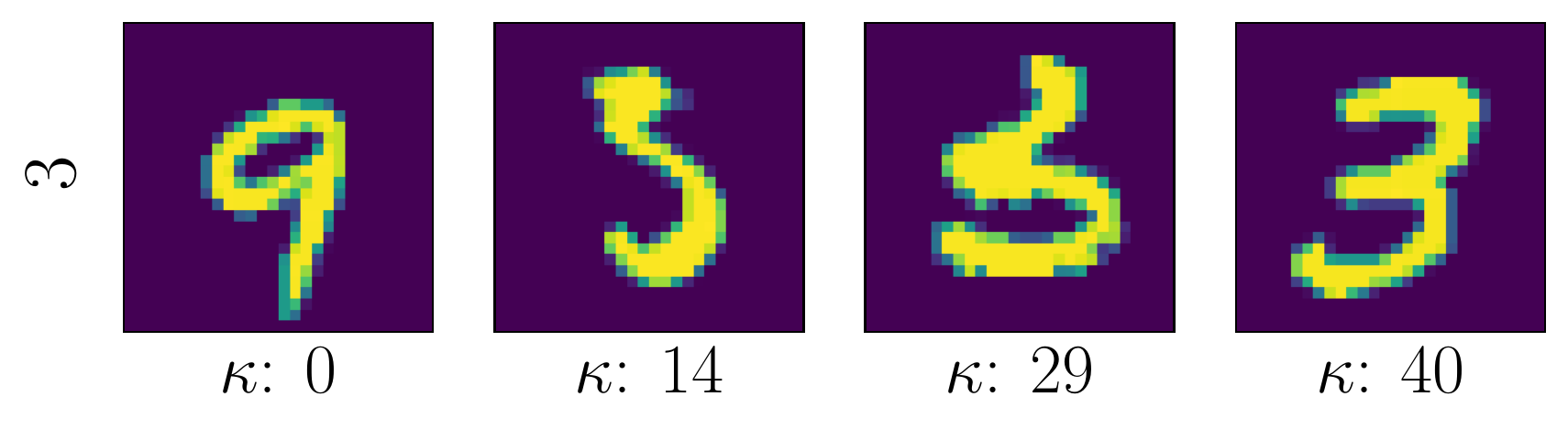}\\
        \includegraphics[scale=0.43]{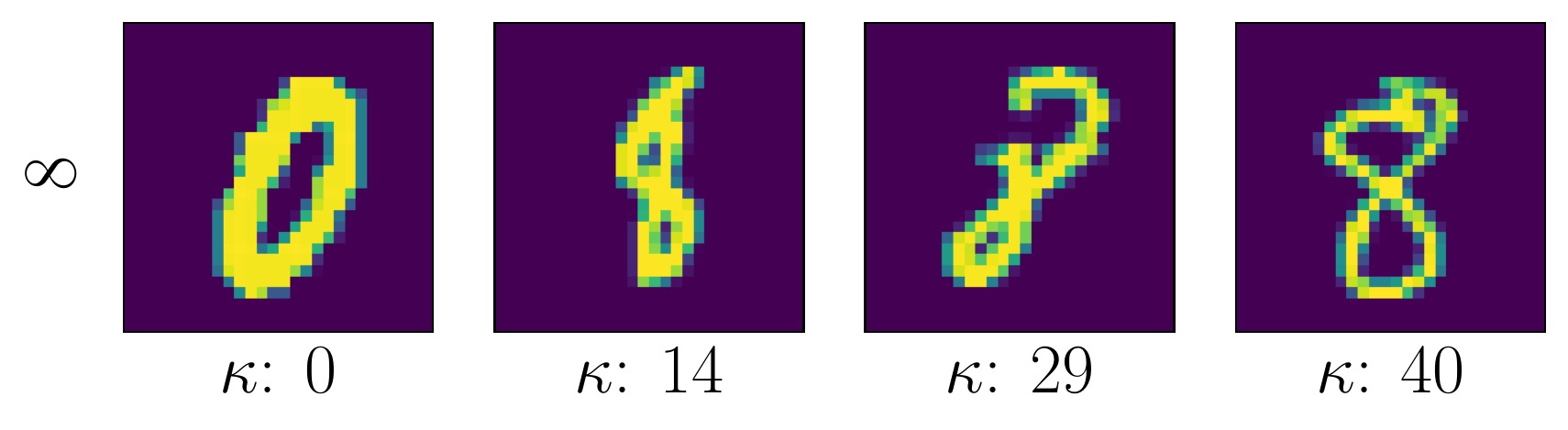}\\
        \label{fig:mnist_rare_pgd_steps}
        \vspace{-5mm}
    \end{minipage}
    }
    \vspace{-3mm}
    \caption{The geometry value $\kappa$ w.r.t. images in \textit{CIFAR-10} and \textit{MNIST} with $20\%$ and $10\%$ symmetric-flipping noise. The leftmost of each subfigure is the given label (i.e., $deer$ and $plane$ or $3$ and $8$) of all images on the right. We randomly select four examples with the different $\kappa$ in each class. As the geometric value $\kappa$ increases from left ($\kappa = 0$) to right ($\kappa = 10$ or $40$), the semantic information of images is more typical and recognizable.}
    \vspace{-4mm}
    \label{fig:rare_classic_pgd_steps}
\end{figure*}

In addition, we find that the geometry value $\kappa$ can provide a fine stratification on typical/rare data. First, we jointly analyze the value $\kappa$ and the loss value in AT (Figure~\ref{fig:resnet18_AT_30_PGD_loss}), where we stratify correct data via $\kappa$. Secondly, by inspecting the semantic information with different $\kappa$, we find that the value $\kappa$ can represent whether the data is relatively typical or rare (Figures~\ref{fig:cifar10_rare_pgd_steps} and~\ref{fig:mnist_rare_pgd_steps}). Moreover, we plot a bivariate graph of the loss value and the value $\kappa$ in Figure~\ref{fig:resnet18_AT_30_PGD_loss}. In this figure, we mainly focus on the correctly classified data (blue scattered dots), since the wrongly classified data (orange scattered dots) has been clearly discriminated by big loss values. Note that, for small-loss (correct) data, the value $\kappa$ can further subdivide such data into typical and rare types.

\subsection{Distinguishable Typical/Rare Data}
\label{sec:rare_classic}

From the \textit{macro} perspective, the loss value can be regarded as a measure to classify correct and incorrect data~\citep{jiang2018mentornet}. Namely, small-loss data can be regarded as correct data, and vice versa. However, such stratification is a bit rough, which motivates us to seek a 
\textit{micro} measure called the geometry value $\kappa$ (the number of PGD steps) in AT. To justify our findings in Figure~\ref{fig:resnet18_AT_30_PGD_loss}, we visualize the semantic information of \textit{CIFAR-10} (Figure~\ref{fig:cifar10_rare_pgd_steps}) and \textit{MNIST} (Figure~\ref{fig:mnist_rare_pgd_steps}) under different $\kappa$. We find that images with large $\kappa$ (rightmost) are prototypical and easier to recognize from the viewpoint of human perception, while images with small $\kappa$ (or $\kappa=0$) seem to be more rare (or incorrect). These rare images have the atypical semantic information, such as some strange shapes (``8" with $\kappa=14$ in Figure~\ref{fig:mnist_rare_pgd_steps}) or confusing backgrounds (``deer" with  $\kappa=2$ in Figure~\ref{fig:cifar10_rare_pgd_steps}). More results about the images with different $\kappa$ can be found in Appendix~\ref{appendix:new_measurement}.

\section{Applications of Geometry Value $\kappa$}
\label{sec:app_kappa}

\begin{figure}[h!]
\vspace{1mm}
    \centering        
    \includegraphics[scale=0.2]{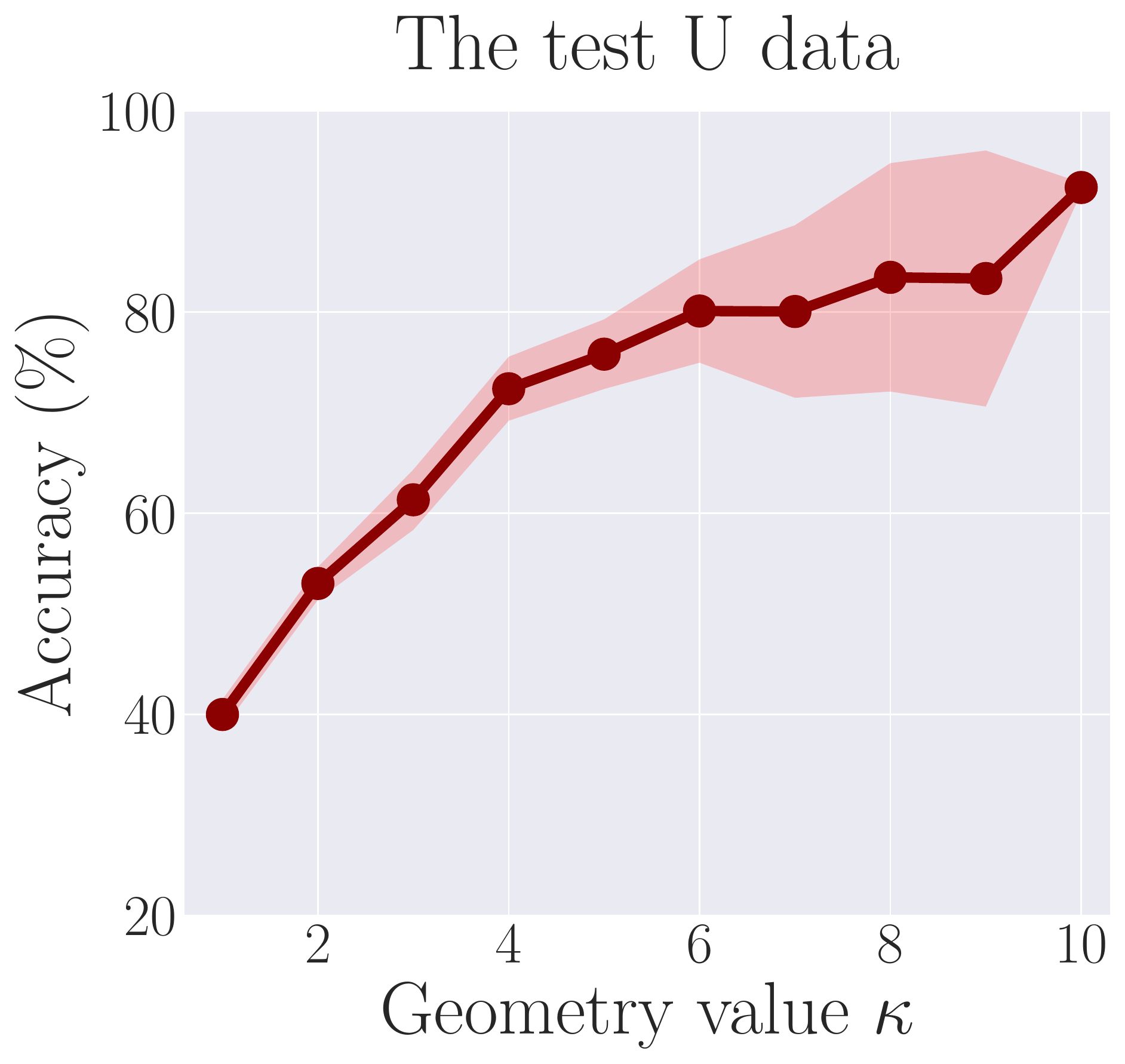}
    \includegraphics[scale=0.2]{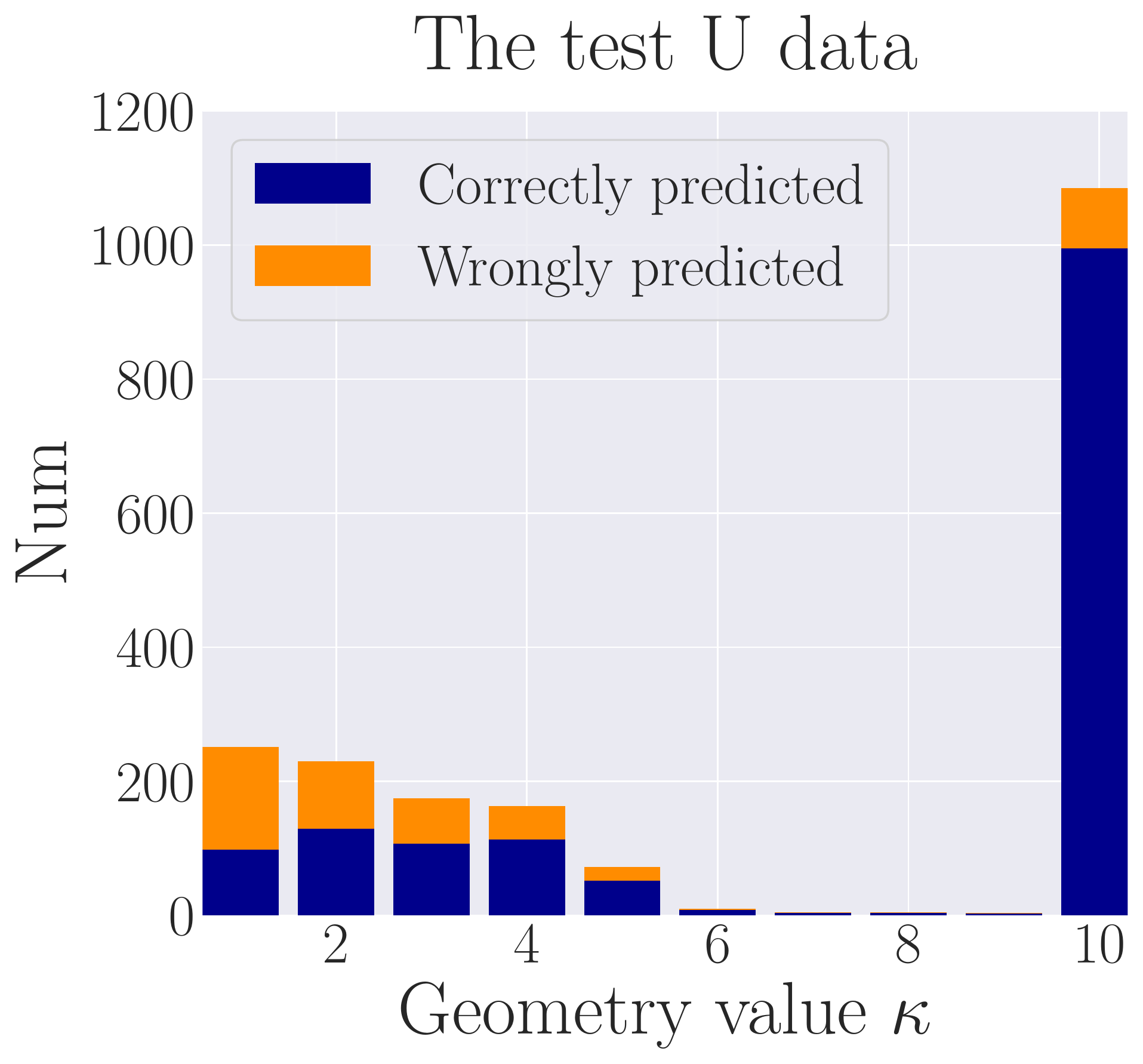}
    %\vspace{-1mm}
    \caption{The accuracy (left panel) and number (right panel) of correctly predicted U data w.r.t. the geometry value $\kappa$. We randomly select $2000$ test data in \textit{CIFAR-10} as unlabeled data. The larger $\kappa$ corresponds to the higher prediction accuracy.}
    \vspace{0mm}
    \label{fig:resnet18_confidence}
\end{figure}

In this section, we provide two applications of our new measure---the geometry value $\kappa$ (the number of PGD steps). Since the value $\kappa$ can differentiate correct/incorrect data in AT (Section~\ref{sec:incorrect_correct}), in the presence of label noise, we can use it to detect noisy labels and correct labels (Figure~\ref{fig:annotator_effect}). Meanwhile, as it can have a fine stratification for typical/rare data (Section~\ref{sec:rare_classic}), we can provide the confidence of annotated labels according to the value $\kappa$ (Figure~\ref{fig:resnet18_confidence}). 

Regardless of ST or AT, high-quality training data are always essential for acquiring a good model~\citep{deng2009imagenet,zhang2019bottleneck}, but the labeling process of high-quality data requires a lot of human resources. To deal with such a problem, many methods used ST to facilitate a \textit{standard annotator} to annotate large-scale unlabeled (U) data~\citep{carmon2019unlabeled,alayrac2019labels}. However, this standard annotator fails when U data are adversarially manipulated.

In practice, label-noise issues widely exist in real-world training datasets, and learning with noisy labels seems inevitable. Meanwhile, the existence of adversarial examples~\citep{szegedy,Goodfellow14_Adversarial_examples} also poses a threat to the annotation for U data. Therefore, we design a robust annotator algorithm (Algorithm~\ref{alg:RA}) to assign reliable labels for U data even in the presence of adversarial manipulations and noisy training labels (Section~\ref{sec:app_robust_annotator}). Compared to human beings, the standard annotator cannot give the information whether the label assignment for U data is reliable. Nonetheless, our new measure could be an alternative to the predictive probability for providing the confidence of annotated labels (Section~\ref{sec:confidence_score}). The detailed experimental setups can be found in Appendix~\ref{sec:appendix_app_kappa}.
\vspace{-2mm}
\subsection{Robust Annotator}
\vspace{-1mm}
\label{sec:app_robust_annotator}
We can construct a robust annotator to assign labels for U data. Here, we consider a real-world scenario, namely, existence of label noise in training data and adversarial manipulations in U data. Our robust annotator has a better labeling performance than the standard annotator, since we use the value $\kappa$ and the loss value jointly to select incorrect training data. We re-annotate high-quality pseudo labels for these incorrect data, and adversarially train on the whole data. Then, our robust annotator can reliably assign labels.

\begin{figure}[h!]
\vspace{1mm}
    \centering
    \includegraphics[scale=0.20]{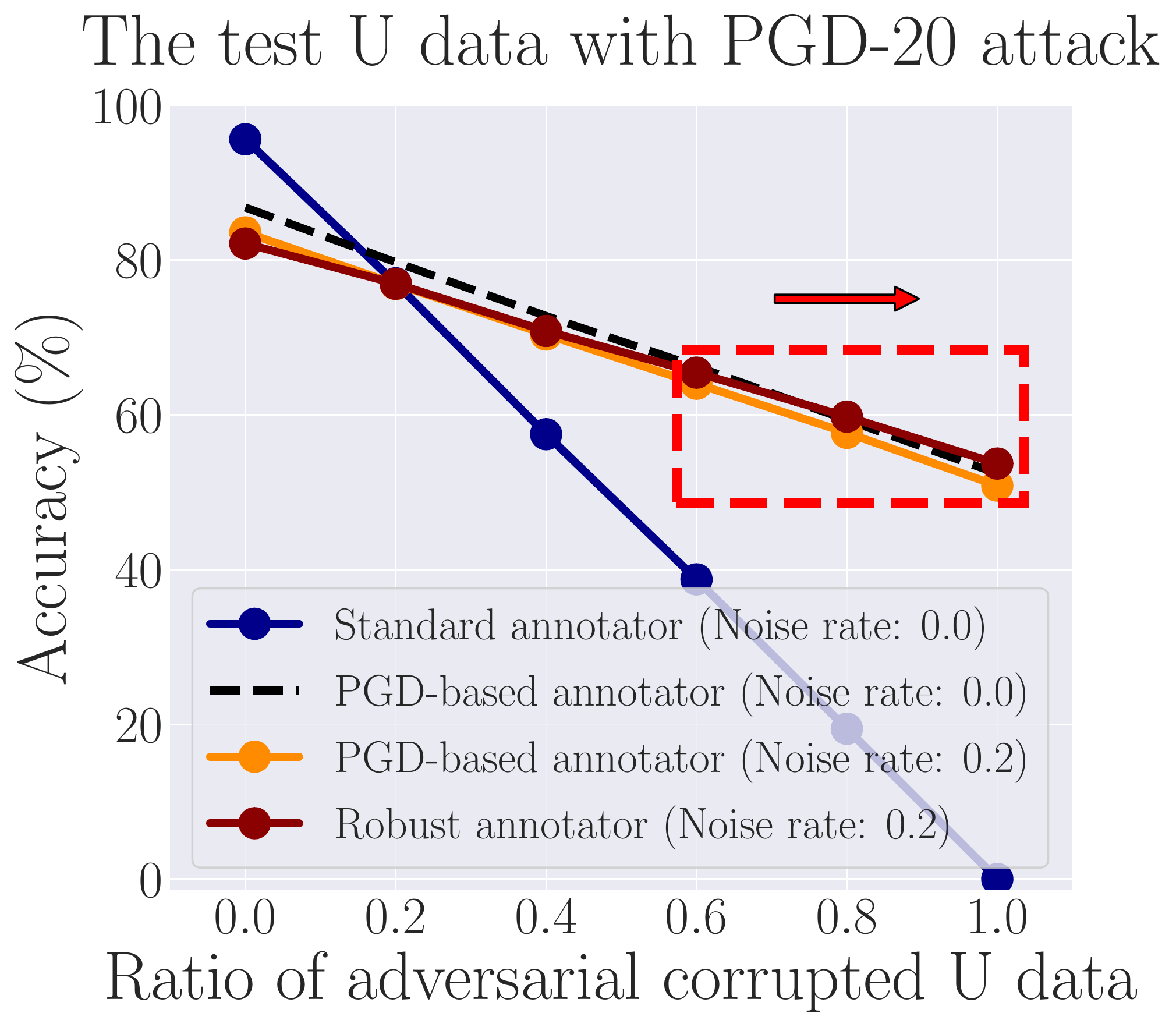}
    \includegraphics[scale=0.20]{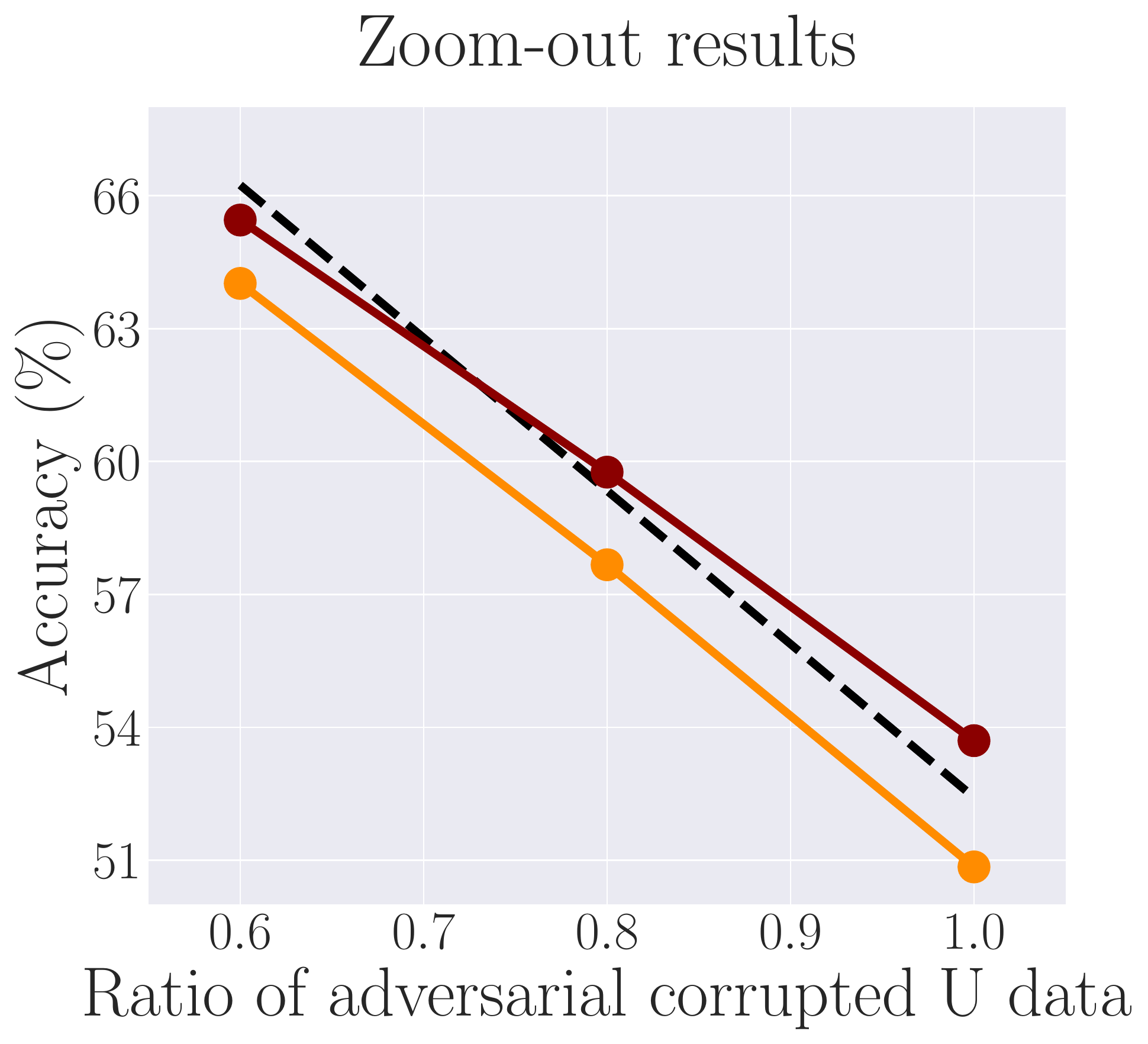}
    %\vspace{1mm}
    \caption{The accuracy of four approaches assigning correct labels to adversarial U data from \textit{CIFAR-10}. \textit{Left panel:} the full results. \textit{Right panel:} the zoom-out results (without standard annotator). Our robust annotator has a satisfactory performance on assigning reliable labels.}
    \vspace{-2mm}
    \label{fig:annotator_effect}
\end{figure}

%\vspace{0mm}

%\vspace{2mm}
\begin{algorithm}[t!]
   \caption{Robust Annotator Algorithm.}
   \label{alg:RA}
   \SetKwInOut{Input}{Input}\SetKwInOut{Output}{Output}

  \Input{network $f_{\mathbf{\theta}}$, training dataset $S = \{(\bx_i, y_i) \}^{n}_{i=1}$, learning rate $\eta$, number of epochs $T$, batch size $m$, number of batches $M$, threshold for geometry value $K$, threshold for loss value $L$.}
  
  \Output{robust annotator $f_{\mathbf{\theta}}$.}
  \For{\rm{epoch} $= 1$, $\dots$, $T$}{
    
    \For{\rm{mini-batch} $=1$, $\dots$, $M$}{
    \textbf{Sample:} a mini-batch $\{(\bx_i, y_i) \}^{m}_{i=1}$ from $S$.
    
    \For{i = 1,\ldots,m (\rm{in parallel})}{
     \textbf{Calculate:} $\kappa_i$ and $\ell_i$ of ($x_i$,$y_i$).
    
    \If{$\kappa_i < K$ and $\ell_i > L$}{
        \textbf{Update:} $y_i \gets \arg\max_{i} f_{\theta} ({ {x} })$.
    }
    \textbf{Generate:} adversarial data $\bxtidle_i$ by PGD method.
    
    }
    
    \textbf{Update:} $\mathbf{\theta} \gets \mathbf{\theta} - \eta \nabla_{\mathbf{\theta}}  \{ \ell(f_{\mathbf{\theta}}(\bxtidle_i), y_i)\}.$
    }
  }
%\vspace{-3mm}
\end{algorithm}
%\vspace{-3mm}
%Adding a certain amount of incorrect data to the adversarial training will improve the robustness of the model to a certain extent, but this is not the main focus of this paper, we will leave it to future work.

In Figure~\ref{fig:annotator_effect}, we test the accuracy of assigning correct labels to U data in the presence of adversarial manipulations. We compare four methods, namely, our robust annotator with $20\%$ symmetric-flipping noise (red line), the PGD-based annotator with $20\%$ symmetric-flipping noise (orange line), the PGD-based annotator without noise (oracle, black dashed line), and the standard annotator without noise (blue line). On normal U data (i.e., zero adversarial ratio), the standard annotator has better performance of labeling. However, when U data is subjected to certain adversarial manipulations (i.e., ratio above $0.2$), the labeling quality of the standard annotator decreases sharply, but that of our robust annotator still remains satisfactory. An extreme case is that, when all U data (ratio $1.0$) are added to adversarial manipulations, labels assigned by the standard annotator become completely unreliable, but our labels assigned by the robust annotator are still better than the PGD-based annotator with $20\%$ symmetric-flipping noise.

\subsection{Confidence Scores}
\label{sec:confidence_score}

For a given data point, the geometry value $\kappa$ can provide a confidence score, which represents the reliability of label annotations. The measure value $\kappa$ can distinguish between typical data (correctly labeled with high probability) and rare data (wrongly labeled with high probability) in U data. In the left panel of Figure~\ref{fig:resnet18_confidence}, we plot the accuracy of correctly predicted data with the value $\kappa$. The larger $\kappa$ corresponds to higher prediction accuracy, which shows that the value $\kappa$ can indeed represent the reliability of label annotations. In the right panel of Figure~\ref{fig:resnet18_confidence}, we further investigate the number of correctly predicted data with the value $\kappa$. Most of the data have the value $\kappa=10$, which corresponds to a high prediction accuracy. Meanwhile, a small part of the data have the value $\kappa \in [0,6]$, which corresponds to a low prediction accuracy. Since the number of data with value $\kappa \in [7,9]$ is small, the standard deviation of the accuracy is large.

%Motivated by the previous observation, we carefully design a method to construct a useful tool , namely Robust Annotator (RA), for robustly assigning correct labels to the unlabeled data, even if the unlabeled data may be adversarially manipulated.

\section{Conclusion}
\label{sec:conclusion}

In this paper, we explored the interaction of adversarial training with noisy labels. We took a closer look at smoothing effects of adversarial training (AT), and further investigated positive knock-on effects of AT. As a result, AT can distinguish correct/incorrect data and alleviate memorization effects in deep networks. Since smoothing effects can make incorrect data be non-robust, the geometry value $\kappa$ (i.e., the number of PGD steps) could be a new measure to differentiate correct/incorrect and typical/rare data. Moreover, we gave two applications of our new measure, i.e., robust annotator and confidence scores. With the robust annotator, we can assign reliable labels for adversarial U data. With confidence scores, we can know the reliability of label annotations. In future, we hope to further explore the direction of combining adversarial training with noisy labels. For example, as the semantic information of some correct data (e.g., $\kappa = 2$ in Figure~\ref{fig:cifar10_rare_pgd_steps}) are ambiguous, we might further investigate the impact of these data for adversarial training.

\section*{Acknowledgements}

BH was supported by the RGC Early Career Scheme No. 22200720 and NSFC Young Scientists Fund No. 62006202. TLL was supported by Australian Research Council Project DE-190101473. MK was supported by the National Research Foundation, Singapore under its Strategic Capability Research Centres Funding Initiative. JZ, GN, and MS were supported by JST AIP Acceleration Research Grant Number JPMJCR20U3, Japan. MS was also supported by the Institute for AI and Beyond, UTokyo.

\clearpage


\nocite{langley00}

\bibliography{main}
\bibliographystyle{plainnat}


\appendix
\onecolumn

\section{The Smoothing Effects of Aadversarial Training}
\label{appendix:smoothing_effect}

\begin{figure}[hp!]
\vspace{0mm}
    \centering
    \includegraphics[scale=0.3]{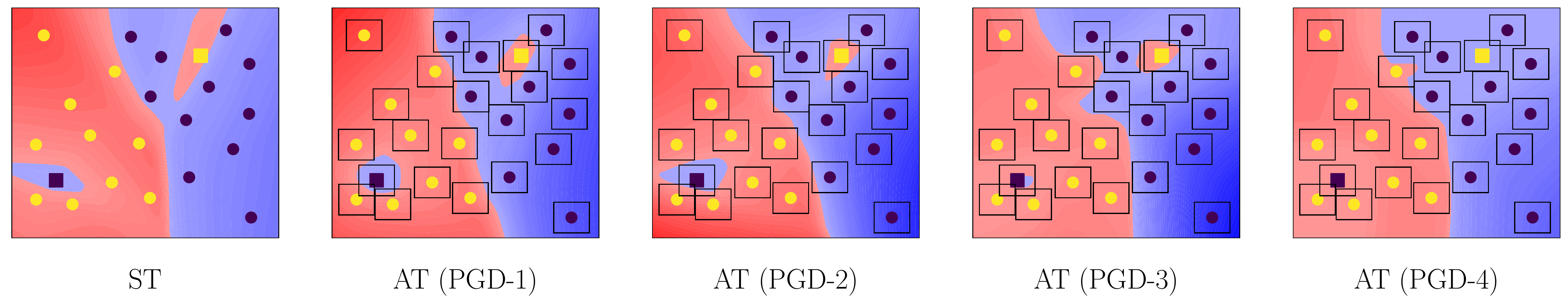}
    \vspace{-2mm}
    \caption{The results of standard training (ST) and adversarial training (AT) on a binary dataset with noisy labels. Dots denote correct data, while squares denote incorrect data. The color gradient represents the prediction confidence: the deeper color represents higher prediction confidence.
    In the leftmost panel, deep networks shapes two small clusters (red and blue ones in cross-over areas) around two incorrect data due to memorization effects in ST. As the number of PGD iterations increases, the smoothing effects in AT gradually strengthens, and two small clusters gradually shrink until they disappear in the rightmost panel. Namely, these clusters have been smoothed out in AT (PGD-4). Boxes represent the norm ball of AT.
    }
    \vspace{0mm}
    \label{fig:appendix_illustration}
\end{figure}

In this section, We provide the detailed setup and more results on the synthetic binary dataset and the real-world dataset (\textit{CIFAR-10}) with noisy labels, which demonstrate the smoothing effects of adversarial training (AT).

\paragraph{Experimental setup.}
To construct synthetic binary dataset, we randomly generate $23$ points (i.e., $(a,b)$, where $a \in (0,1)$ and $b \in (0,1)$) with binary labels (i.e., ``0'' and ``1'') on a two-dimensional plane. Among all data , we choose two points to assign incorrect labels. For the binary classification, we build a simple network contains $5$ linear layers and $4$ ReLU~\citep{nair2010rectified} layers. We train the simple network in ST and AT using Adam with the initial learning rate=$0.001$ for $1000$ iterations. In AT, we set the perturbation bound $\epsilon=0.08$ and the PGD step size $\alpha=0.02$.

\paragraph{Result.}
In Figure~\ref{fig:appendix_illustration}, We plot the classification results in the two-dimensional plane for both ST and AT. We use different PGD iterations to generate adversarial examples, which shows the smoothing process dynamically. In ST, deep network will shape two small clusters around two incorrect data due to memorization effects. While in AT, these small clusters will gradually shrink until they disappear, as the smoothing effects in AT strengthens (i.e., from PGD-1 to PGD-4).

\begin{figure*}[h!]
%\vspace{1mm}
    \centering
    \subfigure[Random direction (Figure 2)]{
    \includegraphics[scale=0.30]{nb100_entropy_at_st_sym_noise.pdf}
    \label{fig:random_dir}
    \vspace{-2mm}
    }
    \hspace{4mm}
    \subfigure[Adversarial direction]{
    \includegraphics[scale=0.30]{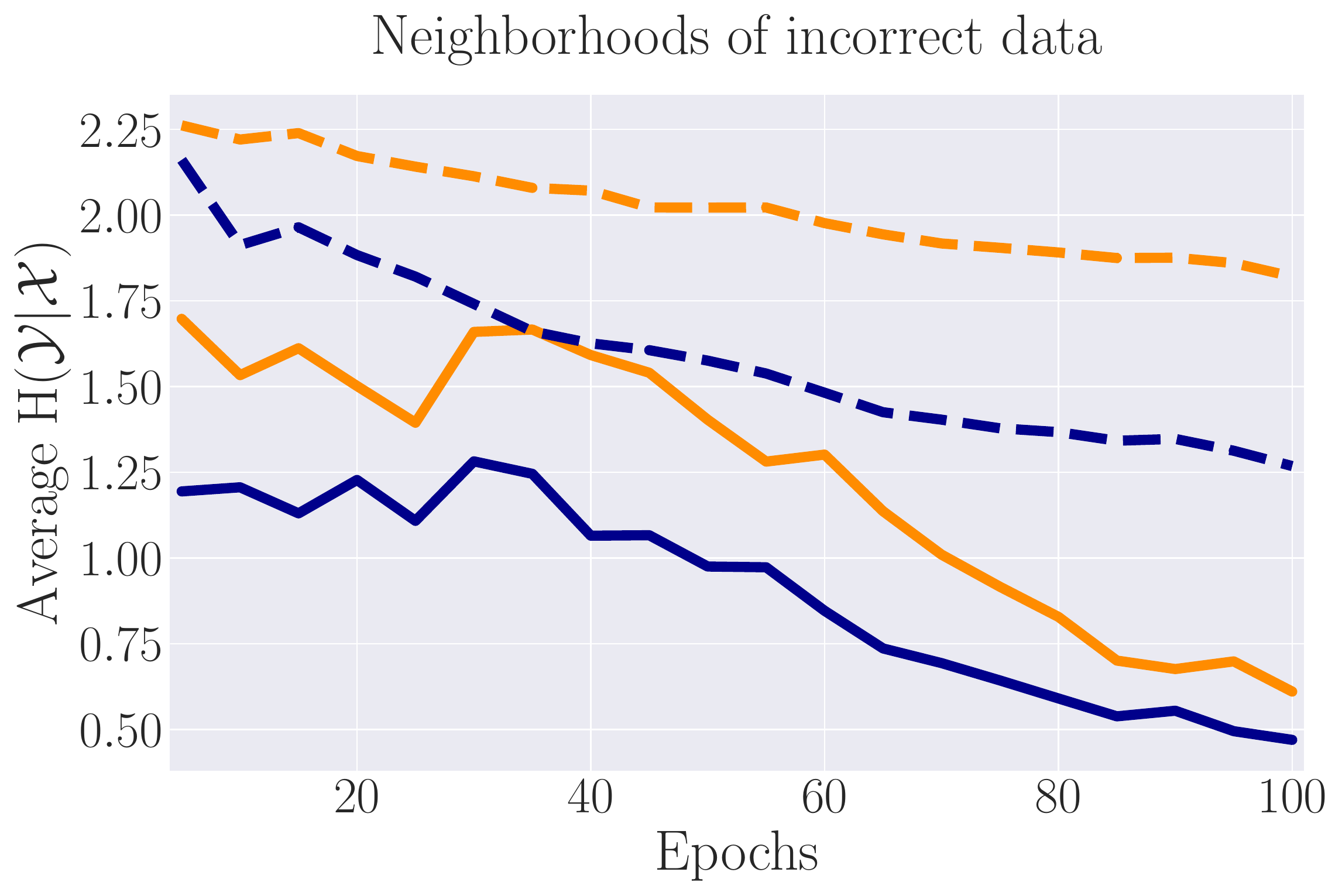}
    \label{fig:adv_dir}
    \vspace{-2mm}
    }
    \caption{The average entropy of models trained by ST and AT. This value is calculated on $100$ points in each neighborhood of incorrect data, using \textit{CIFAR-10} with symmetric-flipping noise. Both \textit{solid} and \textit{dashed} lines represent ST and AT, respectively. Note that ST learns incorrect data more deterministically than AT.}
    \vspace{-2mm}
    \label{fig:appendix_entropy_plot}
\end{figure*}

\paragraph{Result.} 
In Figure~\ref{fig:appendix_entropy_plot}, we plot the entropy values of the model predictions on the \textit{CIFAR-10} dataset. In Figure~\ref{fig:random_dir}, we randomly select $100$ points in each neighborhood (within a small $\epsilon$-ball, where $\epsilon=0.031$) of the incorrect data and calculate their average entropy values in training using the models trained by ST and AT. In Figure~\ref{fig:adv_dir}, we generate the adversarial variant for each incorrect data by PGD-1 attack with $\epsilon=0.031$ and calculate the average entropy values in training. The detailed training settings can be found in Appendix~\ref{sec:app_train_test}. On the whole, compared with ST, the entropy values in AT are always higher. It demonstrates that AT did not learn incorrect data with their neighborhoods deterministically, which confirms that the smoothing effects in AT prevent incorrect data from forming small clusters during training.

\section{Knock-on Effects of Adversarial Training}
\label{appendix:diff_st_at}

In this section, we provide more complementary experiments and analysis for the positive knock-on effects of AT. First, we show the results of standard training and test accuracy on \textit{CIFAR-10} and \textit{MNIST} datasets with different noise rates and types (Appendix~\ref{sec:app_train_test}). Second, we show the analysis of the natural data and adversarial data in AT (Appendix~\ref{sec:app_nat_adv}). Third, we use different networks to investigate positive knock-on effects of AT with noisy labels (Appendix~\ref{sec:app_diff_net}). Finally, we show the results of loss value with different noise rates and types (Appendix~\ref{sec:app_loss_value}).

\subsection{Training Accuracy and Test Accuracy}
\label{sec:app_train_test}

\begin{figure*}[h!]
\vspace{4mm}
    \centering
    \subfigure[CIFAR-10]{
    \includegraphics[scale=0.195]{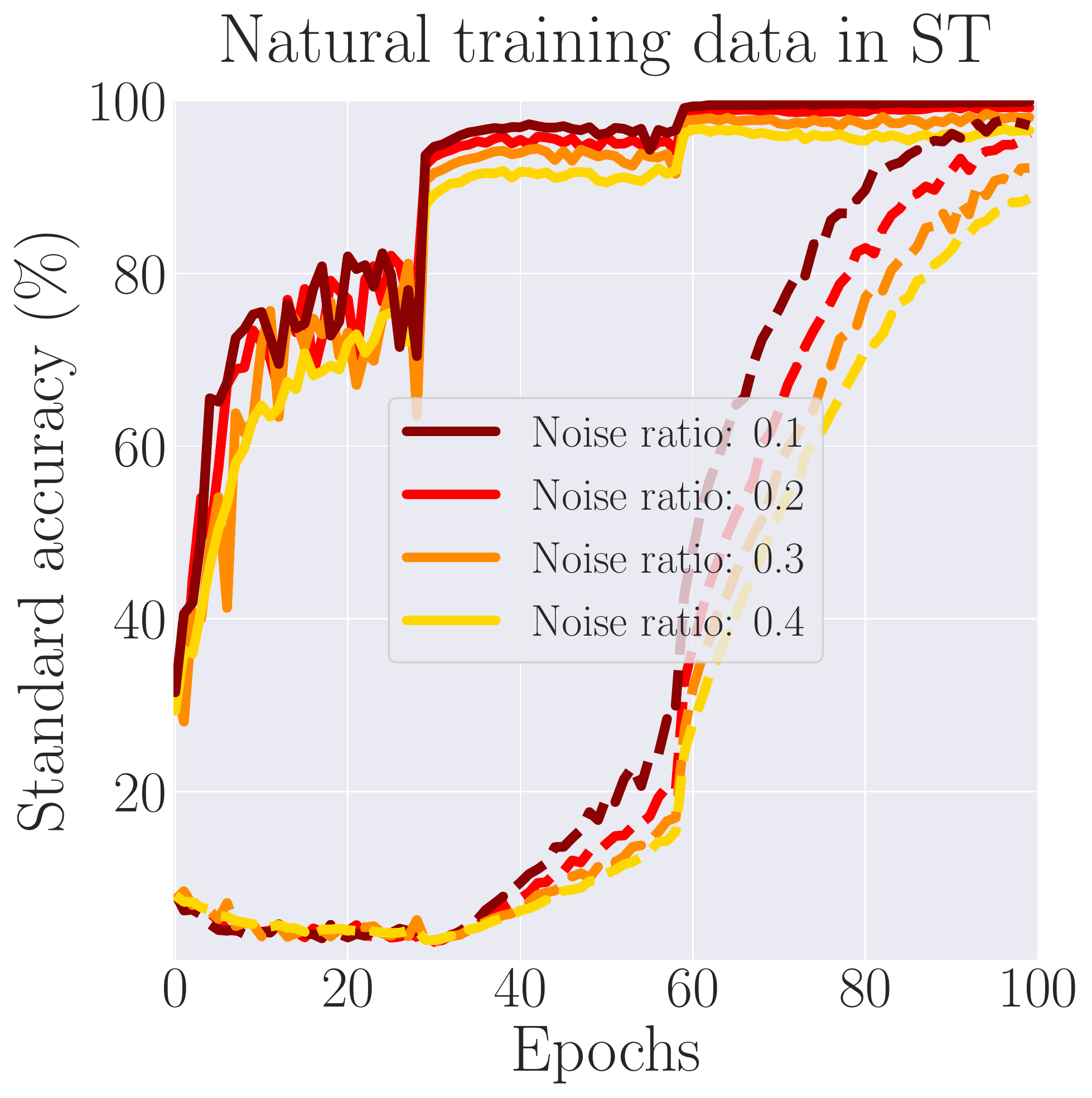}
    \includegraphics[scale=0.195]{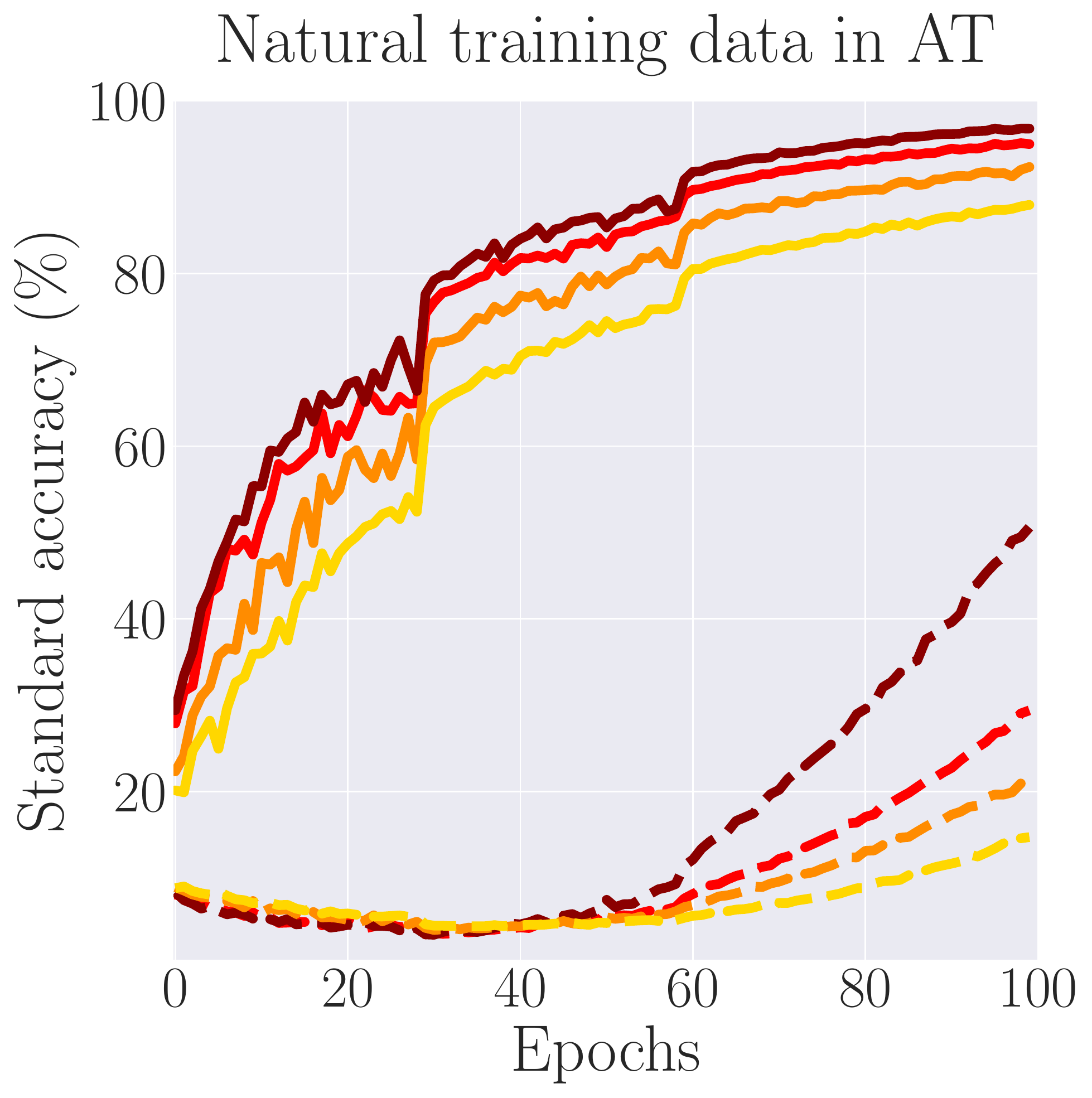}
    }
    \subfigure[MNIST]{
    \includegraphics[scale=0.195]{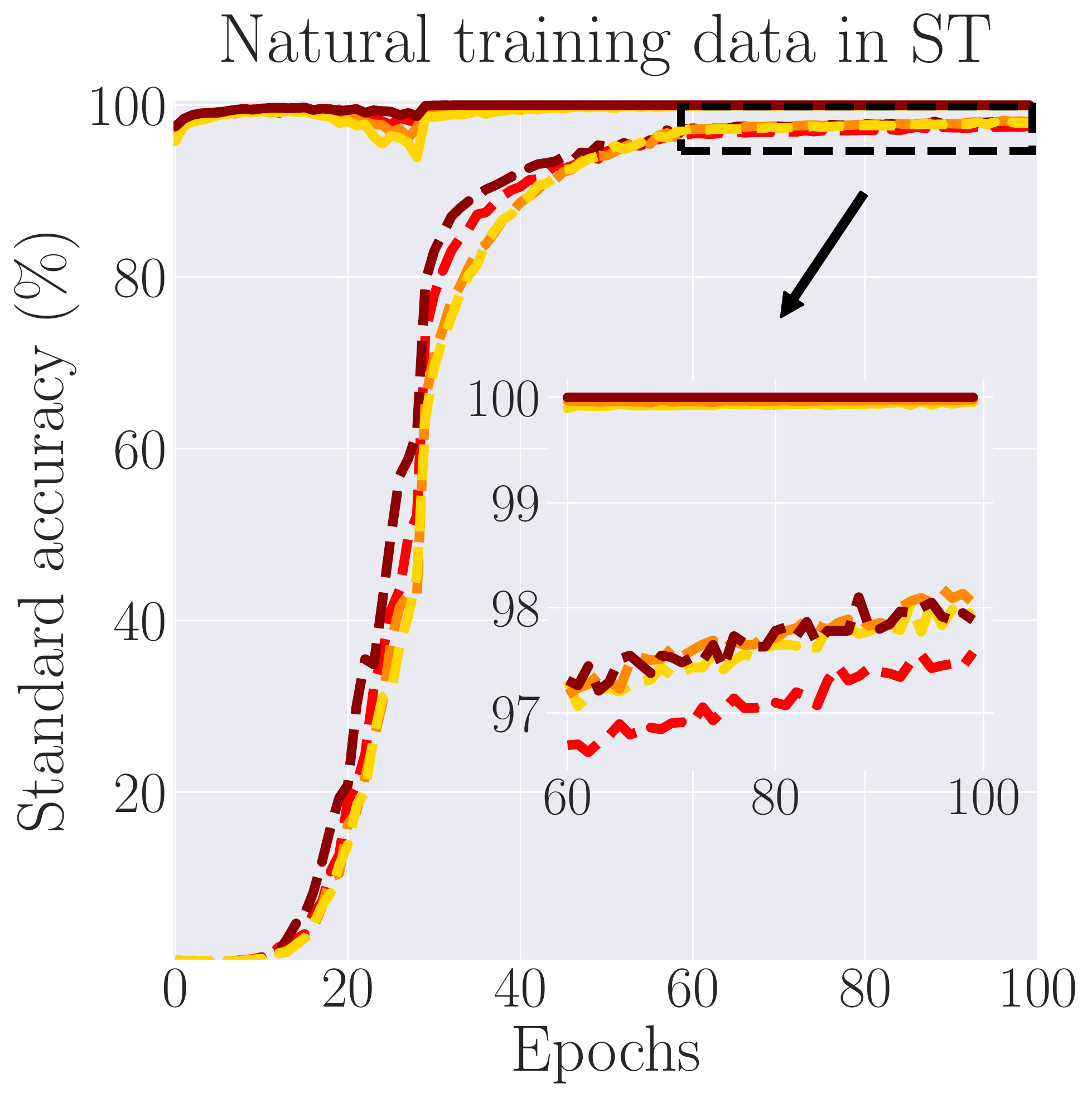}
    \includegraphics[scale=0.195]{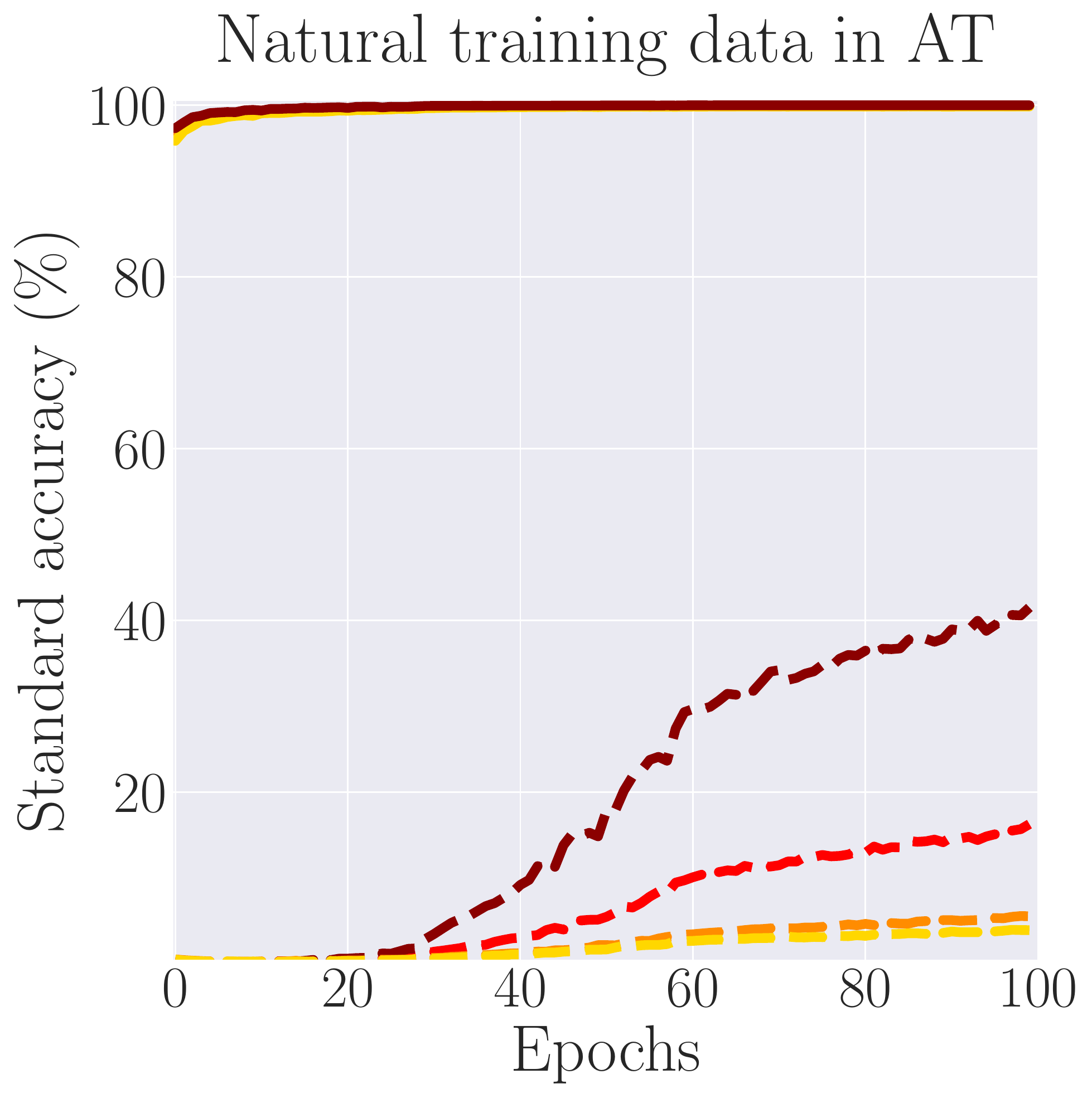}
    }
    \vspace{-2mm}
    \caption{The standard accuracy of ST and AT on correct/incorrect training data using \textit{CIFAR-10} and \textit{MNIST} with \textit{symmetric-flipping} noise. Solid lines denote the accuracy of correct training data, while dashed lines correspond to that of incorrect training data. Compared with ST, there is a large performance gap in the standard accuracy of correct/incorrect training data in AT.}
    %\vspace{2mm}
    \label{fig:appendix_part1_diff_st_at_label_noise_train_acc_sym}
\end{figure*}

\begin{figure*}[h!]
\vspace{4mm}
    \centering
    \subfigure[CIFAR-10]{
    \includegraphics[scale=0.195]{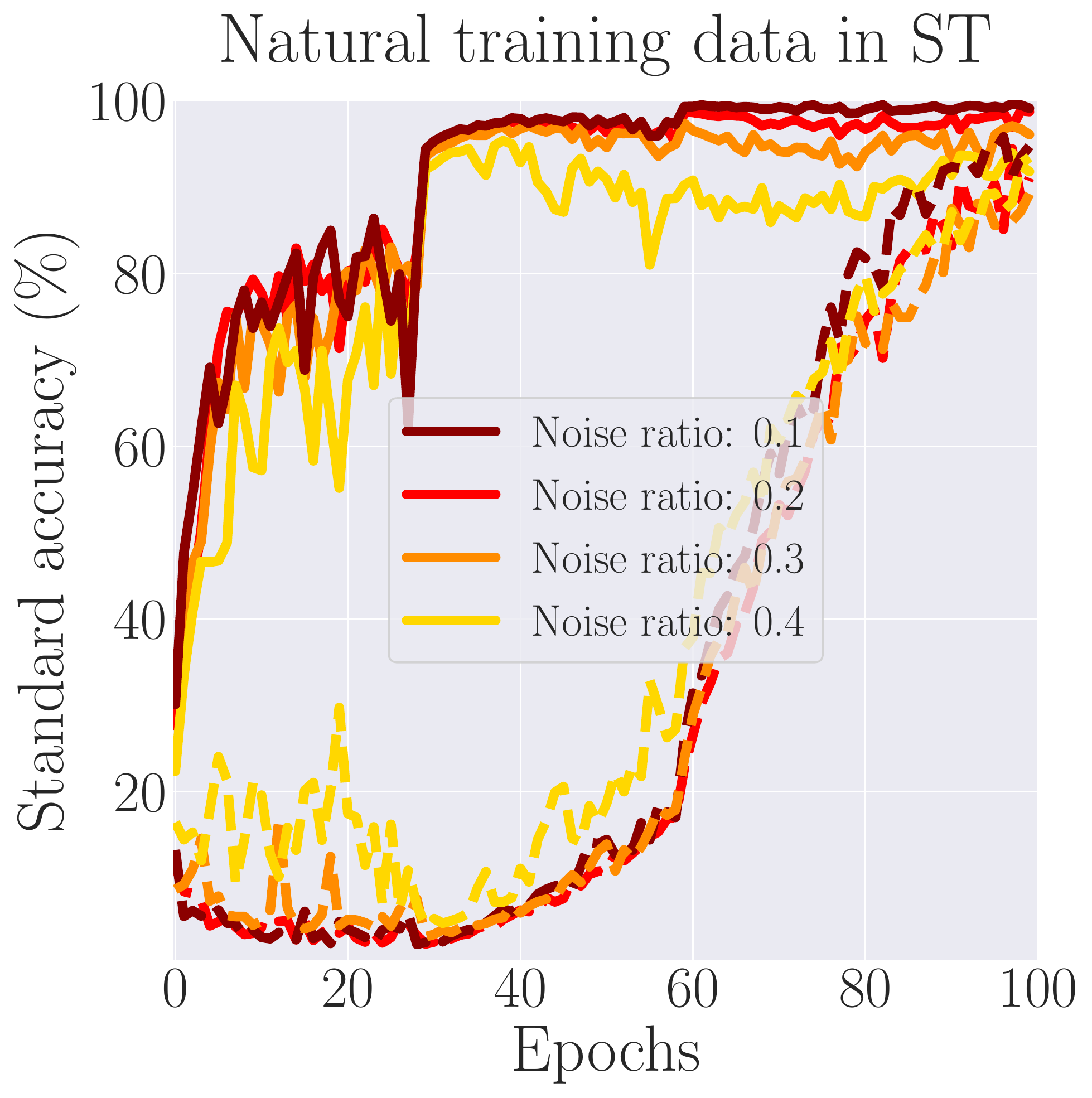}
    \includegraphics[scale=0.195]{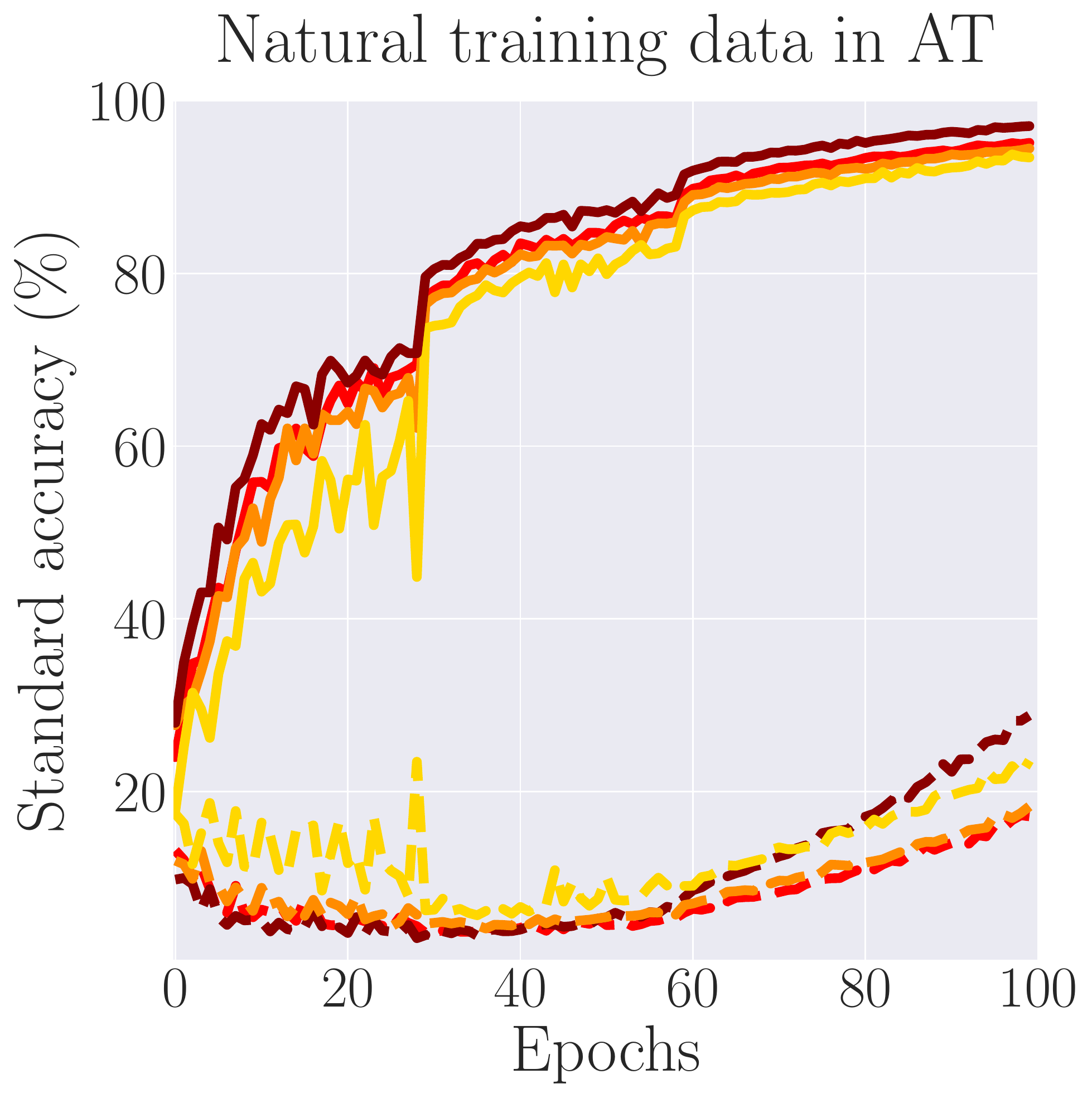}
    }
    \subfigure[MNIST]{
    \includegraphics[scale=0.195]{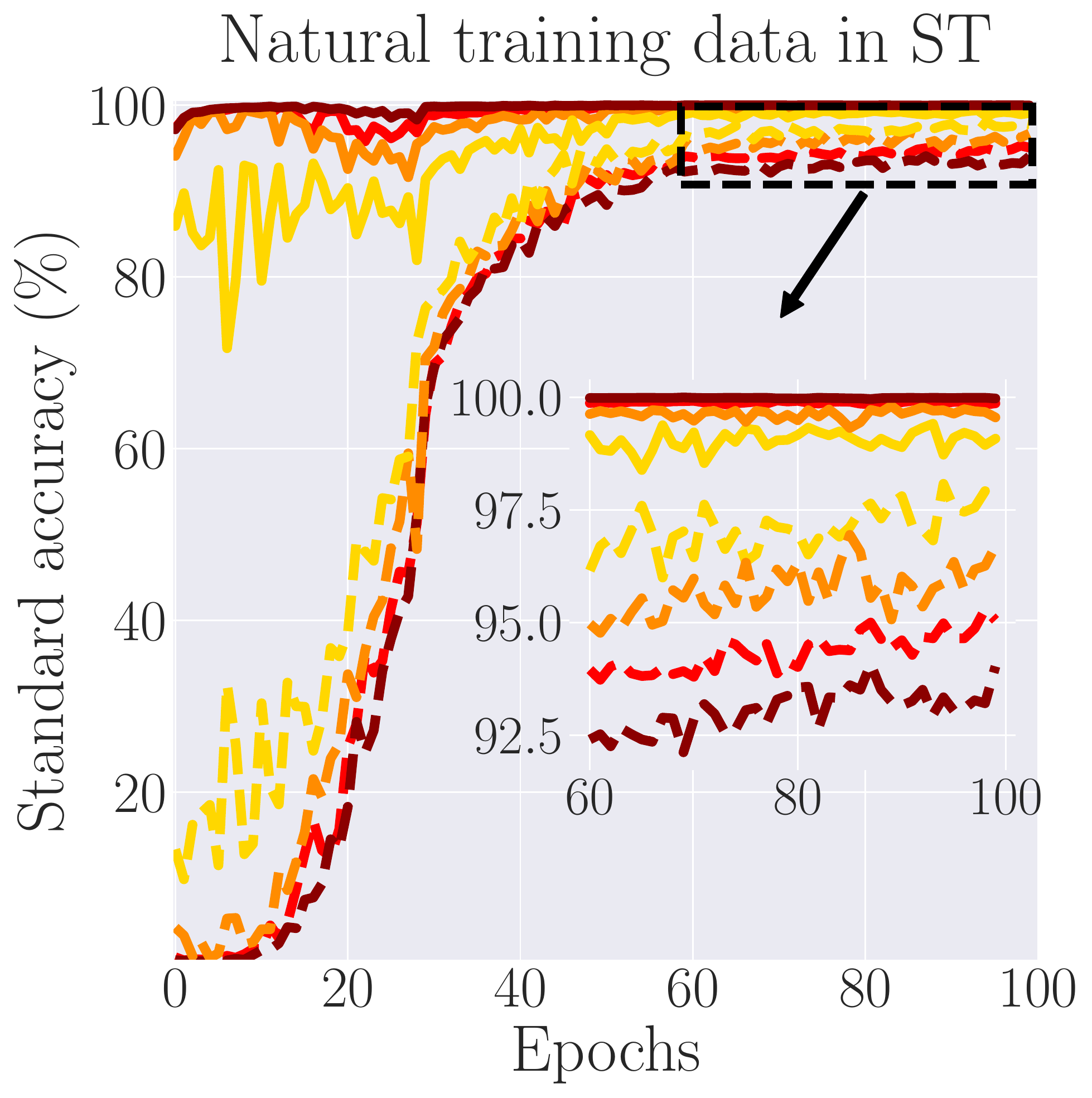}
    \includegraphics[scale=0.195]{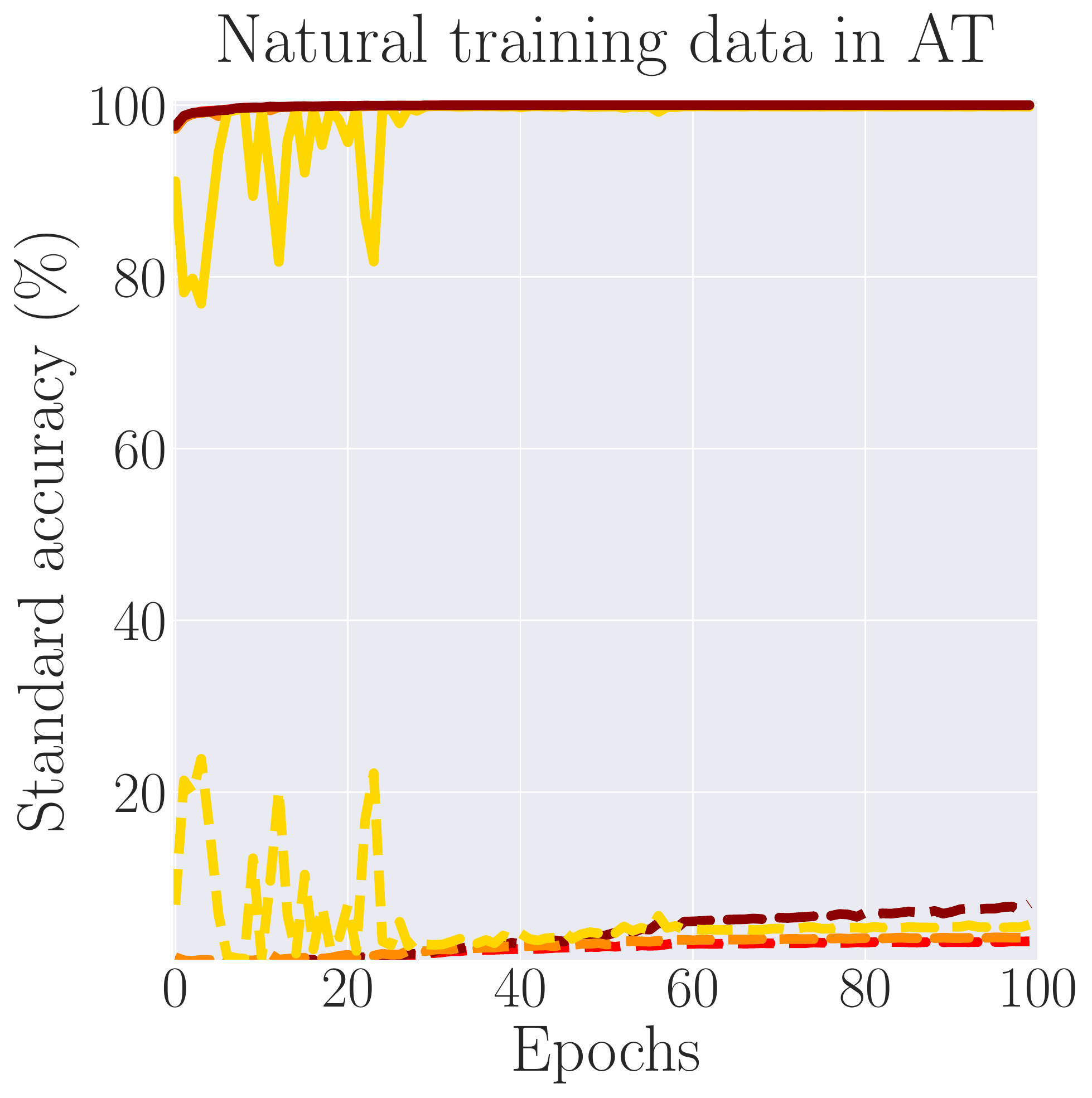}
    }
    \vspace{-2mm}
    \caption{The standard accuracy of ST and AT on correct/incorrect training data using \textit{CIFAR-10} and \textit{MNIST} with \textit{pair-flipping} noise. Solid lines denote the accuracy of correct training data, while dashed lines correspond to that of incorrect training data. Compared with ST, there is a large performance gap in the standard accuracy of correct/incorrect training data in AT.}
    %\vspace{2mm}
    \label{fig:appendix_part1_diff_st_at_label_noise_train_acc_asym}
\end{figure*}

\paragraph{Experimental setup.}
We conduct our experiments on two datasets with different noise rates (e.g., $[0.1,0.4]$) and different noise types (e.g., symmetric-flipping noise and pair-flipping noise). We use the method in~\citet{han2018co} to generate noisy training data. For the \textit{CIFAR-10} dataset, we normalize it into $[0,1]$: Each pixel is scaled by $1/255$. We perform the standard \textit{CIFAR-10} data augmentation: a random $4$ pixel crop followed by a random horizontal flip. For the \textit{MNIST} dataset, we normalize it into $[0,1]$. We train ResNet-18 in ST and AT using SGD with $0.9$ momentum for $100$ epochs on \textit{CIFAR-10} dataset. The initial learning rate is $0.1$ divided by $10$ at Epoch $30$ and $60$ respectively. The weight decay=$0.0005$. For \textit{MNIST} dataset, we use SmallCNN~\citep{Zhang_trades}, and set the initial learning rate as $0.01$. The rest of the settings remain the same as training on \textit{CIFAR-10}. In AT, we set the perturbation bound $\epsilon = 0.031$, the PGD step size $\alpha = 0.007$, and PGD step numbers $K = 10$. For standard evaluation, we obtain standard accuracy on natural training data according to correct/incorrect labels and natural test data with all correct labels. For the robust evaluation, we obtain robust accuracy on adversarial training and adversarial test data. The adversarial test data are generated by PGD-20 attack with the same perturbation bound $\epsilon=0.031$ and the step size $\alpha=0.031/4$, which keeps the same as \cite{Wang_Xingjun_MA_FOSC_DAT}. All PGD generation have a random start, i.e, the uniformly random perturbation of $[-\epsilon,\epsilon]$ added to the natural data before PGD iterations.

\paragraph{Result 1.}
In Figures~\ref{fig:appendix_part1_diff_st_at_label_noise_train_acc_sym} and~\ref{fig:appendix_part1_diff_st_at_label_noise_train_acc_asym}, we plot the standard accuracy of correct/incorrect training data with different noise rates and types. On the whole, compared with ST, correct/incorrect training data can be always distinguishable in AT regardless of noise rates and types. Compared with symmetric-flipping noise, AT has a better performance on distinguishing correct/incorrect data with pair-flipping noise. Note that, under the same noise rates, the standard accuracy on incorrect training data in AT with pair-flipping noise is lower than that with symmetric-flipping noise.

\begin{figure*}[h!]
\vspace{4mm}
    \centering
    \subfigure[CIFAR-10]{
    \includegraphics[scale=0.195]{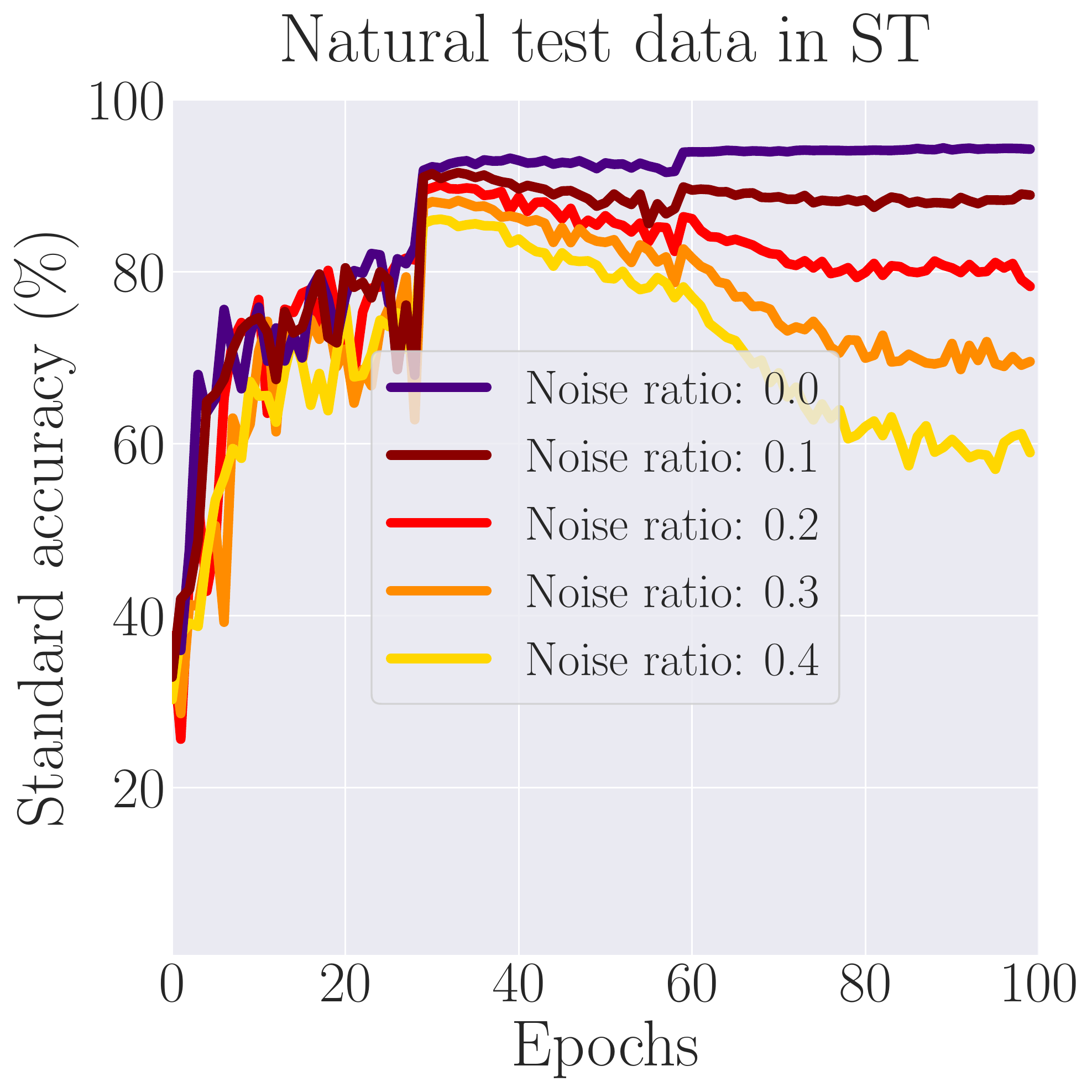}
    \includegraphics[scale=0.195]{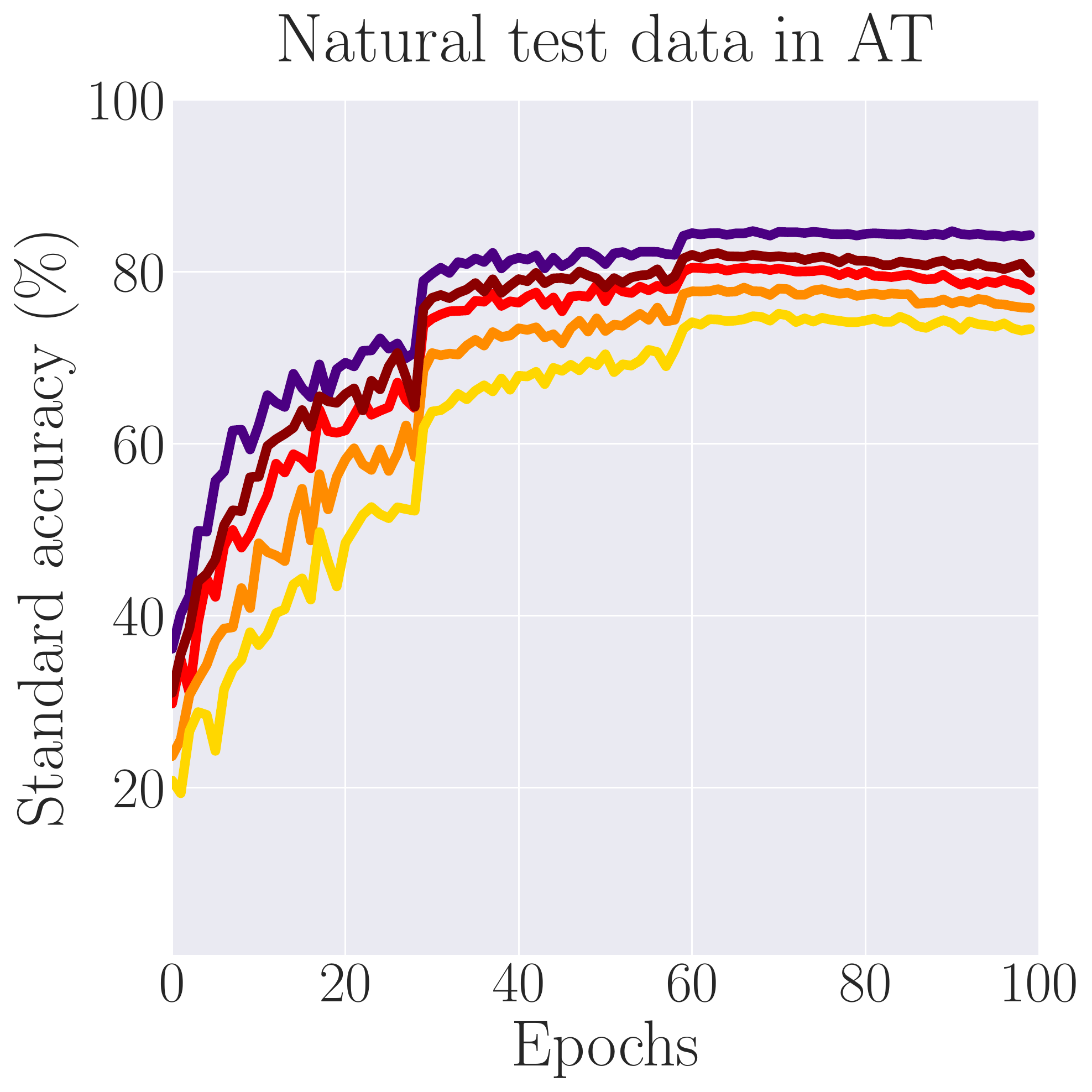}
    }
    \subfigure[MNIST]{
    \includegraphics[scale=0.195]{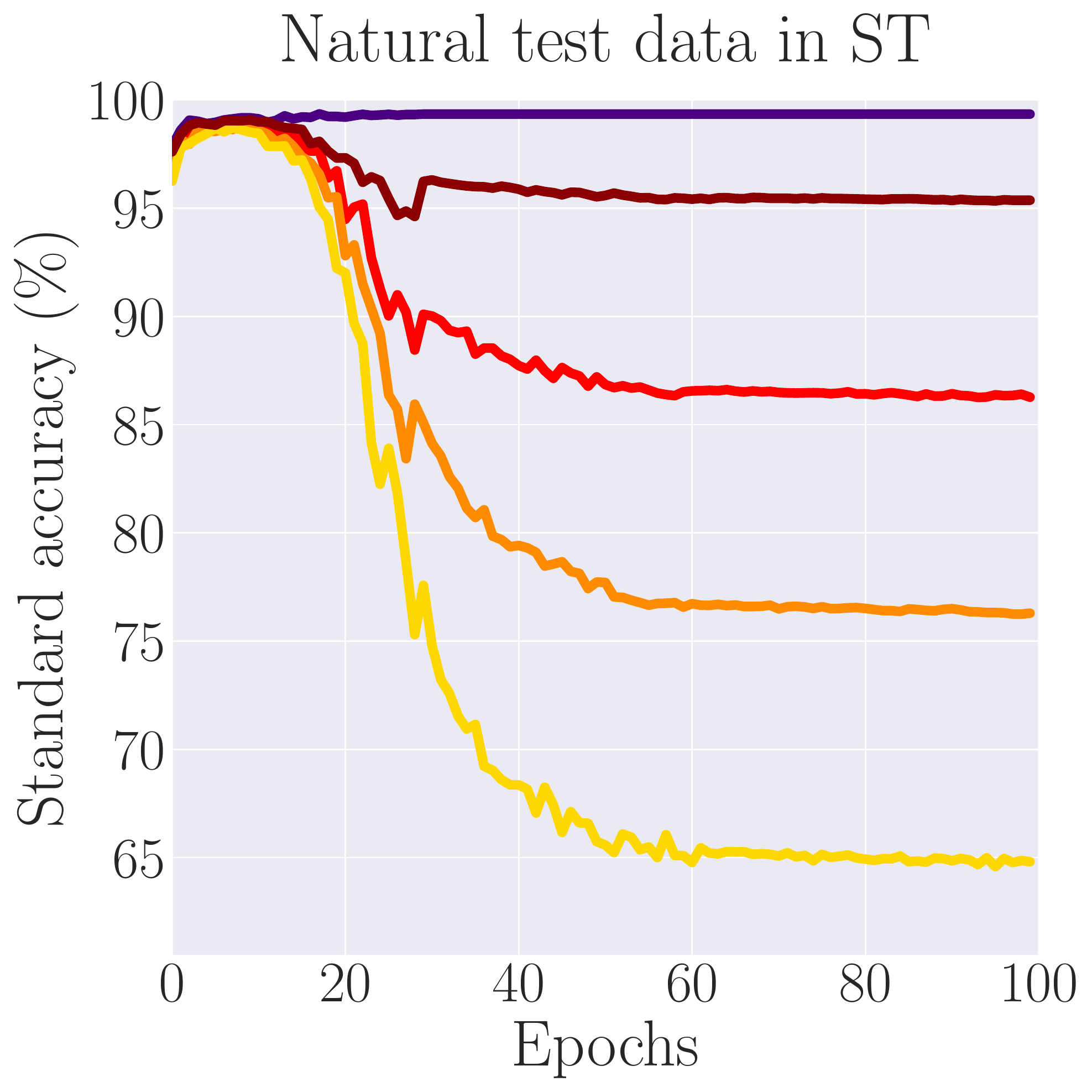}
    \includegraphics[scale=0.195]{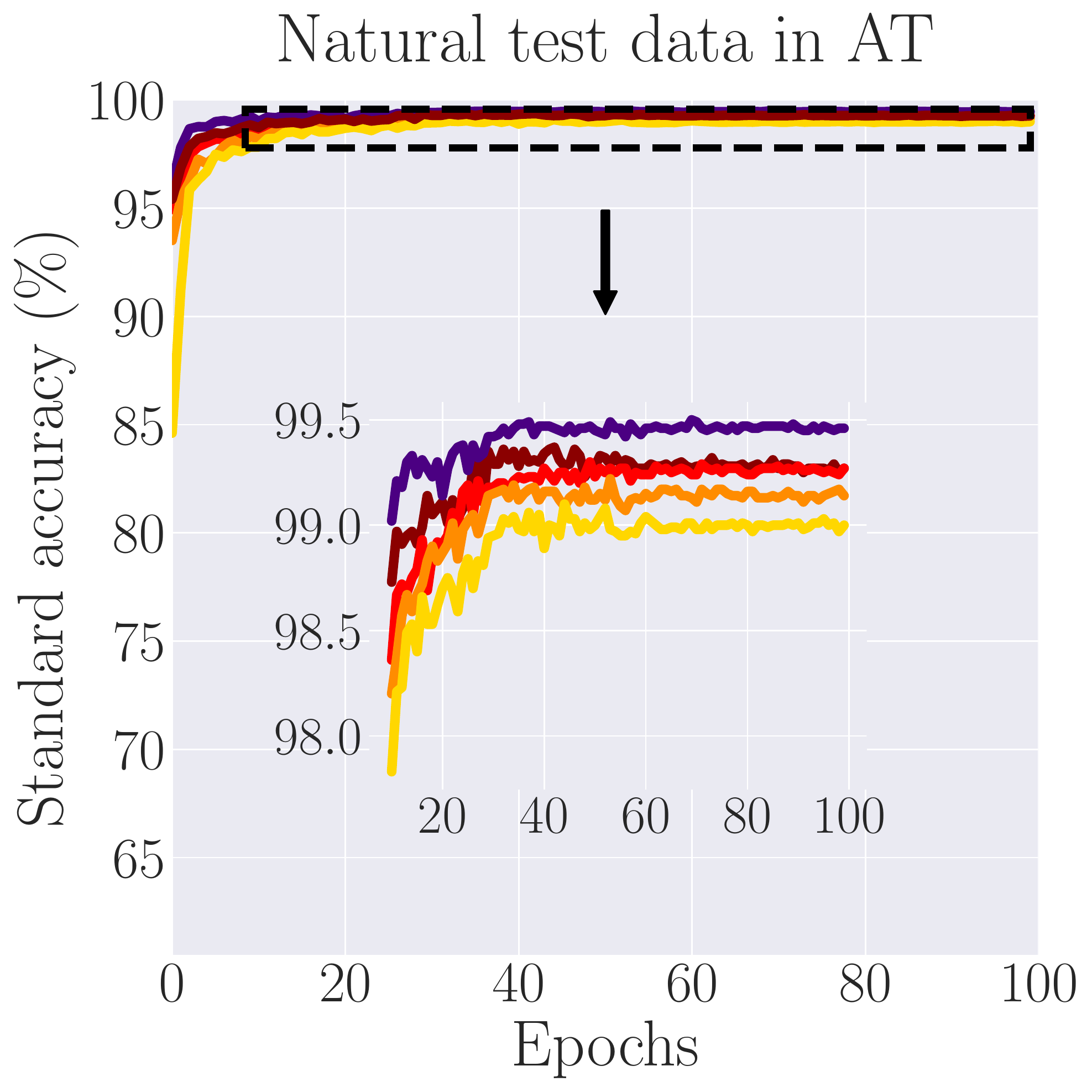}
    }
    \vspace{-2mm}
    \caption{The standard accuracy of ST and AT on natural test data, where training data using \textit{CIFAR-10} and \textit{MNIST} with \textit{symmetric-flipping} noise. Note that the larger noise rate causes the test accuracy of ST dropping more seriously due to memorization effects in deep learning, while AT alleviates such negative effects.}
    %\vspace{2mm}
    \label{fig:appendix_part1_diff_st_at_label_noise_test_acc_sym}
\end{figure*}

\begin{figure*}[h!]
\vspace{4mm}
    \centering
    \subfigure[CIFAR-10]{
    \includegraphics[scale=0.195]{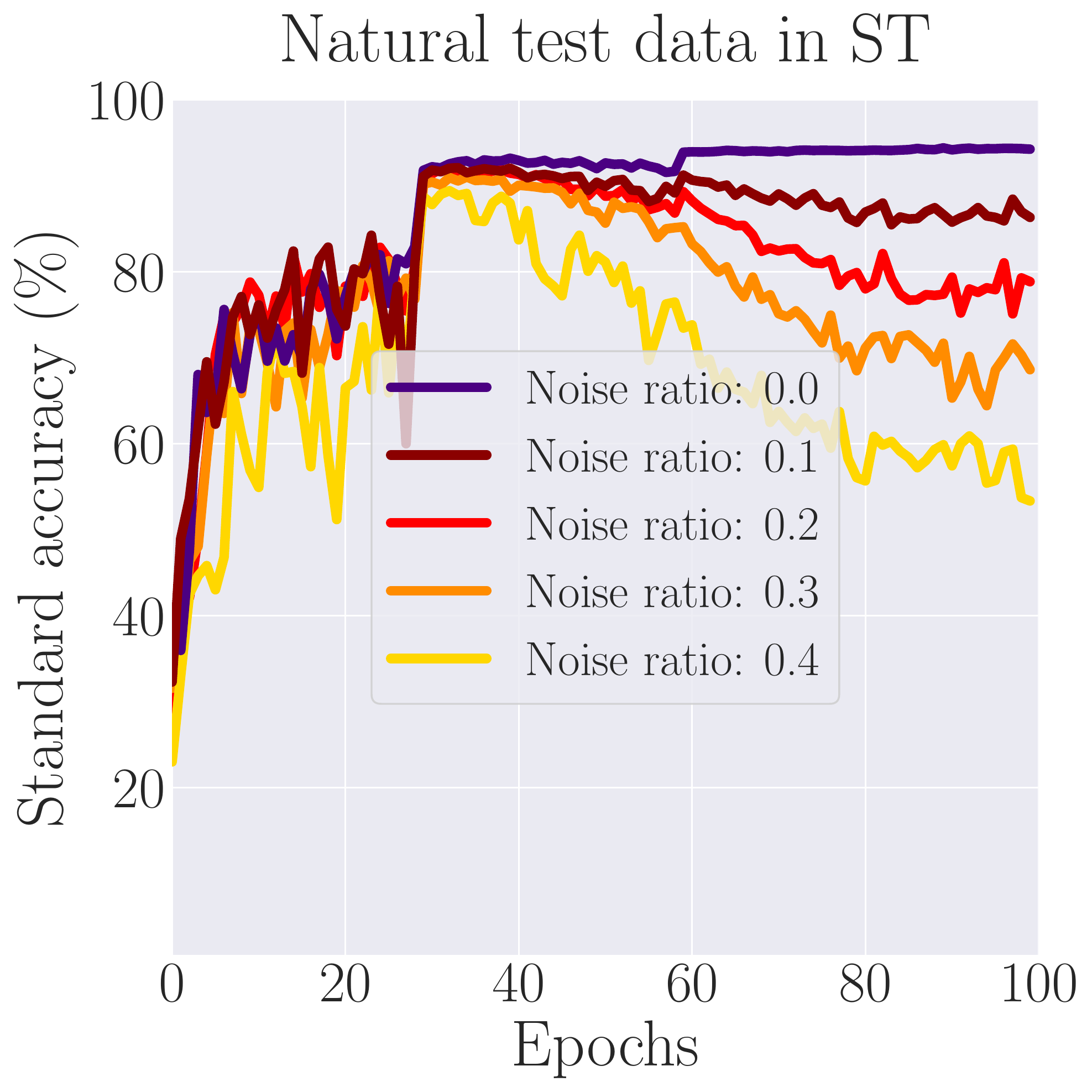}
    \includegraphics[scale=0.195]{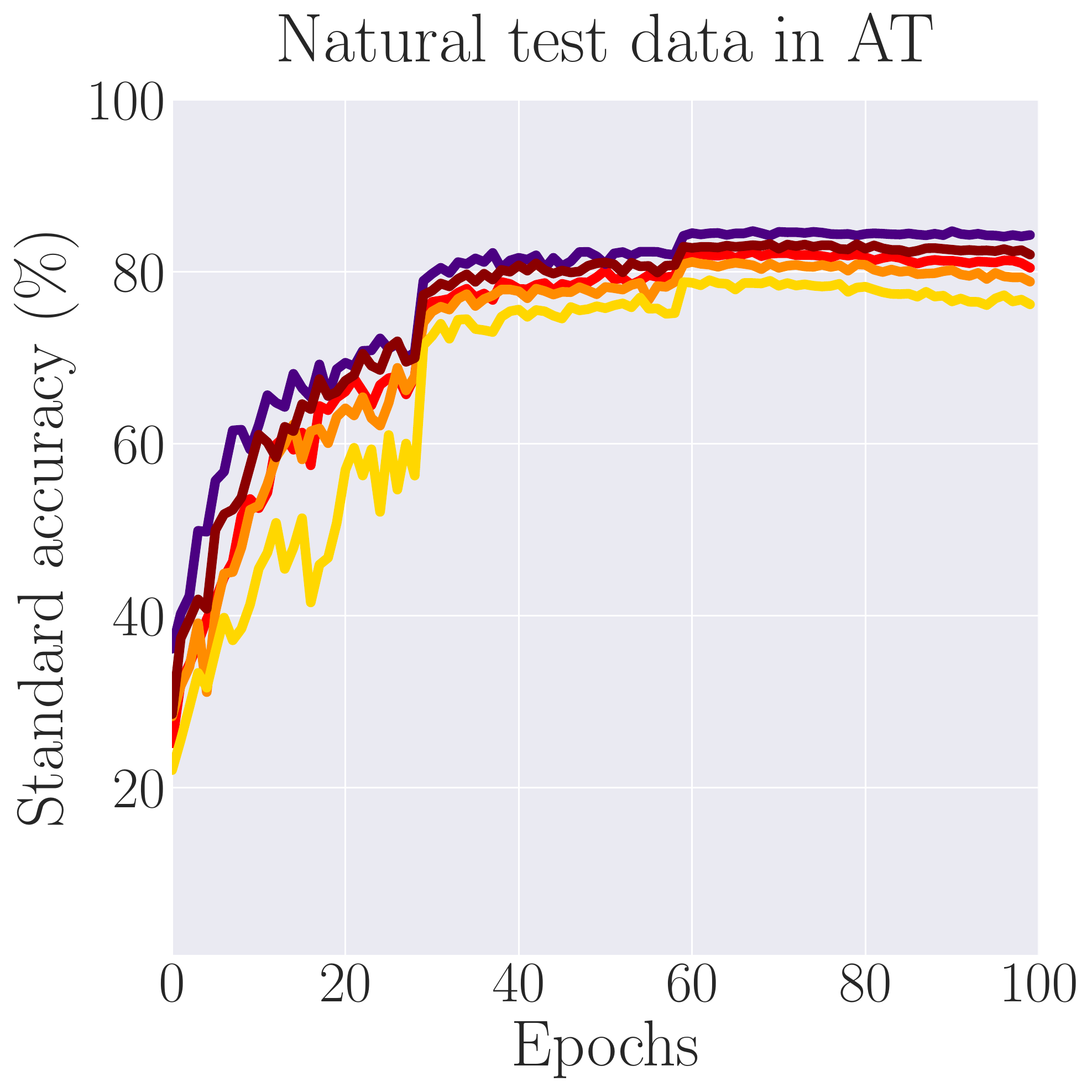}
    }
    \subfigure[MNIST]{
    \includegraphics[scale=0.195]{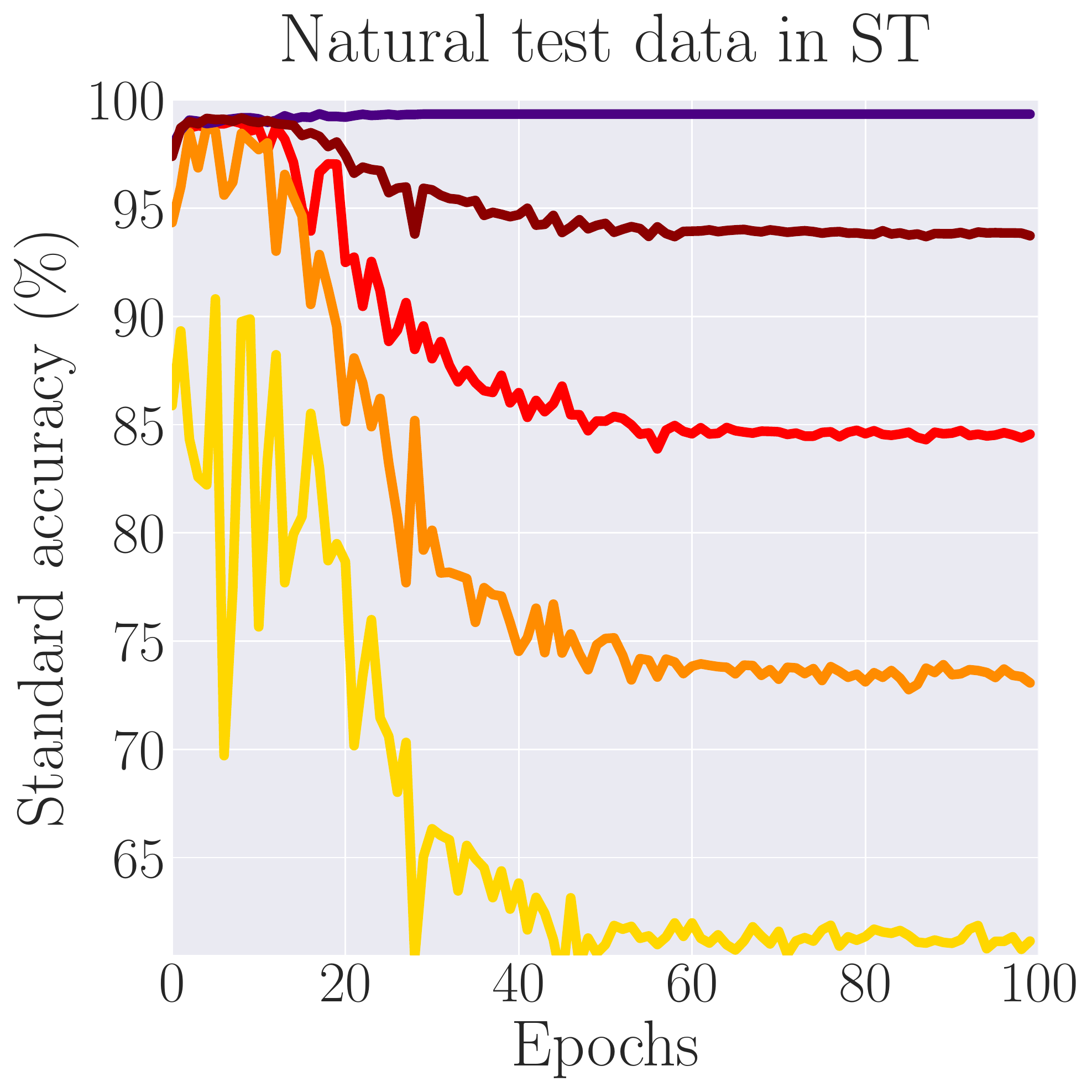}
    \includegraphics[scale=0.195]{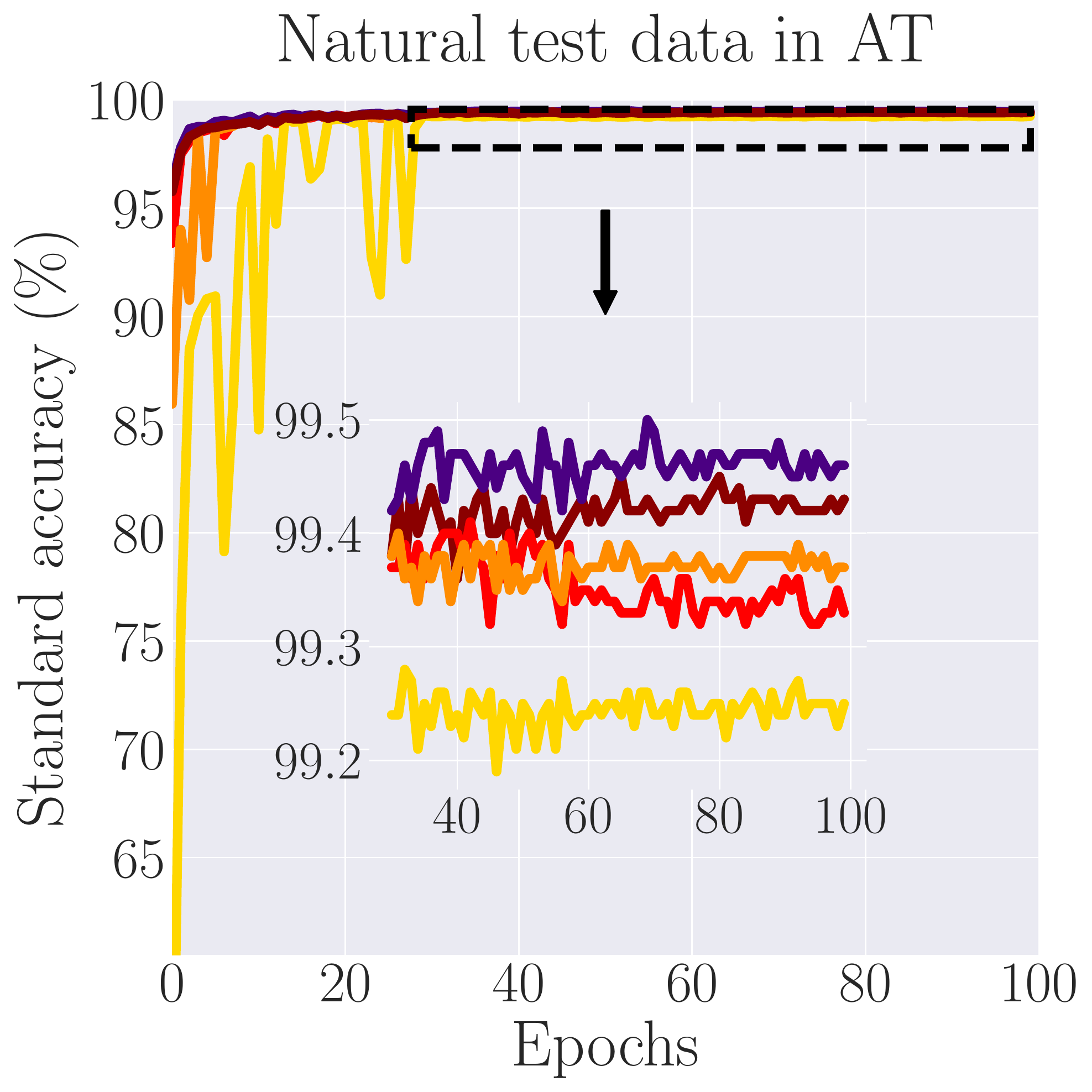}
    }
    \vspace{-2mm}
    \caption{The standard accuracy of ST and AT on natural test data, where training data using \textit{CIFAR-10} and \textit{MNIST} with \textit{pair-flipping} noise. Note that the larger noise rate causes the test accuracy of ST dropping more seriously due to memorization effects in deep learning, while AT alleviates such negative effects.}
    %\vspace{2mm}
    \label{fig:appendix_part1_diff_st_at_label_noise_test_acc_asym}
\end{figure*}

\paragraph{Result 2.}
In Figures~\ref{fig:appendix_part1_diff_st_at_label_noise_test_acc_sym} and~\ref{fig:appendix_part1_diff_st_at_label_noise_test_acc_asym}, we plot the standard accuracy of natural test data with different noise rates and types. On the whole, AT can alleviate the negative effects of label noise due to memorization effects in deep learning. Under each noise type, as the noise rate increases, standard accuracy of ST on natural test data drops more seriously in the later stage of training (e.g., after 60 epochs in \textit{CIFAR-10}). In AT, we only observe that the larger noise rate causes the lower standard accuracy on natural test data, while the overfitting phenomenon is not obvious. Compared with \textit{CIFAR-10}, both symmetirc-flipping and pair-flipping noise have more serious negative effects on \textit{MNIST}, while simply using AT can alleviate these effects to a greater extent.

\subsection{Natural Data and Adversarial Data}
\label{sec:app_nat_adv}

\begin{figure*}[h!]
\vspace{4mm}
    \centering
    \subfigure[Training accuracy]{
    \includegraphics[scale=0.195]{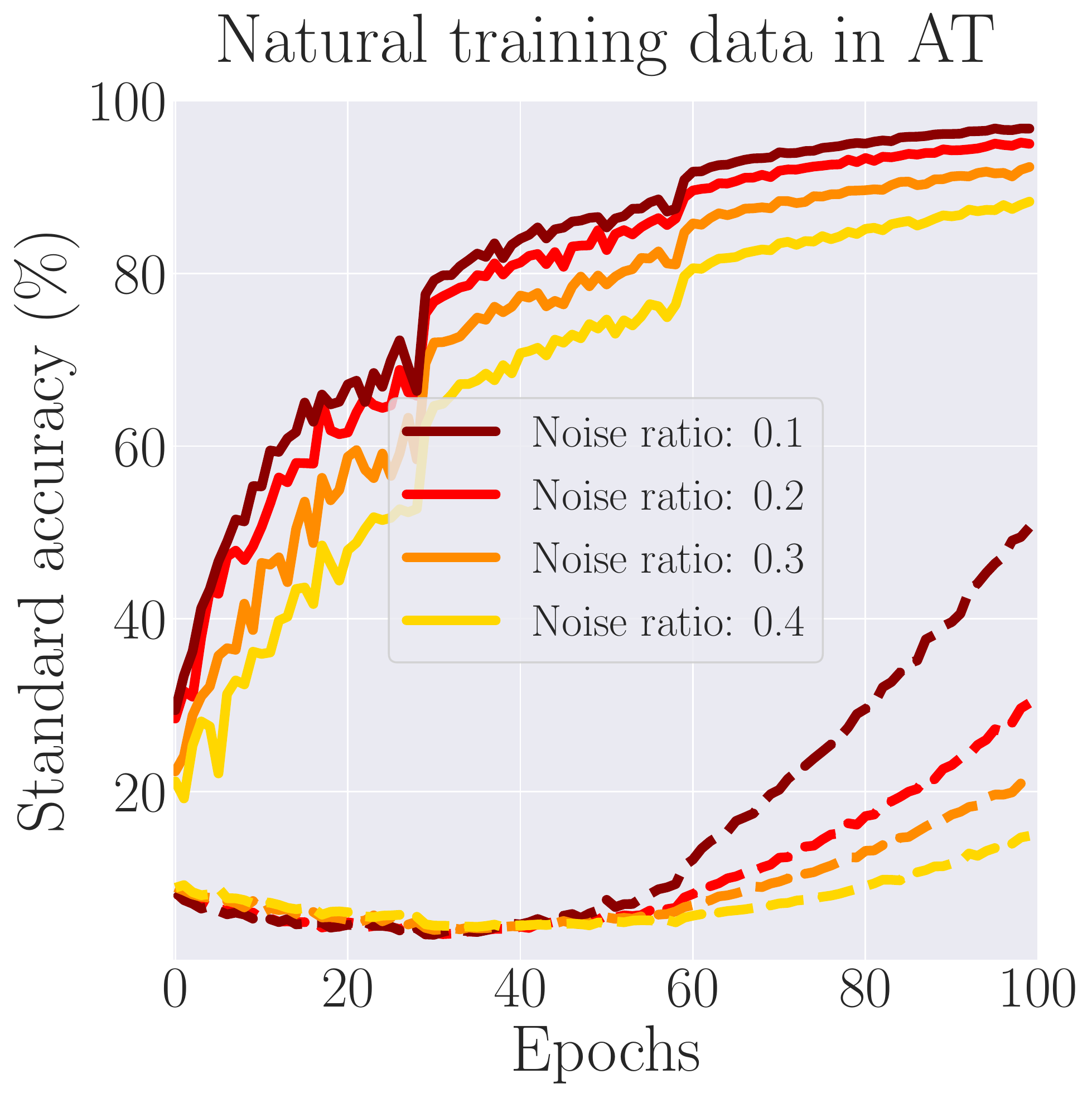}
    \includegraphics[scale=0.195]{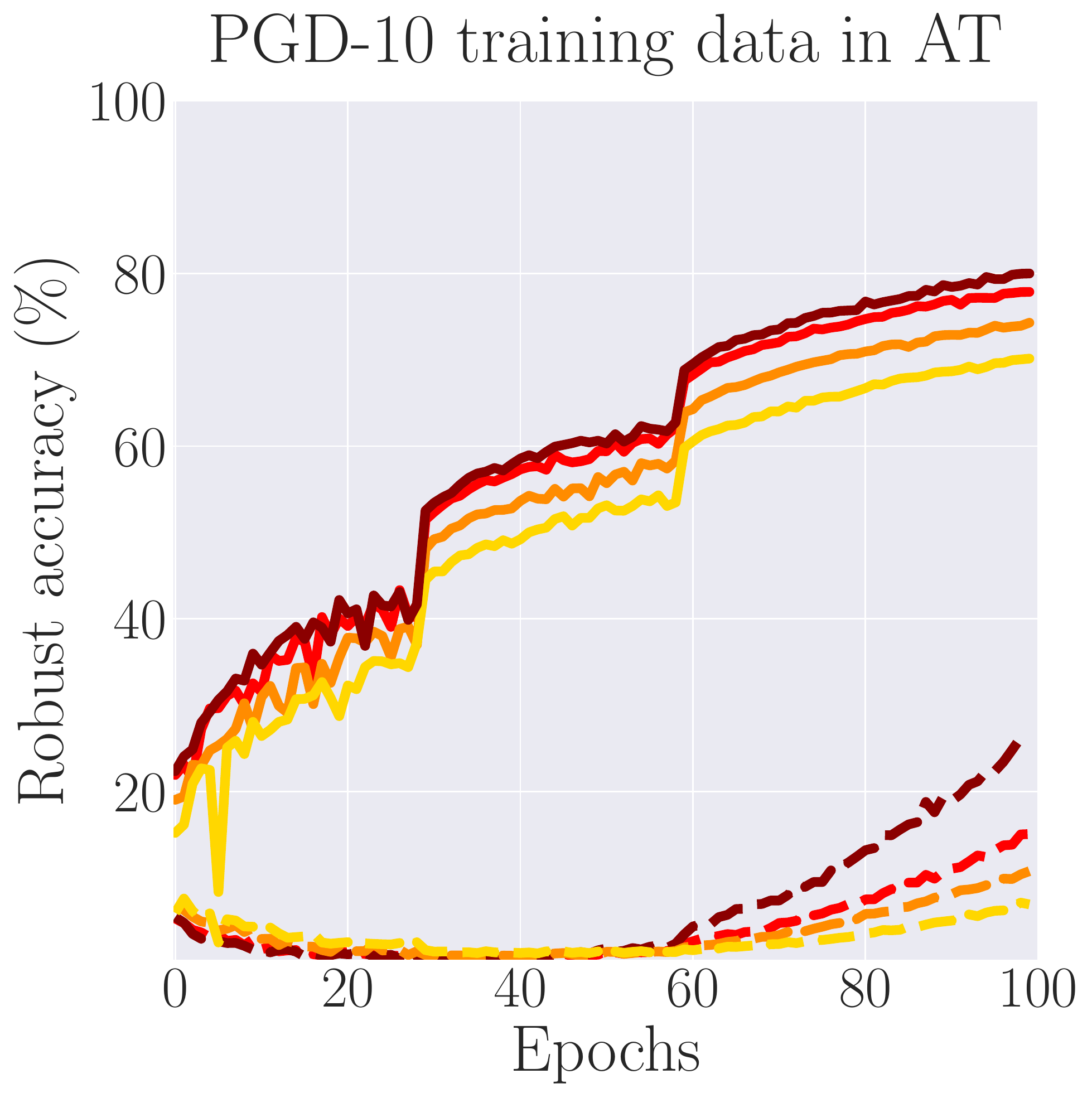}
    \label{a:train_adv}
    }
    \subfigure[Test accuracy]{
    \includegraphics[scale=0.195]{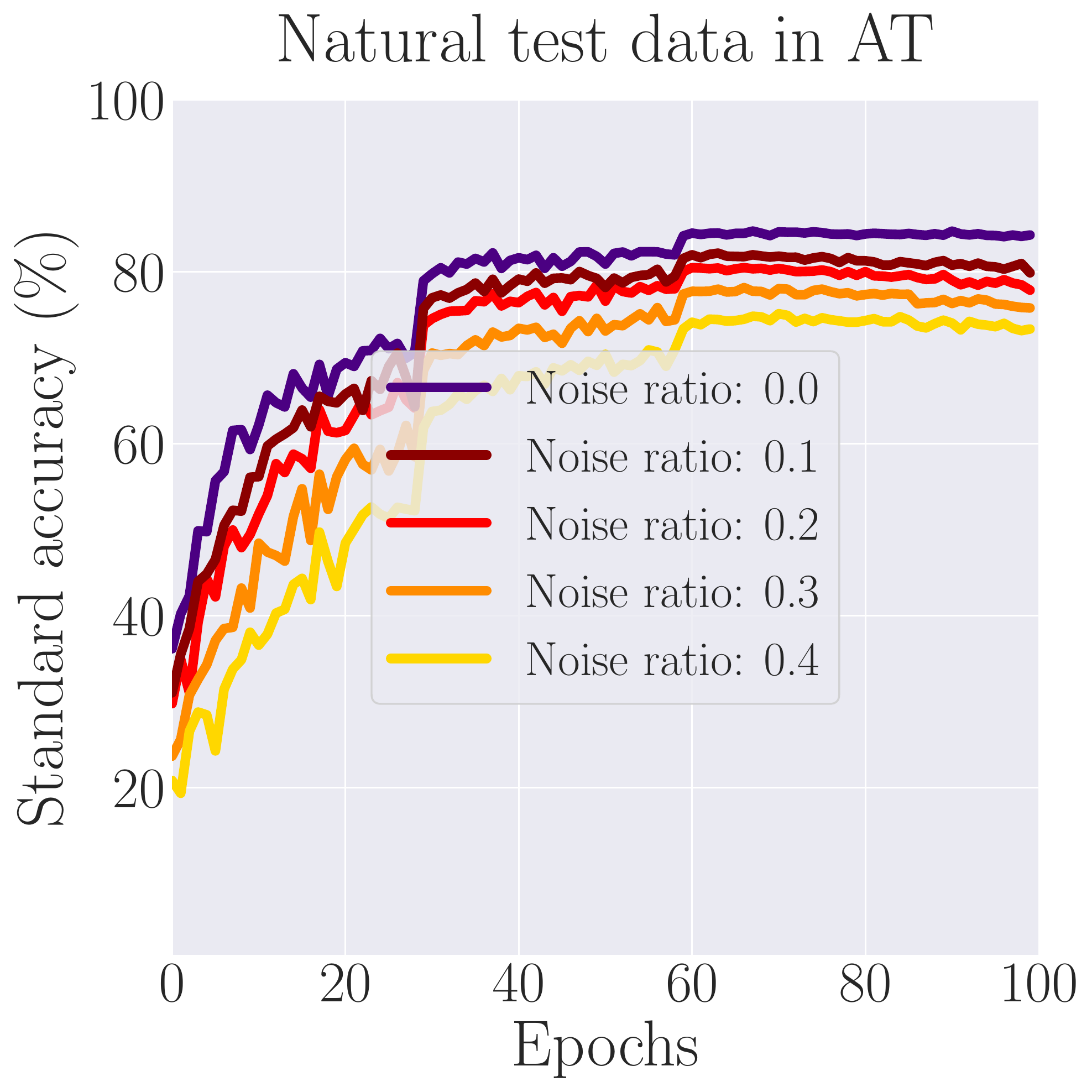}
    \includegraphics[scale=0.195]{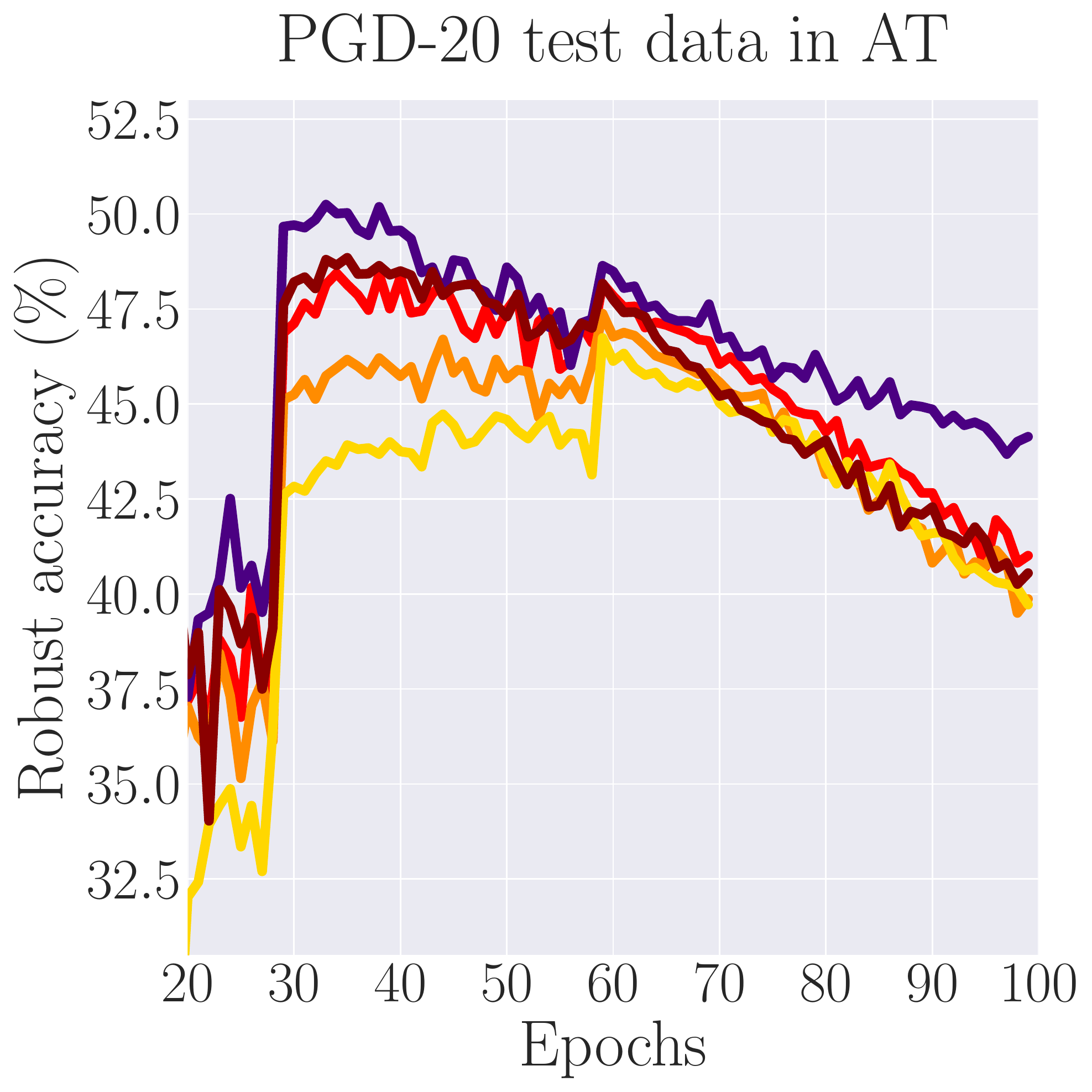}
    }
    \vspace{-2mm}
    \caption{The standard/robust accuracy of AT on natural training data, adversarial training data (PGD-10), adversarial test data (PGD-20) using the \textit{CIFAR-10} dataset with \textit{symmetric-flipping} noise.}
    %\vspace{2mm}
    \label{fig:appendix_nat_adv_acc_sym}
\end{figure*}

%Previous results confirm that the smoothing effect of AT prevents the model from memorizing the incorrect data, thus the standard accuracy on the incorrect training data are always lower than that of correct training data in AT. 

\paragraph{Result.} In Figure~\ref{fig:appendix_nat_adv_acc_sym}, we plot the standard and robust accuracy on natural data and adversarial data (e.g., PGD-10 training data and PGD-20 test data) using \textit{CIFAR-10} with symmetirc-flipping noise. Different from ST, each natural training data will generate a corresponding adversarial data in AT. We also check the difference in robust accuracy between correct and incorrect adversarial data during training. We found that AT can also distinguish correct/incorrect adversarial data over the whole training process. However, we find that the difference between correct and incorrect adversarial data (right panel in Figure~\ref{a:train_adv}) is smaller than that between incorrect and correct natural data (left panel in Figure~\ref{a:train_adv}).

\subsection{Different Networks}
\label{sec:app_diff_net}

\begin{figure*}[h!]
\vspace{4mm}
    \centering
    \subfigure[Training accuracy]{
    \includegraphics[scale=0.195]{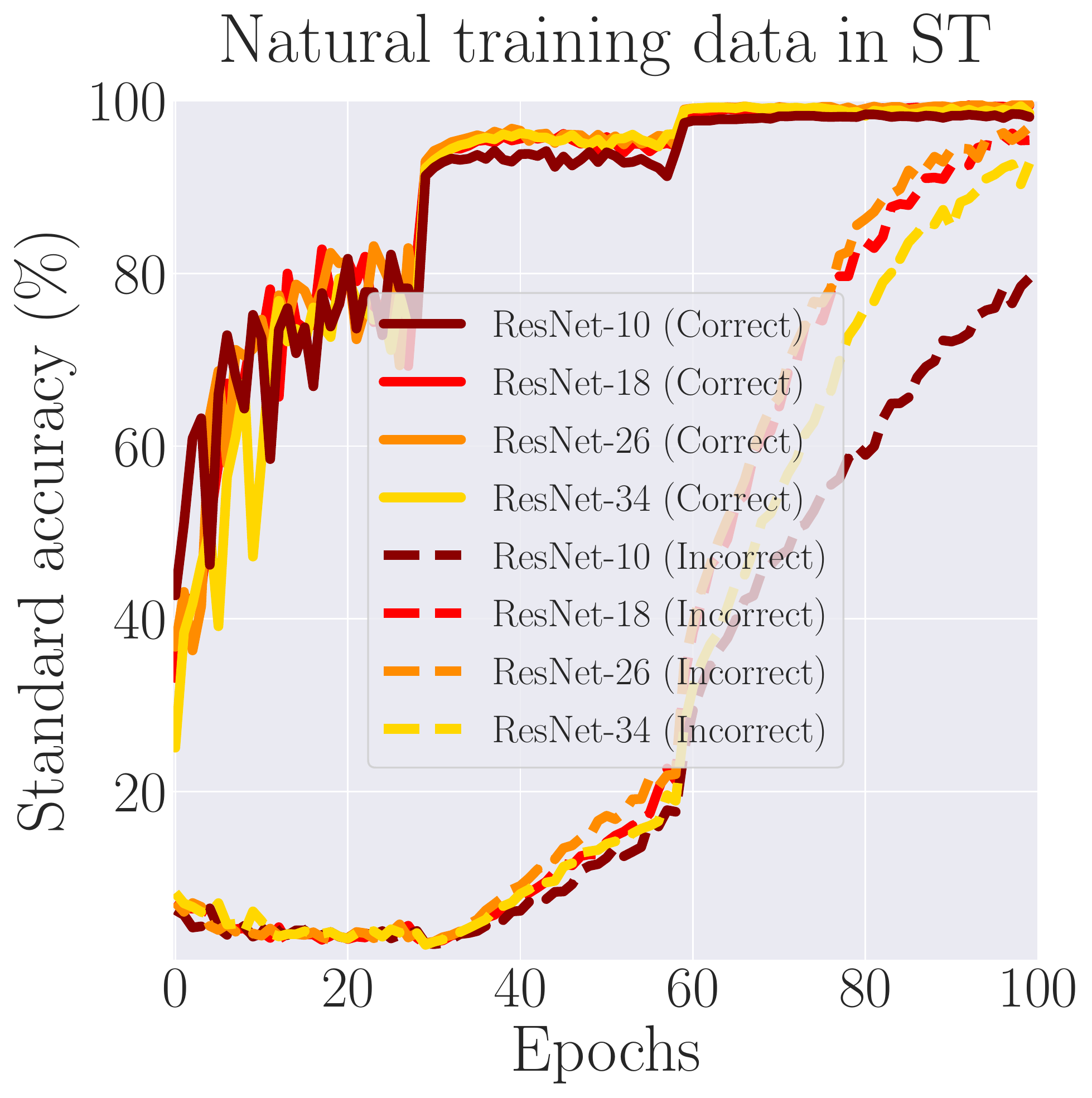}
    \includegraphics[scale=0.195]{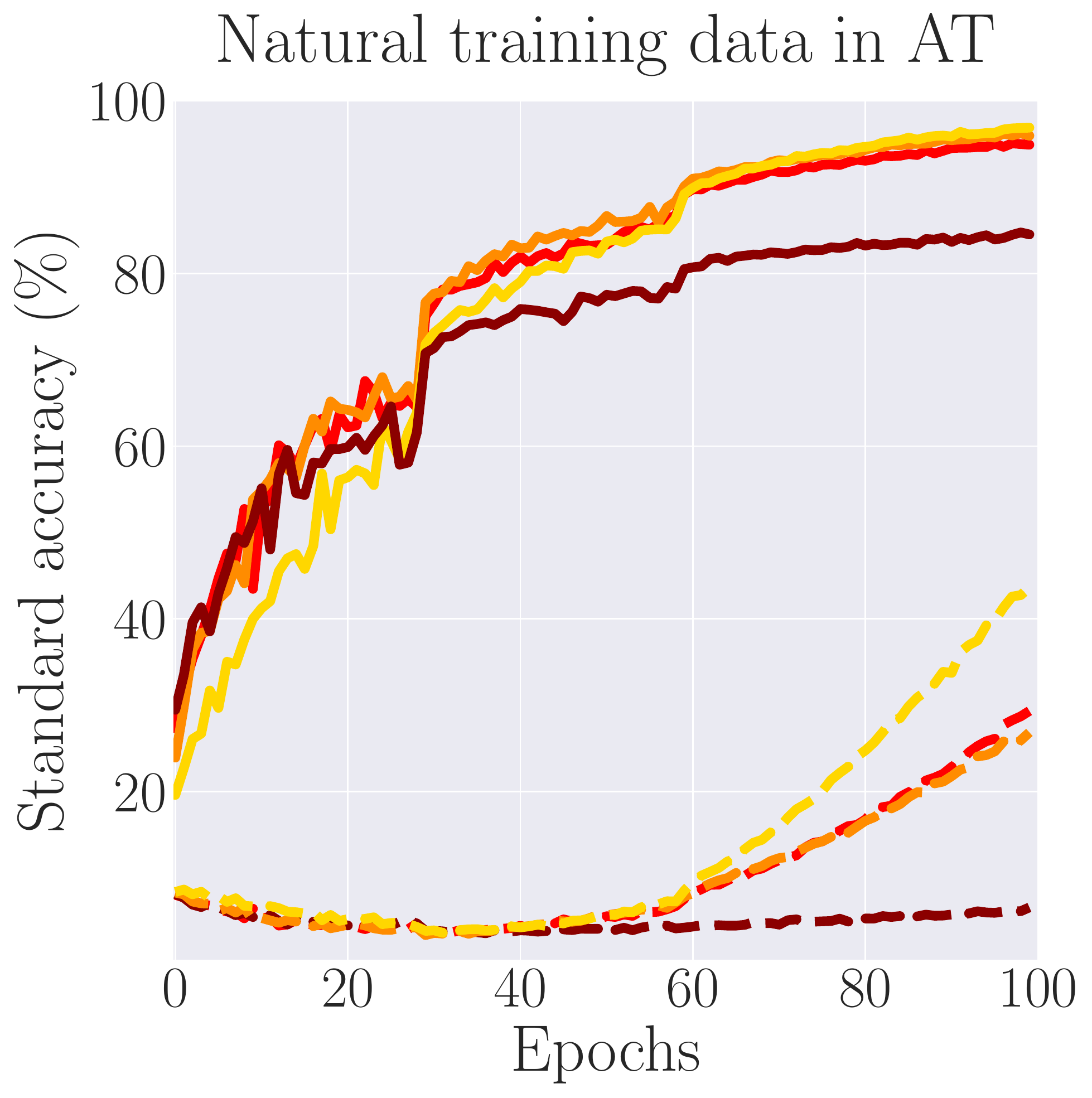}
    \label{a:train_adv}
    }
    \subfigure[Test accuracy]{
    \includegraphics[scale=0.195]{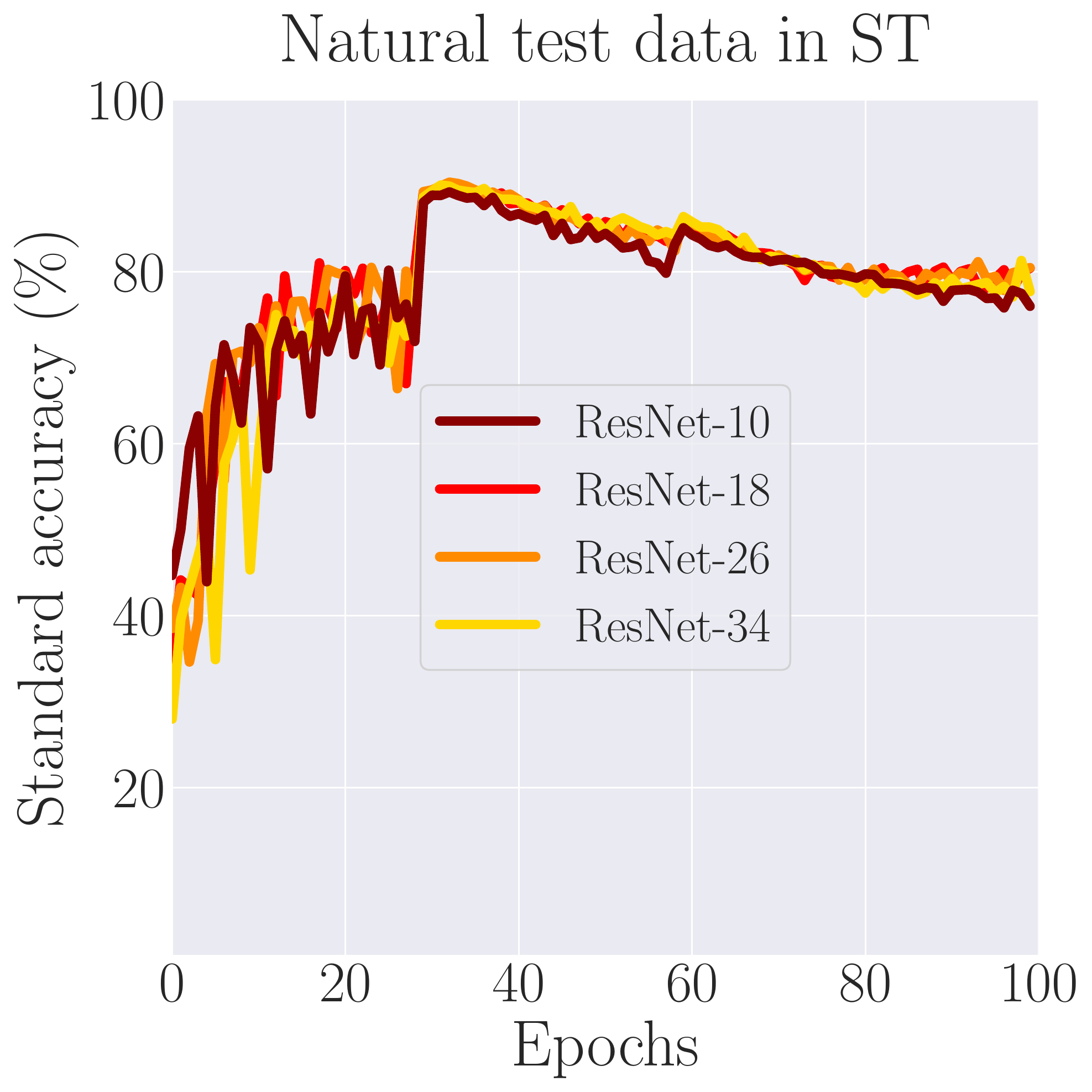}
    \includegraphics[scale=0.195]{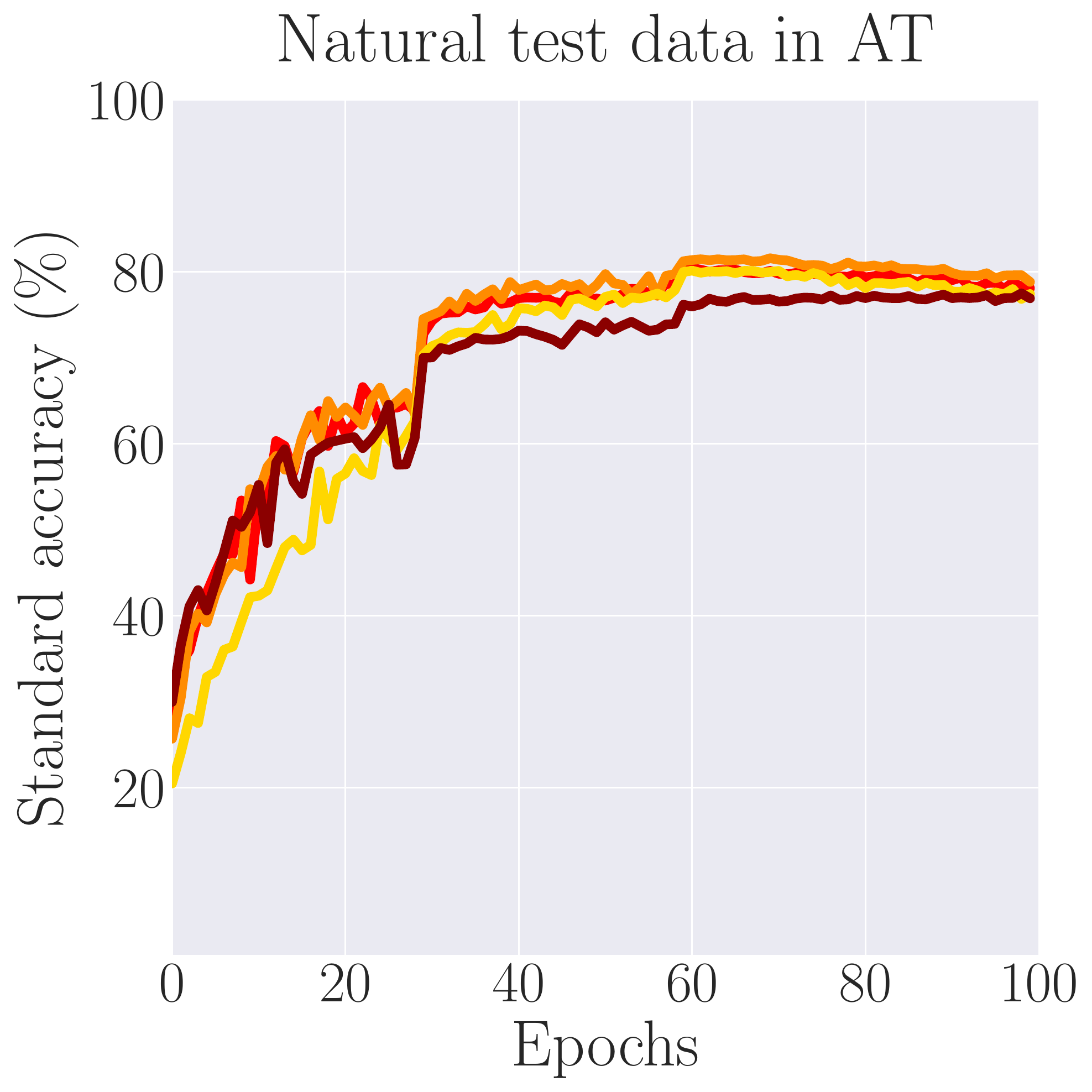}
    }
    \vspace{-2mm}
    \caption{The standard accuracy of AT on natural training/test data using the \textit{CIFAR-10} dataset with $20\%$ \textit{symmetric-flipping} noise. We conduct the experiments using ResNet-10, ResNet-18, ResNet-26 and ResNet-34.}
    %\vspace{2mm}
    \label{fig:appendix_nat_acc_sym_diffnet}
\end{figure*}

\paragraph{Result.} In Figure~\ref{fig:appendix_nat_acc_sym_diffnet}, we plot the standard accuracy on natural training/test data using ResNet-10, ResNet-18, ResNet-26 and ResNet-34 trained by ST and AT. We conduct the experiments using \textit{CIFAR-10} dataset with $20\%$ symmetric-flipping noise. The training settings keep the same as Appendix~\ref{sec:app_train_test}. We find that, using different networks, AT still has a better performance on distinguishing correct/incorrect data compared with ST and can alleviate the negative effects of label noise. 

\begin{figure*}[h!]
\vspace{4mm}
    \centering
    \subfigure[symmetirc-flipping noise]{
    \includegraphics[scale=0.195]{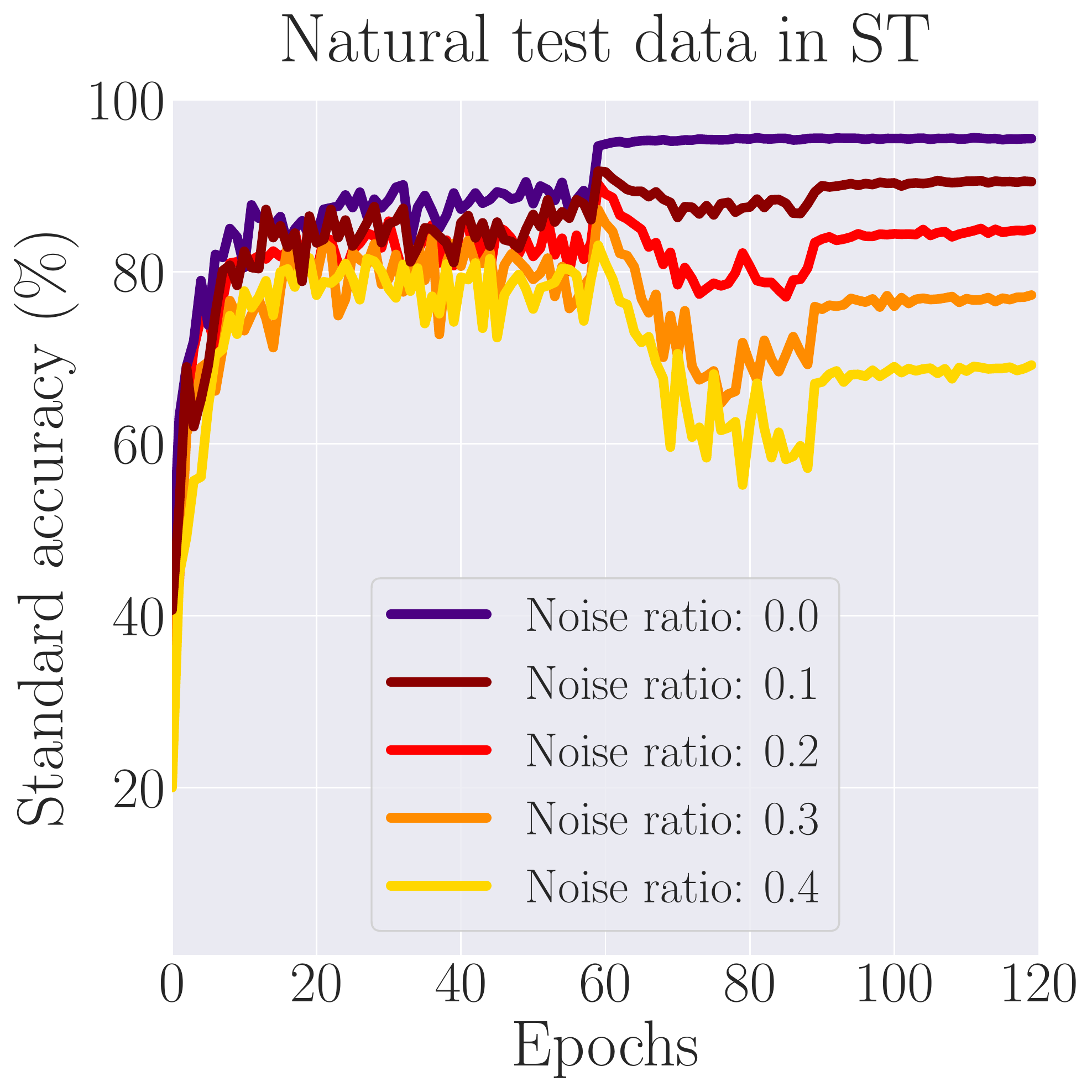}
    \includegraphics[scale=0.195]{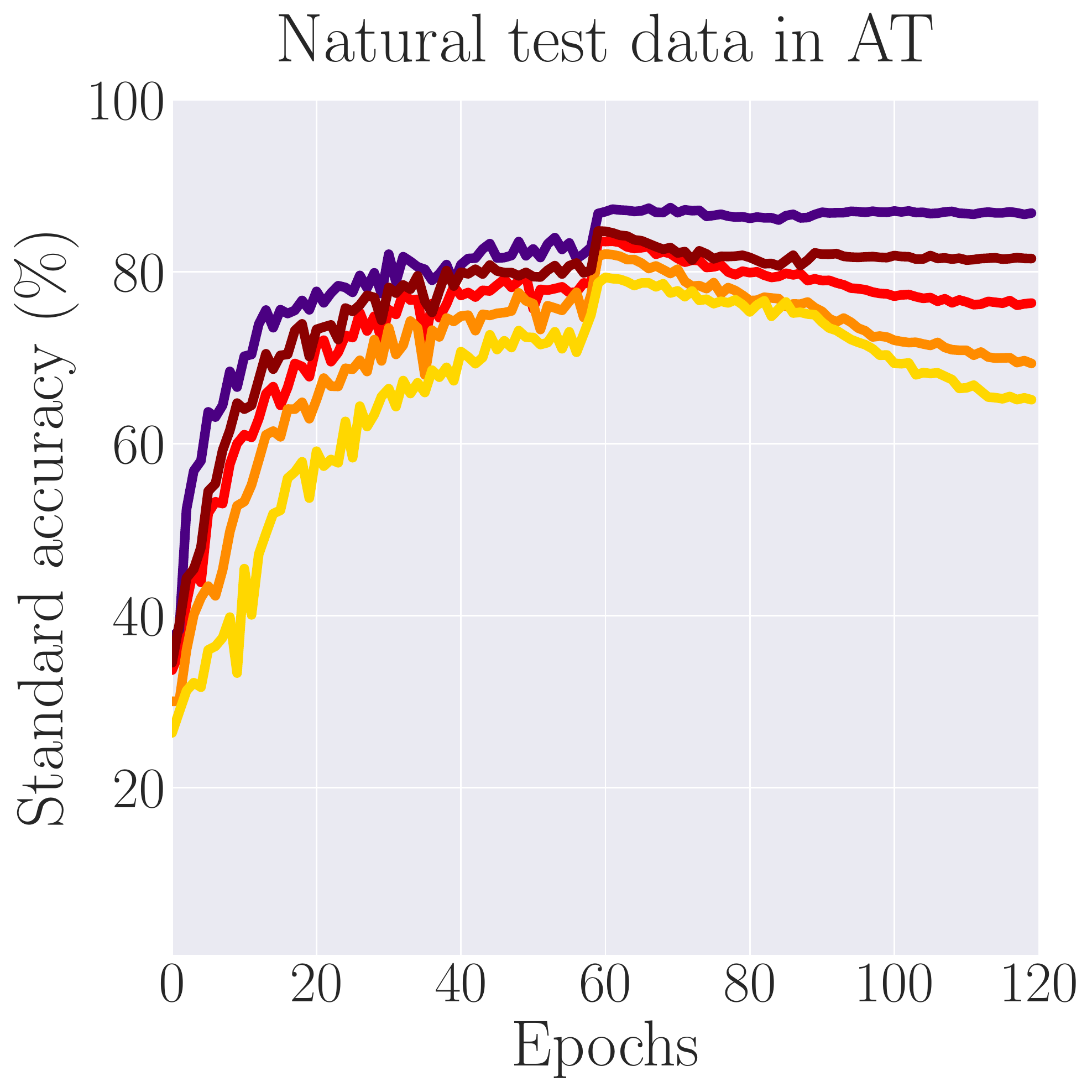}
    }
    \subfigure[pair-flipping noise]{
    \includegraphics[scale=0.195]{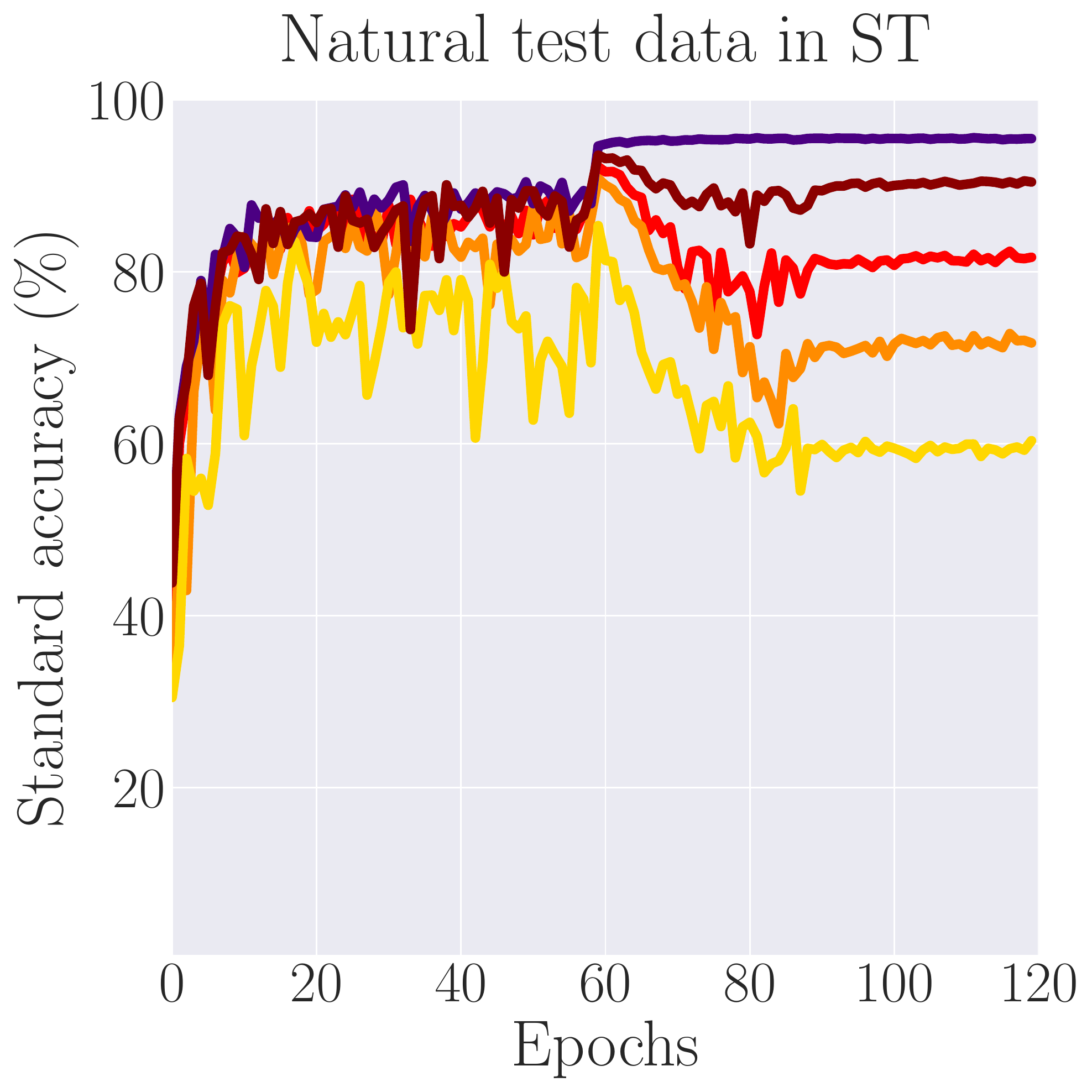}
    \includegraphics[scale=0.195]{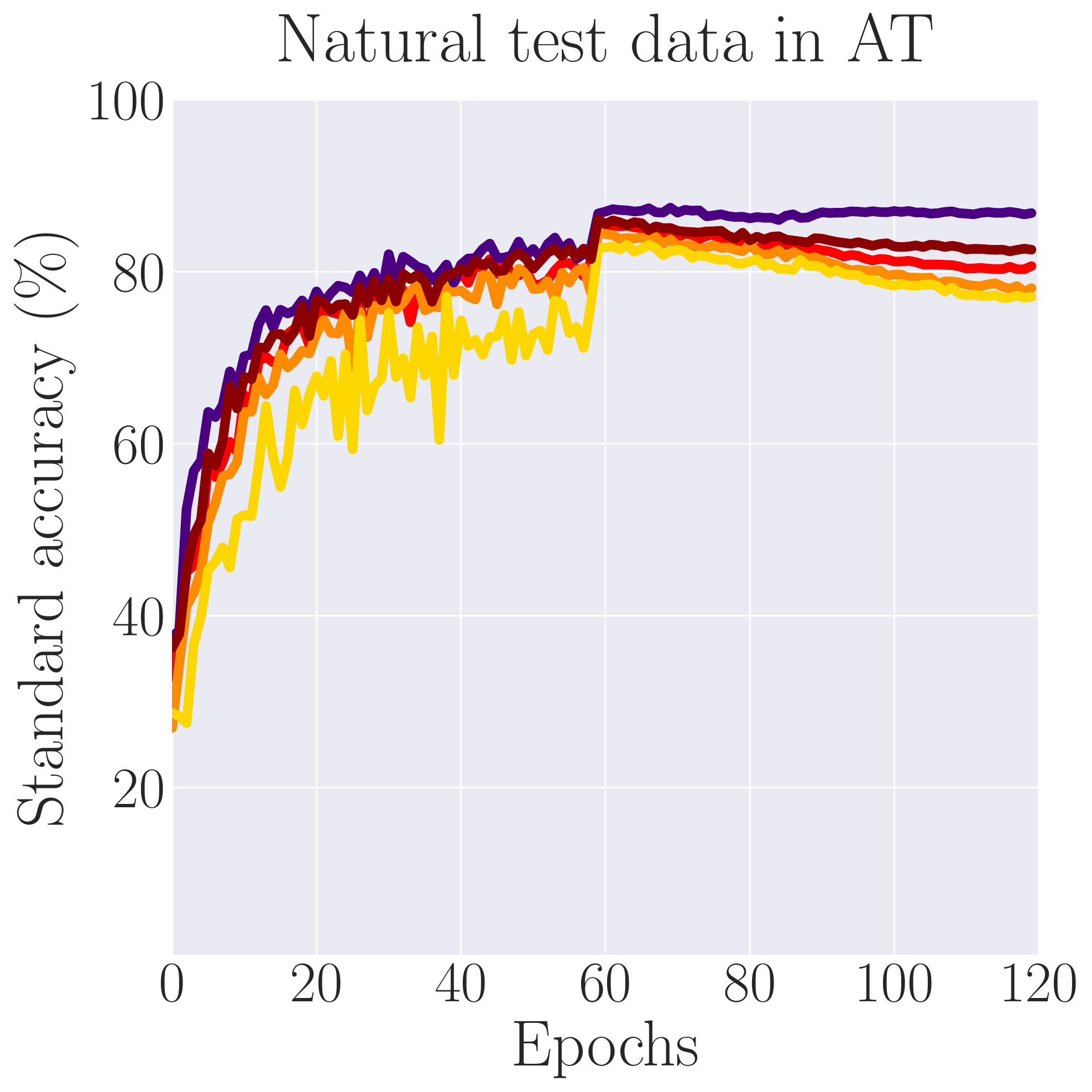}
    }
    \vspace{-2mm}
    \caption{The standard accuracy of AT on natural test data using the \textit{CIFAR-10} dataset with symmetric-flipping and pair-flipping noise. We conduct the experiments using WRN-32-10.}
    %\vspace{2mm}
    \label{fig:appendix_wrn_test_acc_sym}
\end{figure*}

\paragraph{Result.} In Figure~\ref{fig:appendix_wrn_test_acc_sym}, we plot the standard accuracy on natural test data using a large deep network, Wide-ResNet (e.g.,WRN-32-10~\citep{zagoruyko2016WRN}),  trained by ST and AT. We conduct the experiments using the \textit{CIFAR-10} dataset with different noise rates and types. We train the network for $120$ epochs and set the weight decay=$0.0002$, the rest of the settings keep the same as Appendix~\ref{sec:app_train_test}. We find that AT can still alleviate negative effects of label noise due to memorization effects of deep networks. Particularly, compared with the symmetric-flipping noise, AT has a better performance on avoiding memorization of pair-flipping noise, which can be confirmed by Figures~\ref{fig:appendix_part1_diff_st_at_label_noise_test_acc_sym} and~\ref{fig:appendix_part1_diff_st_at_label_noise_test_acc_asym}.

\subsection{The Loss Value}
\label{sec:app_loss_value}

\begin{figure*}[h!]
\vspace{4mm}
    \centering
    \hspace{1mm}
    \subfigure[CIFAR-10]{
    \includegraphics[scale=0.195]{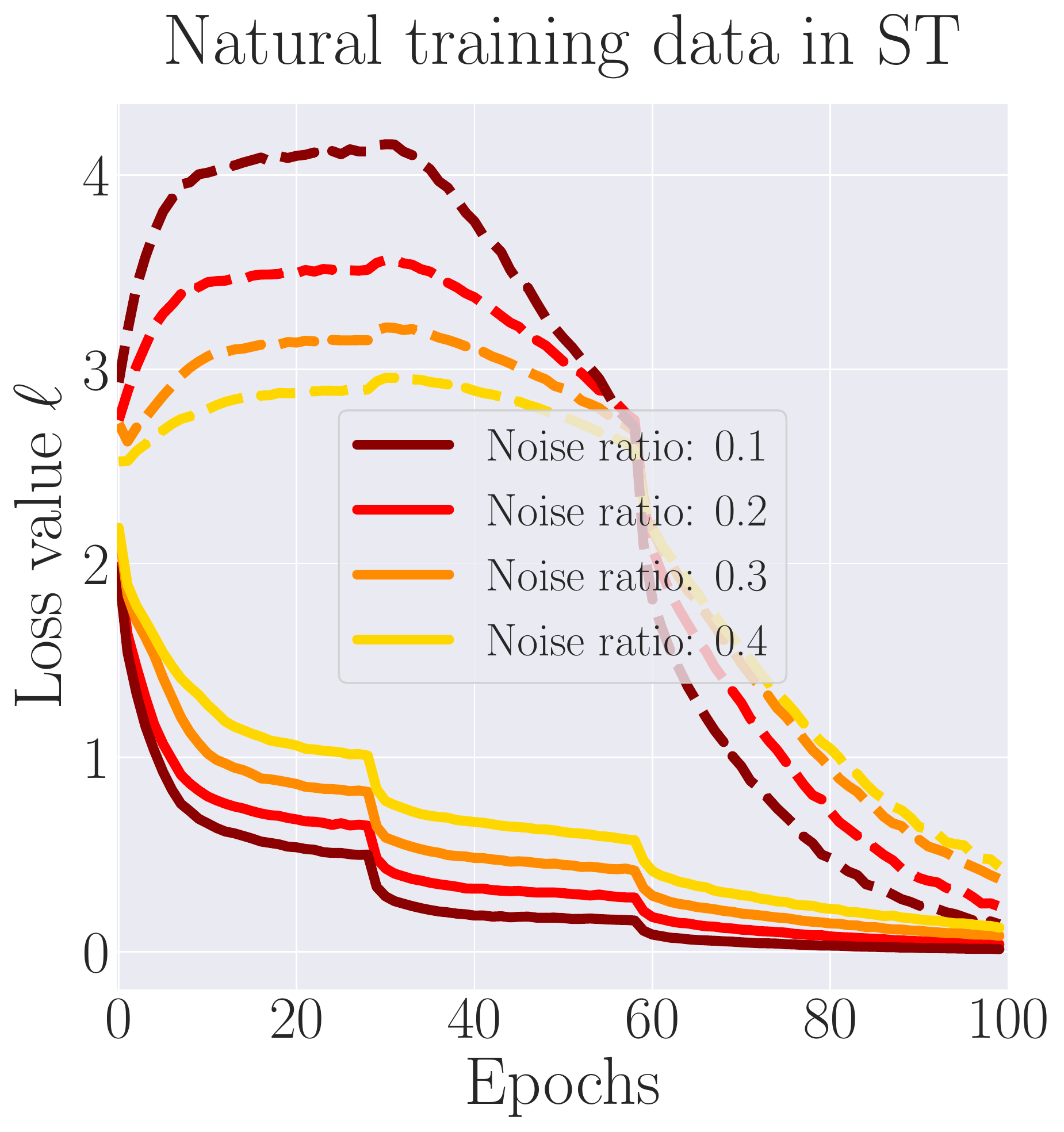}
    \includegraphics[scale=0.195]{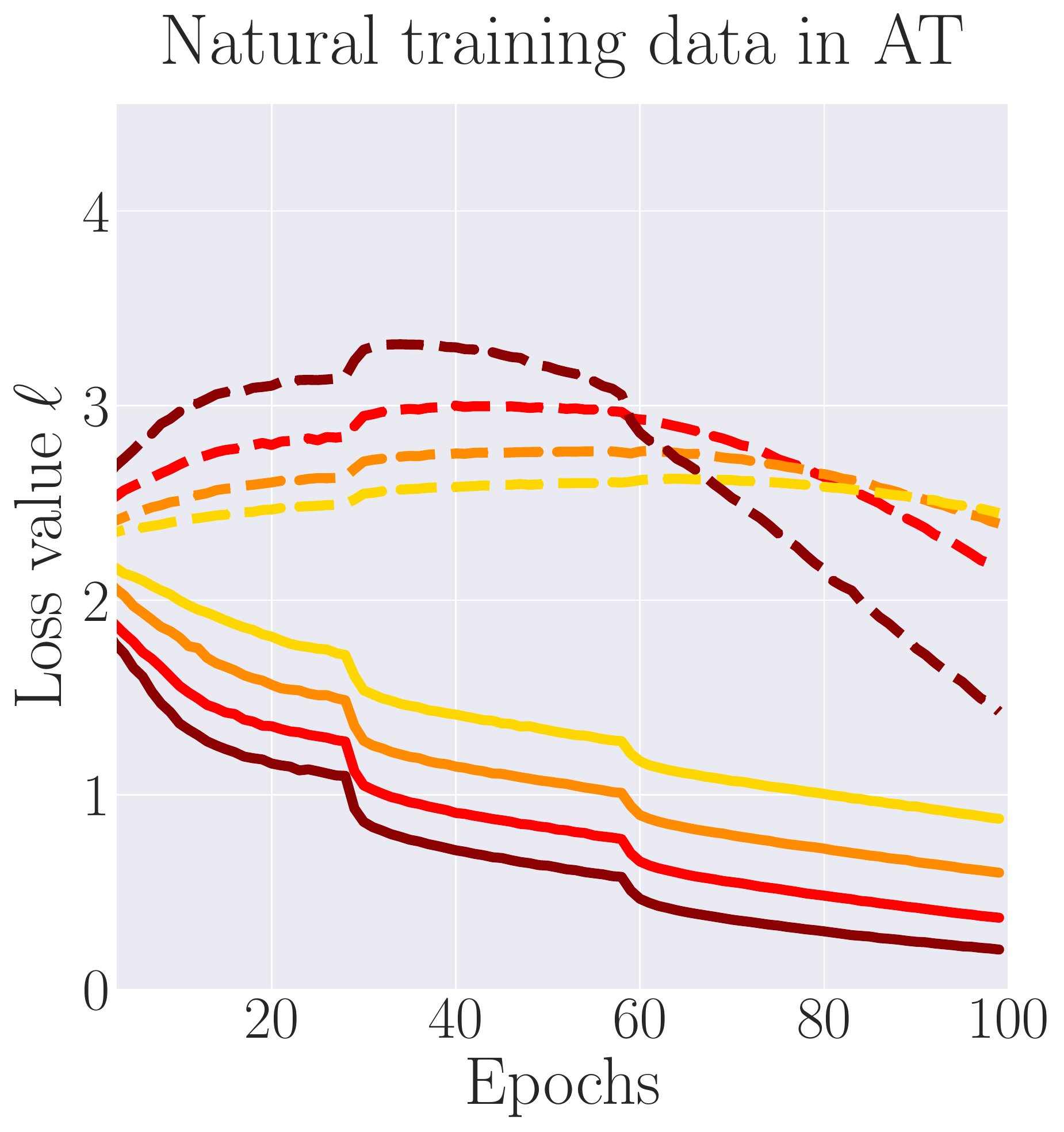}
    }
    \subfigure[MNIST]{
    \includegraphics[scale=0.195]{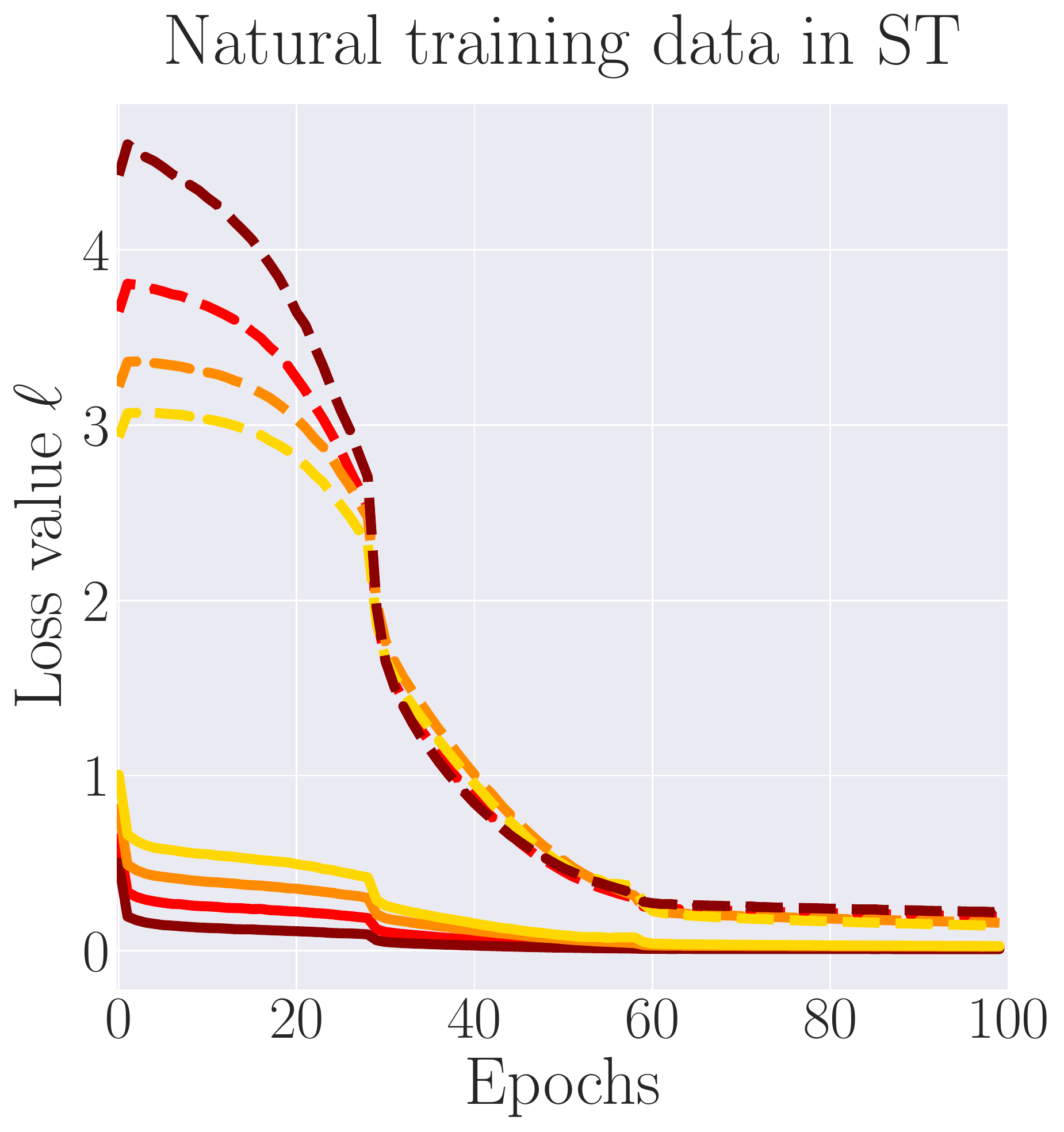}
    \includegraphics[scale=0.195]{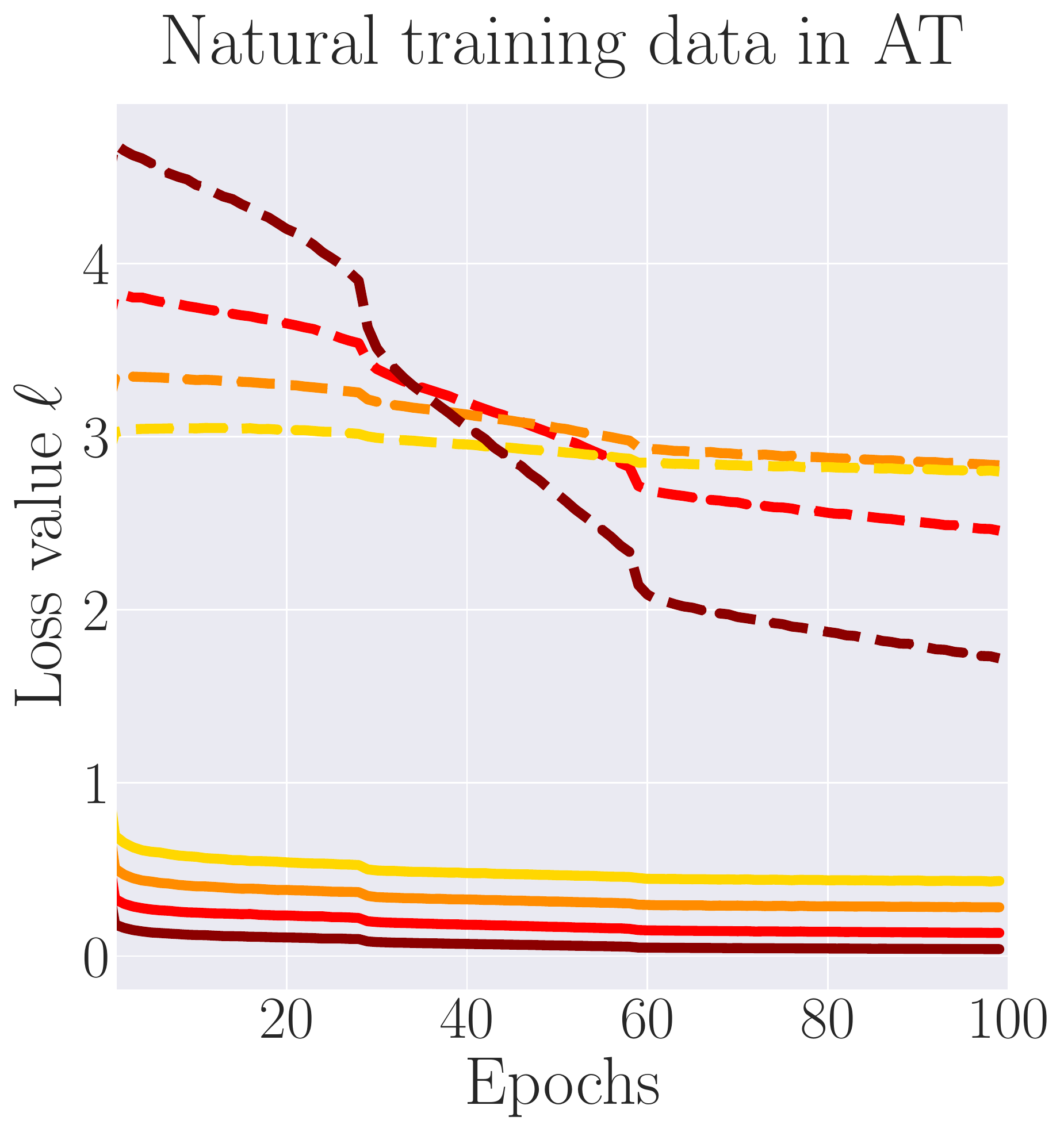}
    }
    \vspace{-2mm}
    \caption{The loss value of ST and AT on correct/incorrect training data using \textit{CIFAR-10} and \textit{MNIST} with \textit{symmetric-flipping} noise. Solid lines denote the loss value of correct training data, while dashed lines correspond to that of incorrect training data. Compared with ST, there is a large gap in the loss value of correct/incorrect training data in AT.}
    %\vspace{2mm}
    \label{fig:appendix_part1_diff_st_at_label_noise_train_loss_sym}
\end{figure*}

\begin{figure*}[h!]
\vspace{4mm}
    \centering
    \hspace{1mm}
    \subfigure[CIFAR-10]{
    \includegraphics[scale=0.195]{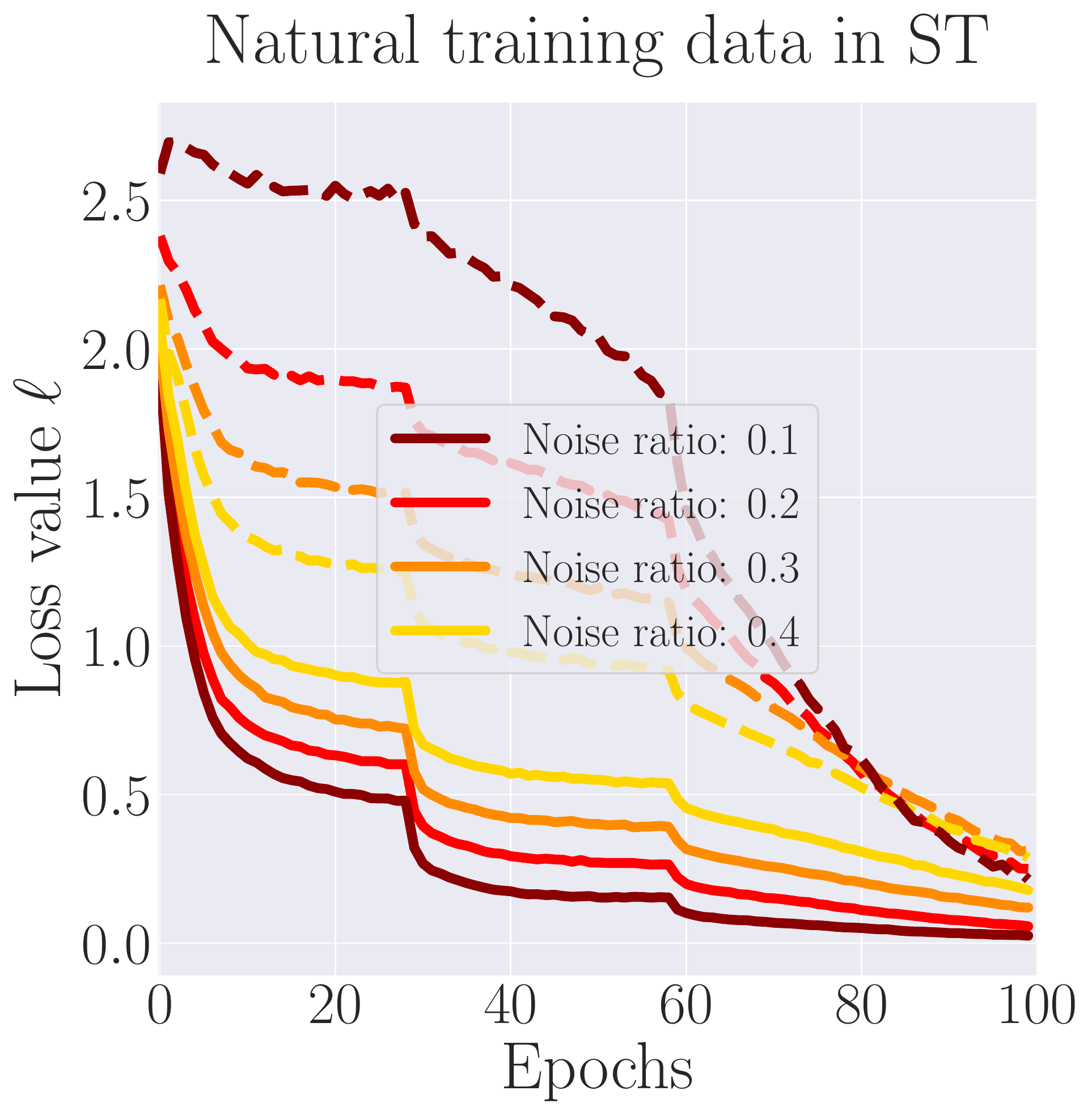}
    \includegraphics[scale=0.195]{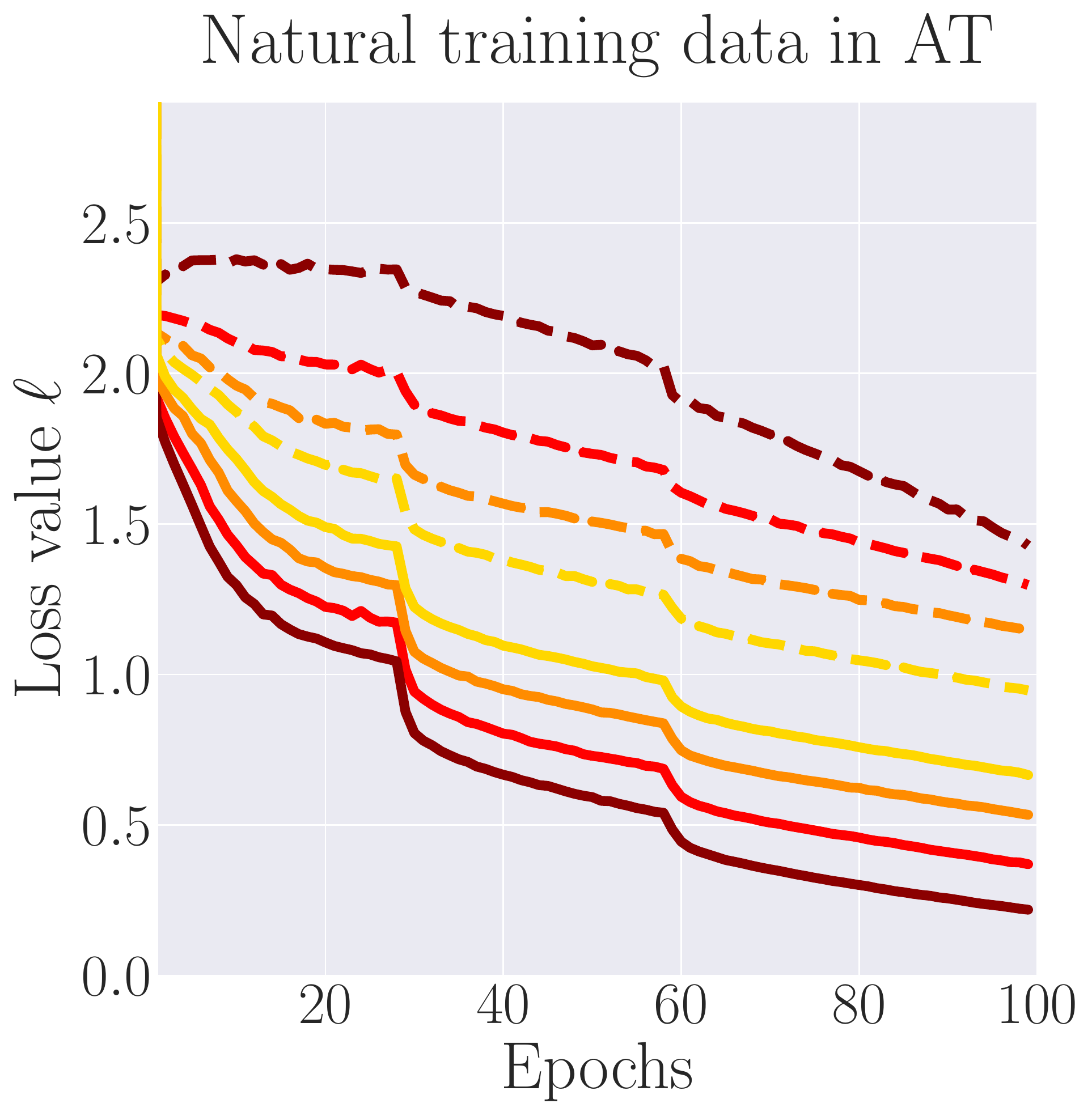}
    }
    \subfigure[MNIST]{
    \includegraphics[scale=0.195]{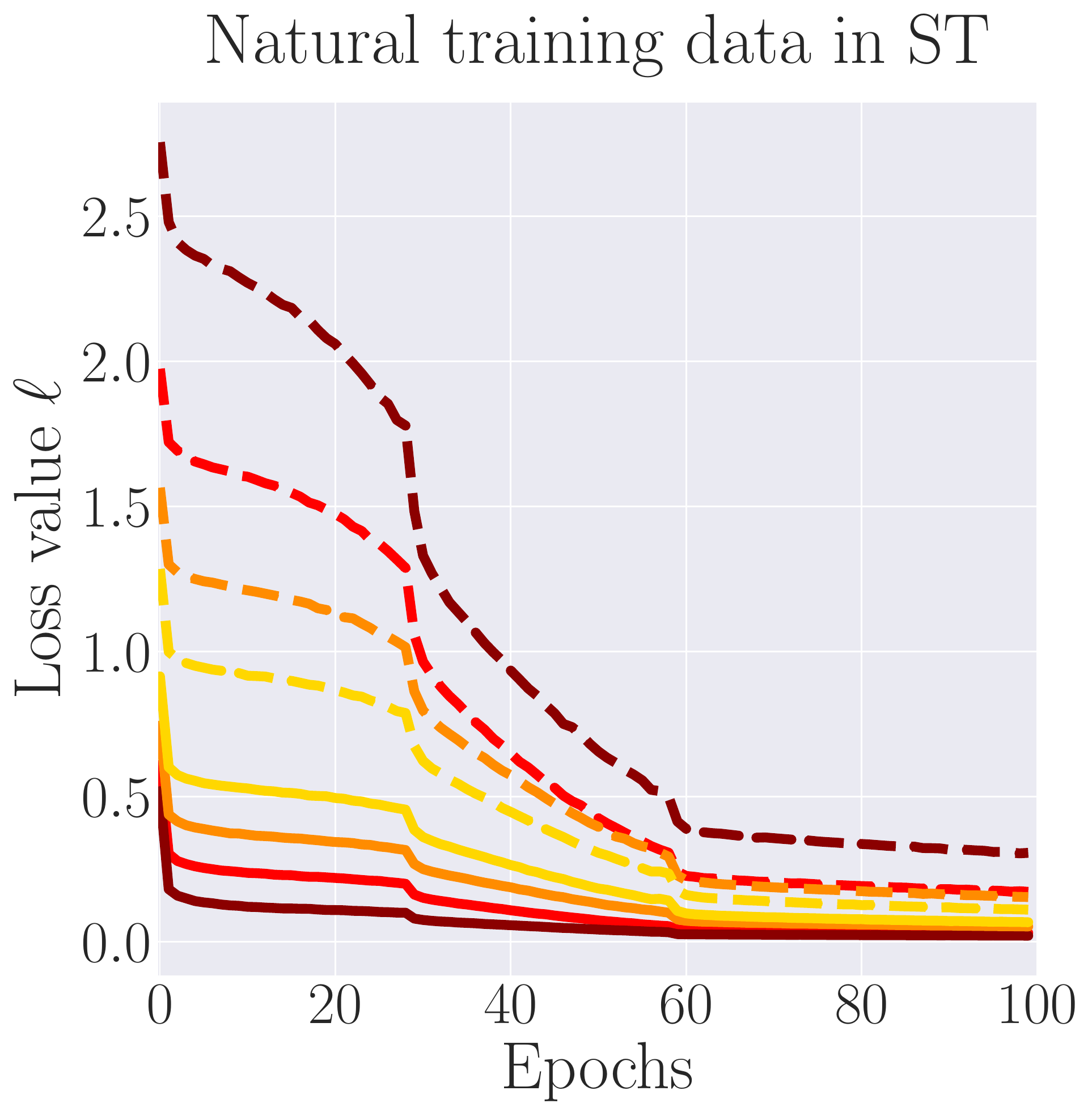}
    \includegraphics[scale=0.195]{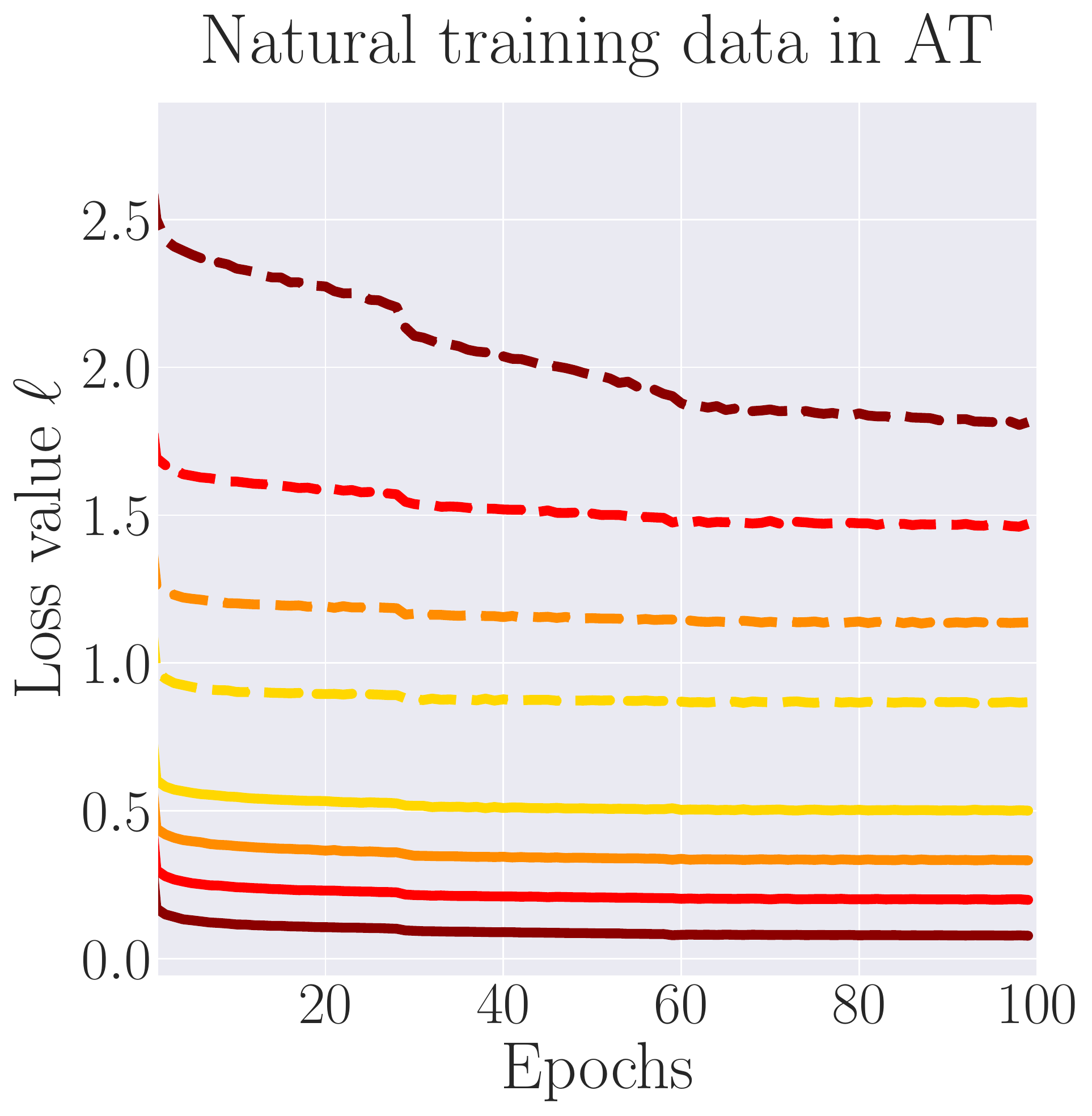}
    }
    \vspace{-2mm}
    \caption{The loss value of ST and AT on correct/incorrect training data using \textit{CIFAR-10} and \textit{MNIST} with \textit{pair-flipping} noise. Solid lines denote the loss value of correct training data, while dashed lines correspond to that of incorrect training data. Compared with ST, there is a large gap in the loss value of correct/incorrect training data in AT.}
    %\vspace{2mm}
    \label{fig:appendix_part1_diff_st_at_label_noise_train_loss_asym}
\end{figure*}

\paragraph{Result.}
In Figures~\ref{fig:appendix_part1_diff_st_at_label_noise_train_loss_sym} and~\ref{fig:appendix_part1_diff_st_at_label_noise_train_loss_asym}, we check the loss value of correct/incorrect training data with different noise rates and types. On the whole, compared with ST, correct/incorrect training data can also be more distinguishable in AT using the loss value, regardless of noise rates and types.

\clearpage

\section{New Measure: Geometry Value $\kappa$}
\label{appendix:new_measurement}

In this section, we provide more experimental results of the geometry value $\kappa$ vs. that of the loss value, and provide more visualization about the specific semantic information corresponds to our new measure. First, we calculate the loss value and geometry value $\kappa$ of correct/incorrect data in AT with different noise rates and types (Appendix~\ref{sec:app_kappa_loss}). Second, we display more visualization results on \textit{CIFAR-10} and \textit{MNIST} datasets to show the relationship between the geometry value $\kappa$ and image data (Appendix~\ref{sec:app_rare_classic}).

%In this section, we showed the geometry value $\kappa$ could be a new measurement for data stratification. First, the geometry value $\kappa$ can differentiate correct/incorrect data in AT (Figures~\ref{fig:loss_value_pgd_steps_AT},~\ref{fig:dis_pgd_loss_sym_02}  and~\ref{fig:dis_pgd_loss_asym_04}). Compare with loss value, which is wildly used in sample selection~\citep{jiang2018mentornet,han2018co,yu2019does}, we showed that the geometry value $\kappa$ can have a better performance to filter incorrect data with different noise types. Second, we demonstrated that the geometry value $\kappa$ can provide a finer stratification on classic/rare data (Figures~\ref{fig:resnet18_AT_30_PGD_loss}, \ref{fig:mnist_rare_pgd_steps}, and~\ref{fig:cifar10_rare_pgd_steps}).

\subsection{Geometry Value vs. Loss Value}
\label{sec:app_kappa_loss}

%Under different noise ratio of the training data, the difference between correct and incorrect data in the term of the mean value calculated on the $\kappa$ is greater than that of the loss value. Many previous studies of label noise utilize the loss value to do sample selection~\citep{jiang2017mentornet, han2018co}. With the similar distinguishing ability, the geometry value $\kappa$ can also be used as a new indicator of sample selection for training in the presence of label noise.

\begin{figure}[h!]
\vspace{-2mm}
    \centering
    \subfigure[Noise rate: 0.1]{
    \begin{minipage}[b]{0.465\linewidth}
        \includegraphics[scale=0.19]{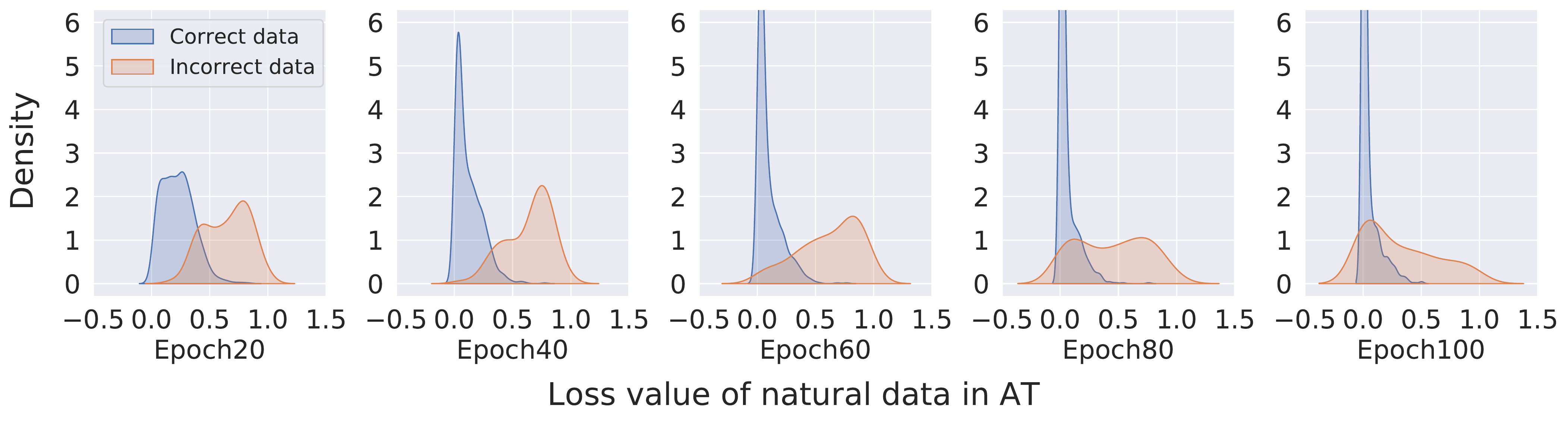}\\
        \includegraphics[scale=0.19]{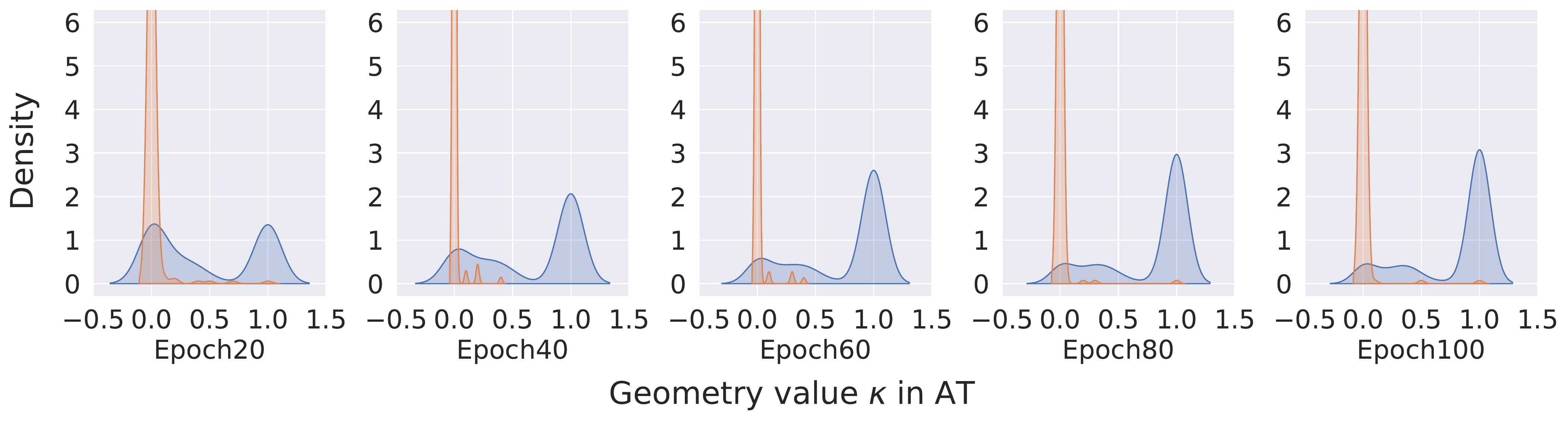}
    \end{minipage}
    }
    \subfigure[Noise rate: 0.2]{
    \begin{minipage}[b]{0.465\linewidth}
        \includegraphics[scale=0.19]{dis_AT_loss_sym_02.pdf}\\
        \includegraphics[scale=0.19]{dis_AT_pgd_sym_02.pdf}
    \end{minipage}
    }\\
    \subfigure[Noise rate: 0.3]{
    \begin{minipage}[b]{0.465\linewidth}
        \includegraphics[scale=0.19]{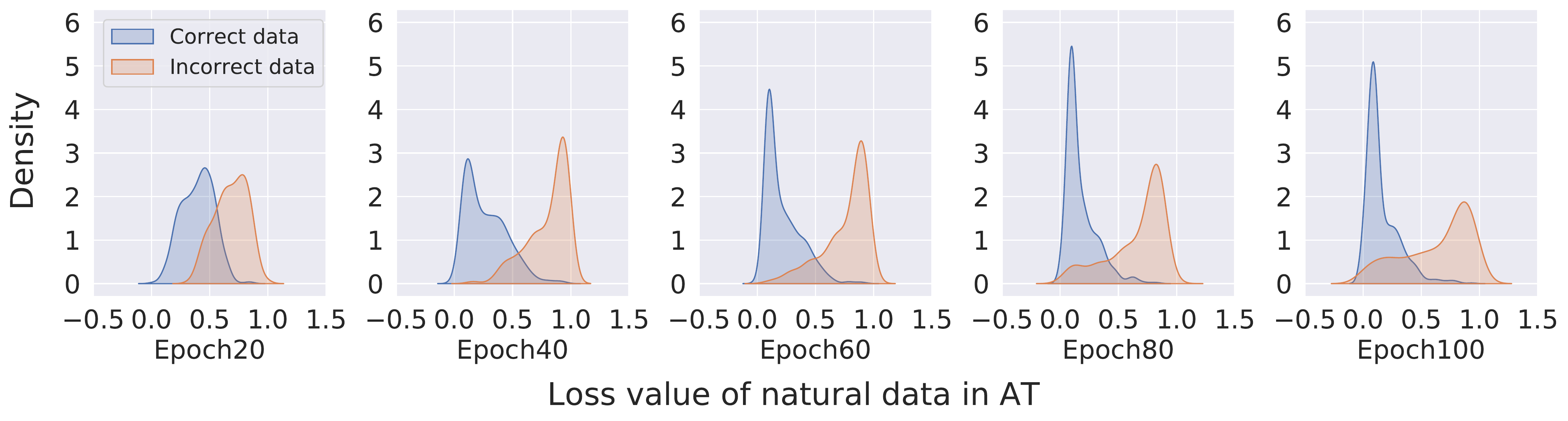}\\
        \includegraphics[scale=0.19]{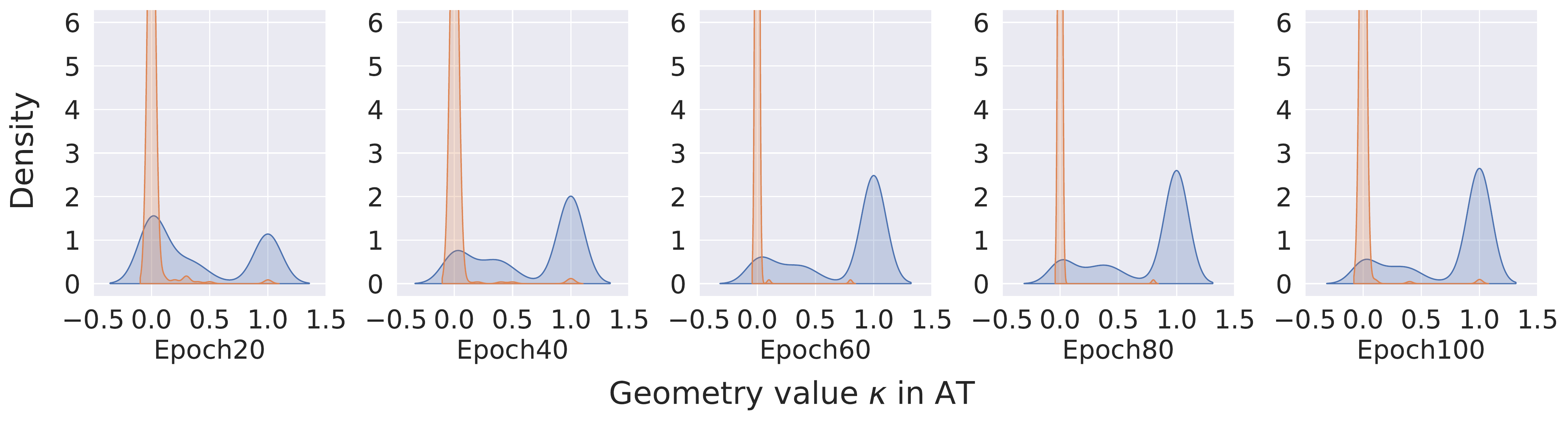}
    \end{minipage}
    }
    \subfigure[Noise rate: 0.4]{
    \begin{minipage}[b]{0.465\linewidth}
        \includegraphics[scale=0.19]{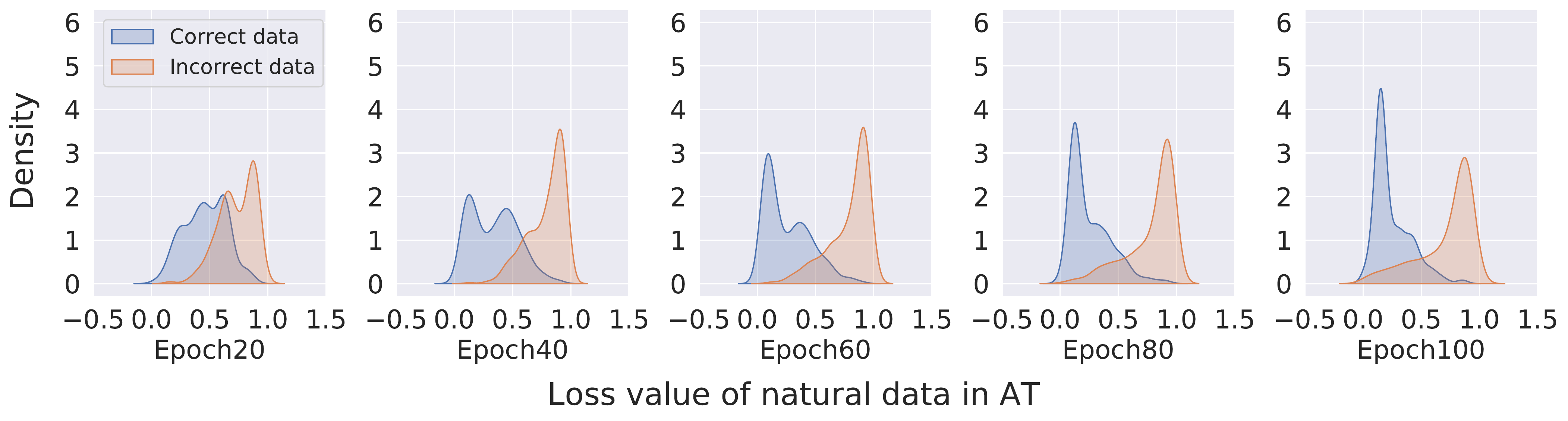}\\
        \includegraphics[scale=0.19]{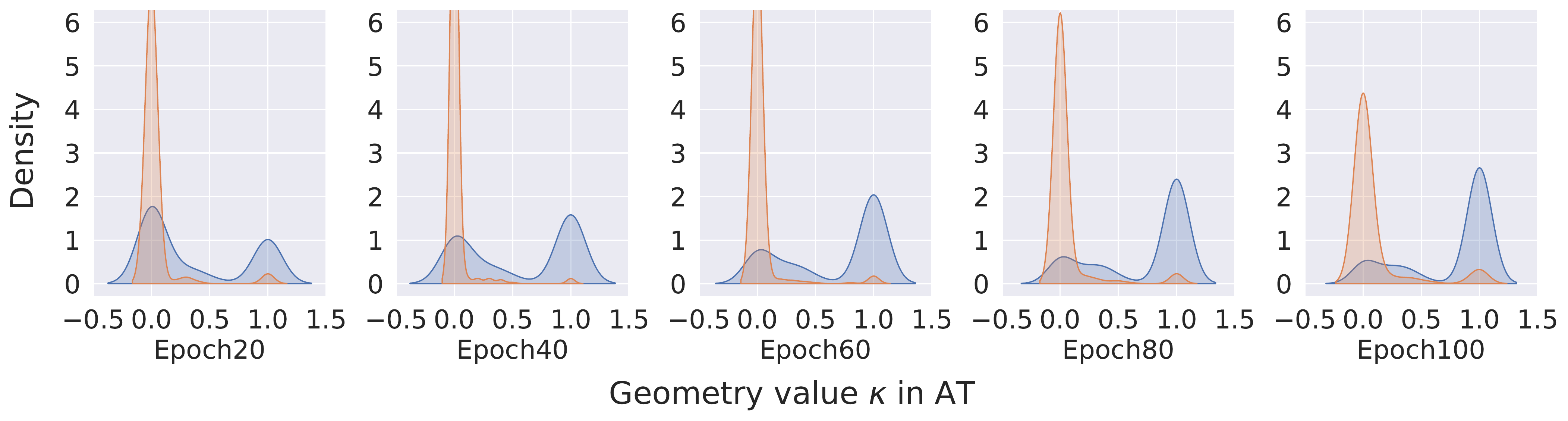}
    \end{minipage}
    }
    \vspace{-2mm}
    \caption{The density of AT on correct/incorrect data using \textit{CIFAR-10} with \textit{symmetric-flipping} noise. \textit{Top panels}: the loss value in AT. \textit{Bottom panels}: the geometry value $\kappa$ in AT. Note that the geometry value $\kappa$ has a better distinction on correct/incorrect data.}
    %\vspace{-5mm}
    \label{fig:appendix_dis_pgd_loss_sym}
\end{figure}

\begin{figure}[t!]
\vspace{10mm}
    \centering
    \subfigure[Noise rate: 0.1]{
    \begin{minipage}[b]{0.465\linewidth}
        \includegraphics[scale=0.19]{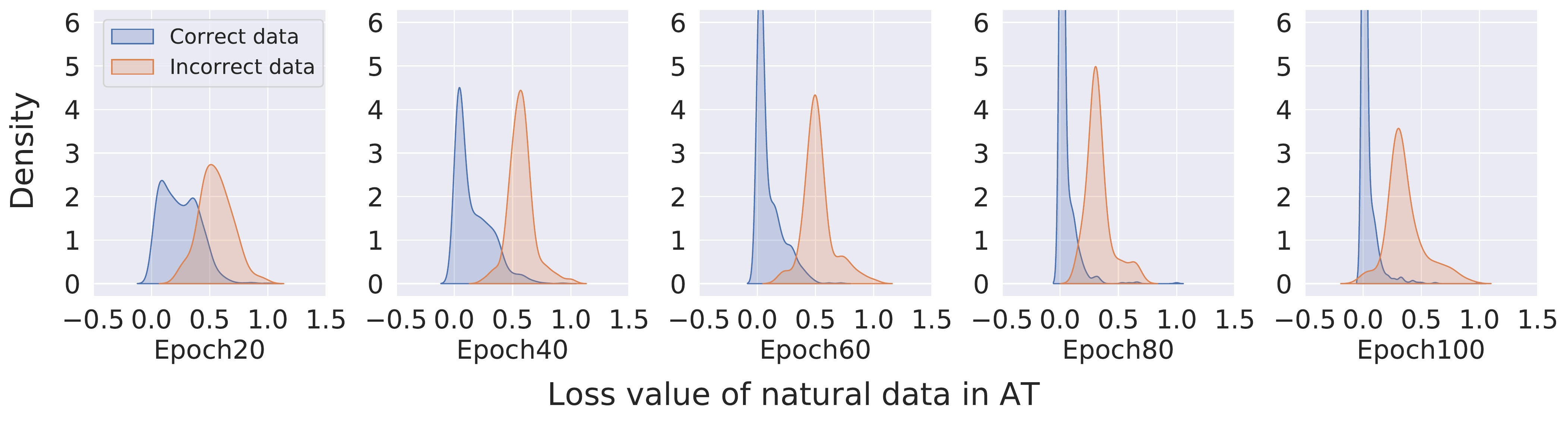}\\
        \includegraphics[scale=0.19]{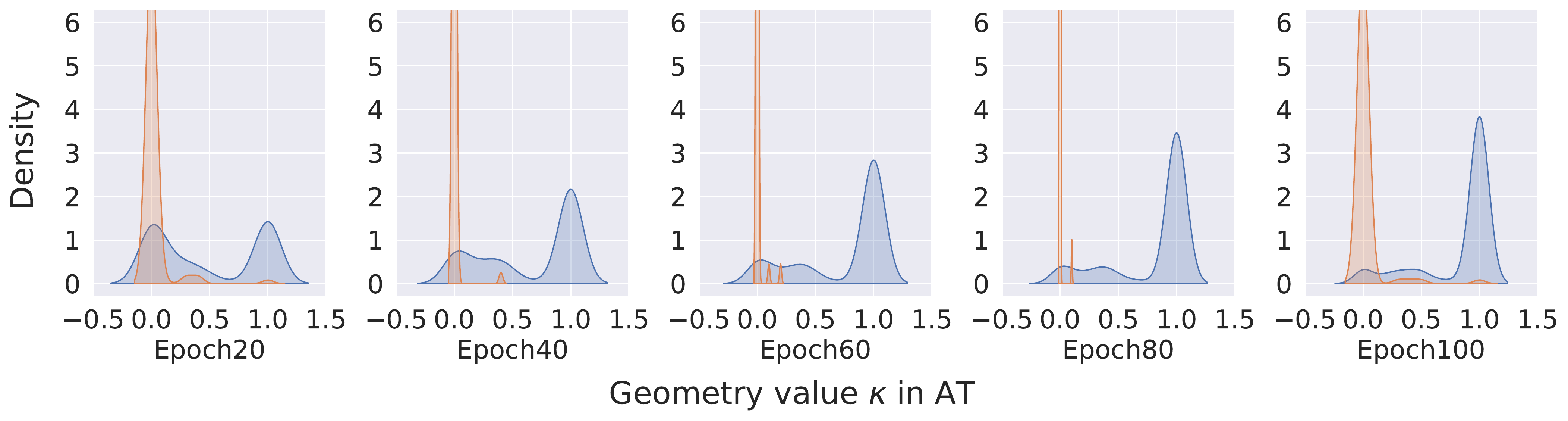}
    \end{minipage}
    }
    \subfigure[Noise rate: 0.2]{
    \begin{minipage}[b]{0.465\linewidth}
        \includegraphics[scale=0.19]{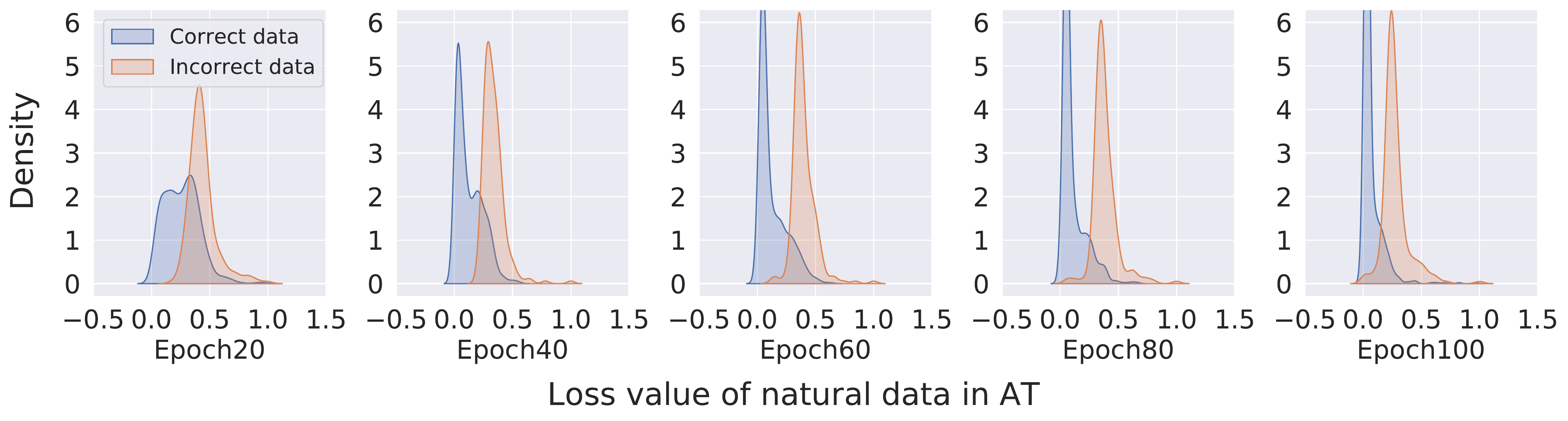}\\
        \includegraphics[scale=0.19]{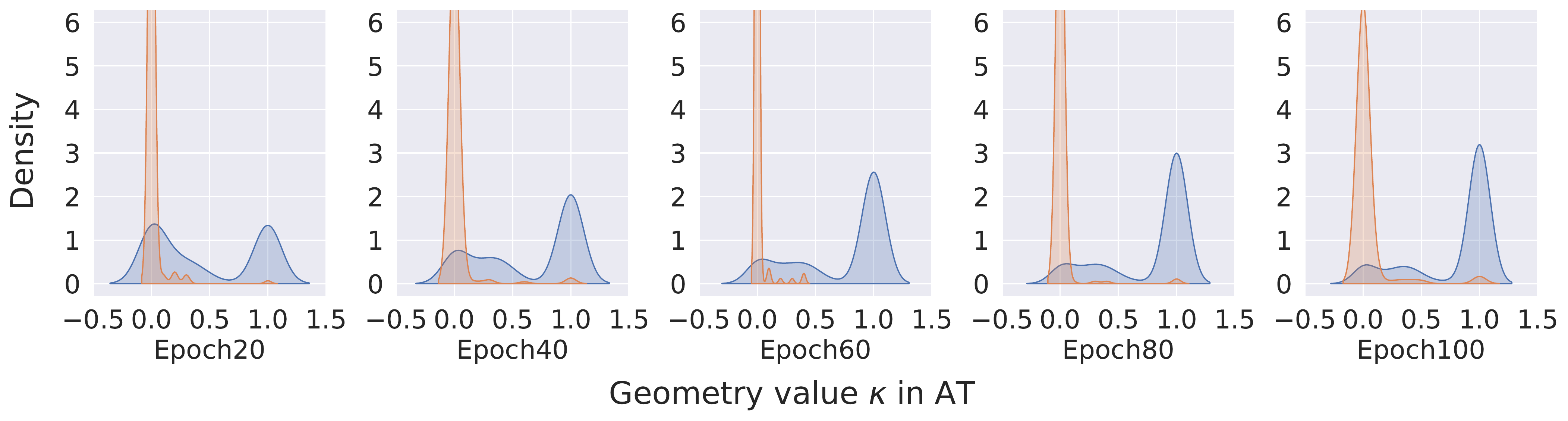}
    \end{minipage}
    }\\
    \subfigure[Noise rate: 0.3]{
    \begin{minipage}[b]{0.465\linewidth}
        \includegraphics[scale=0.19]{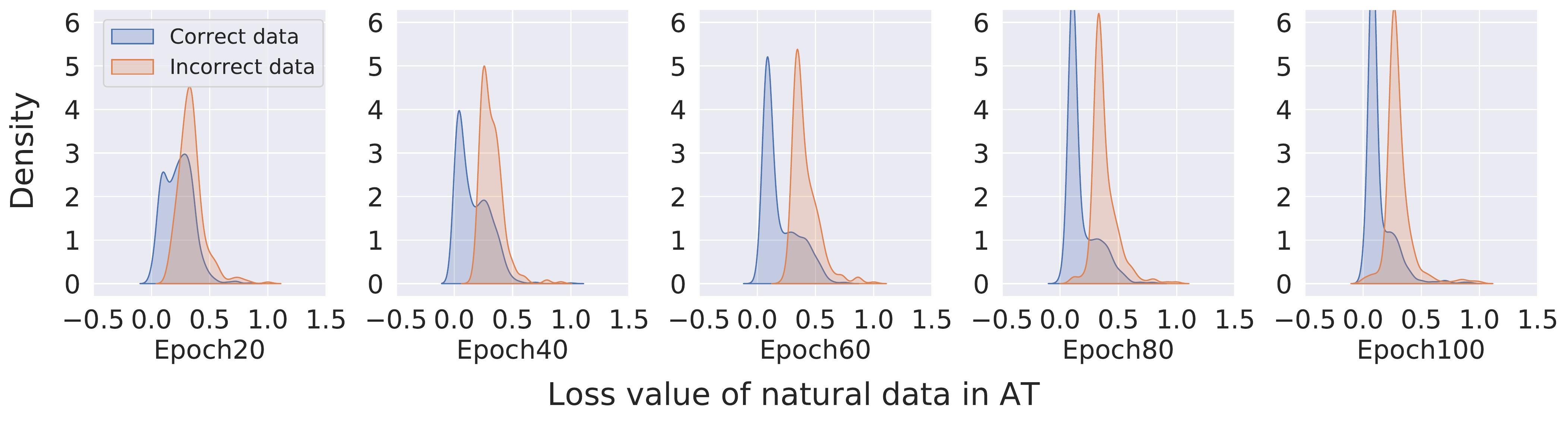}\\
        \includegraphics[scale=0.19]{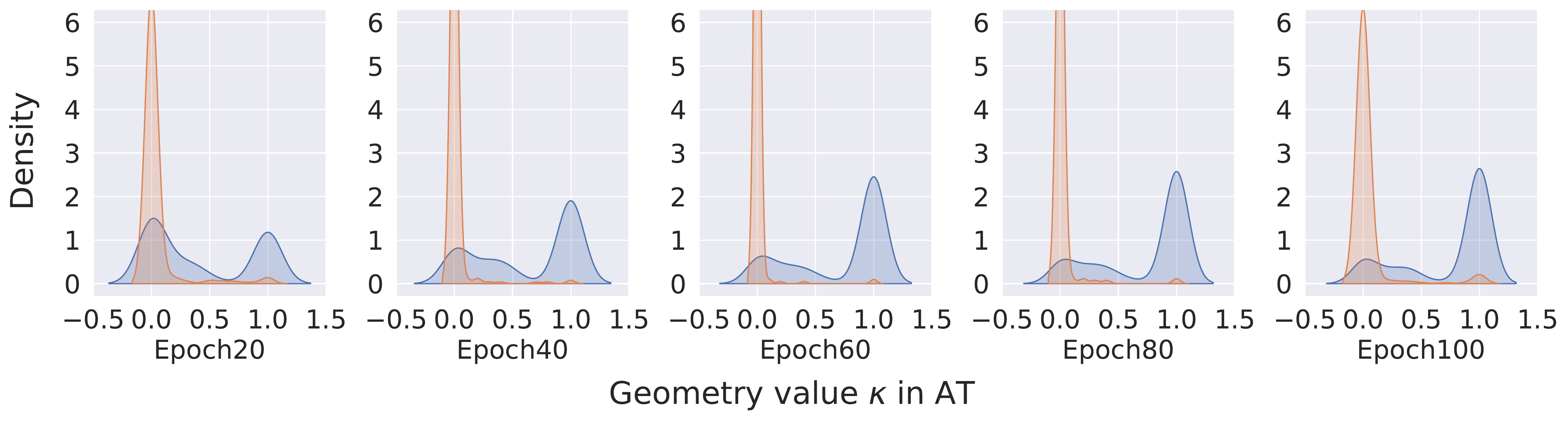}
    \end{minipage}
    }
    \subfigure[Noise rate: 0.4]{
    \begin{minipage}[b]{0.465\linewidth}
        \includegraphics[scale=0.19]{dis_AT_loss_asym_04.pdf}\\
        \includegraphics[scale=0.19]{dis_AT_pgd_asym_04.pdf}
    \end{minipage}
    }
    \vspace{-2mm}
    \caption{The density of AT on correct/incorrect data using \textit{CIFAR-10} with \textit{pair-flipping} noise. \textit{Top panels}: the loss value in AT. \textit{Bottom panels}: the geometry value $\kappa$ in AT. Note that the geometry value $\kappa$ has a better distinction on correct/incorrect data.}
    %\vspace{-5mm}
    \label{fig:appendix_dis_pgd_loss_asym}
\end{figure}

\paragraph{Result.} In Figures~\ref{fig:appendix_dis_pgd_loss_sym} and~\ref{fig:appendix_dis_pgd_loss_asym}, we plot the density maps of two measures on \textit{CIFAR-10} dataset with symmetric-flipping and pair-flipping noise. We calculate the loss value of natural data and the geometry value in AT using $5$ checkpoints at different epochs (e.g., Epoch20, Epoch40, Epoch60, Epoch80, Epoch100), which trained with the same settings in Appendix~\ref{sec:app_train_test}. We perform the min-max normalization~\citep{tax2000feature} on both loss value and geometric value $\kappa$, which scales the range of values in $[0,1]$. On the whole, it is clear that the geometry value $\kappa$ has a stable performance of distinguishing correct/incorrect data under different noise rates and types. Specifically, under the pair-flipping noise with the large noise rate (e.g., Noise rate: 0.4), the loss value cannot differentiate correct/incorrect data well, while the geometry value $\kappa$ can still have a satisfied distinguishing performance.

\subsection{Distinguish Rare and Typical Data}
\label{sec:app_rare_classic}

\begin{figure}[h!]
    \vspace{2mm}
    \centering
    \includegraphics[scale=0.43]{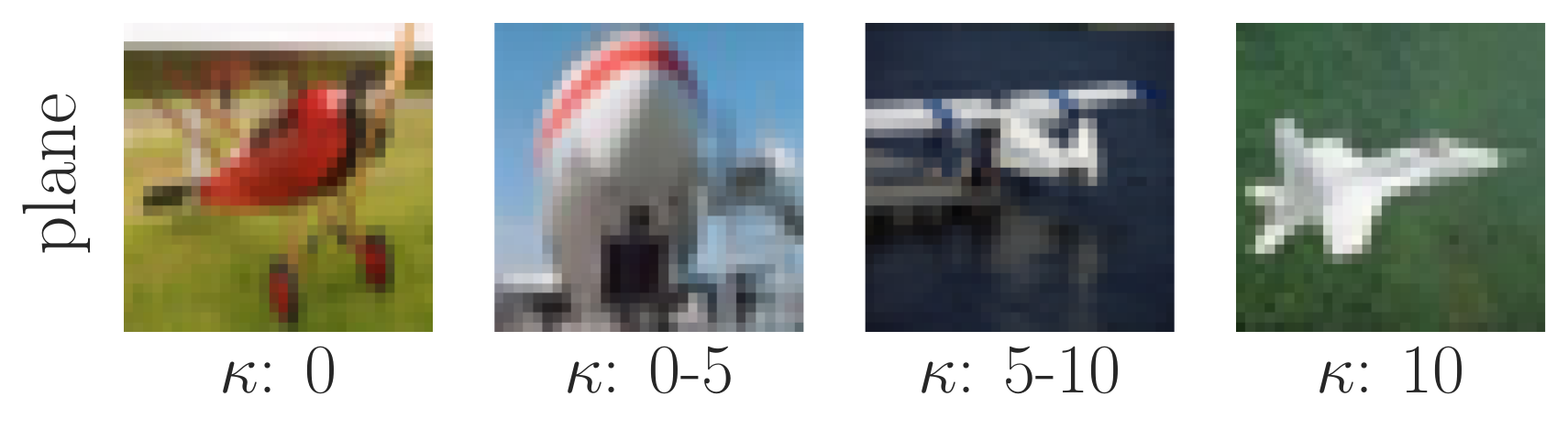}
    \hspace{2mm}
    \includegraphics[scale=0.43]{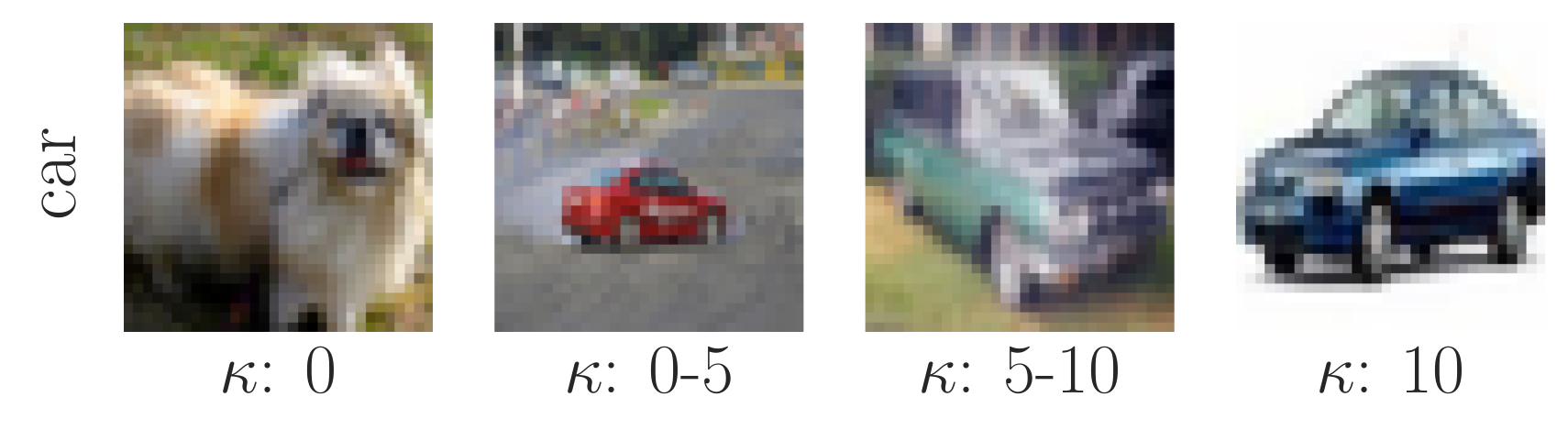}\\
    %\hspace{1mm}
    \includegraphics[scale=0.43]{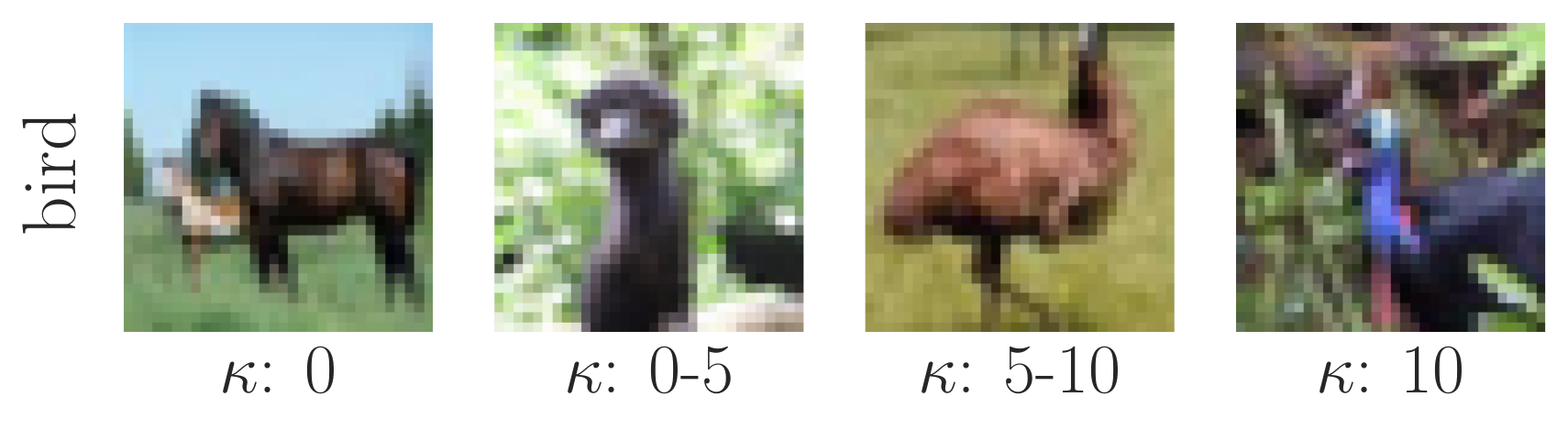}
    \hspace{2mm}
    \includegraphics[scale=0.43]{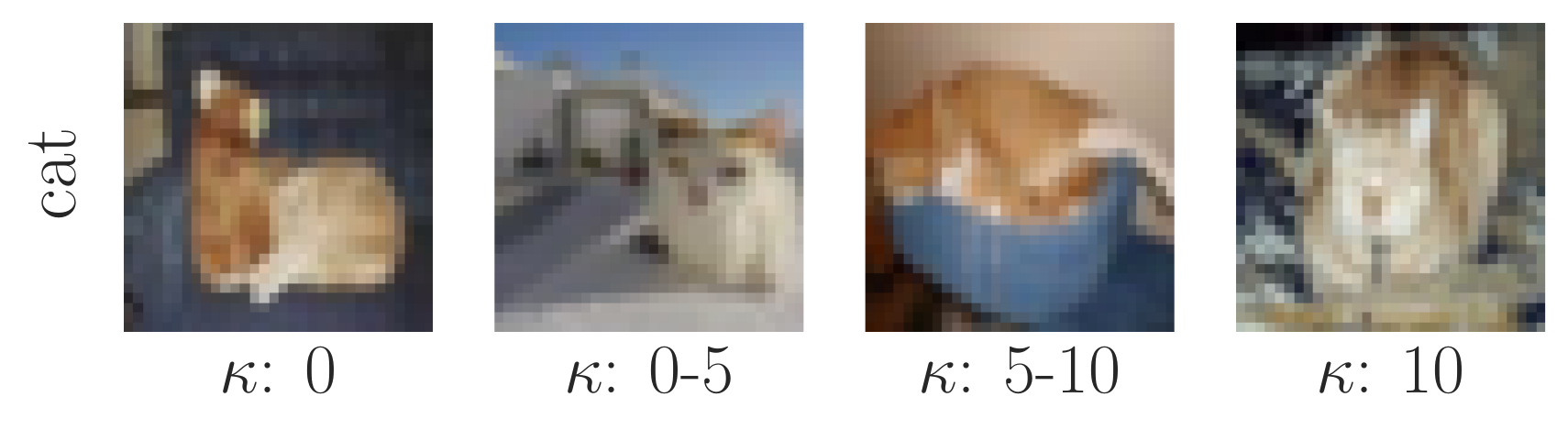}\\
    %\hspace{1mm}
    \includegraphics[scale=0.43]{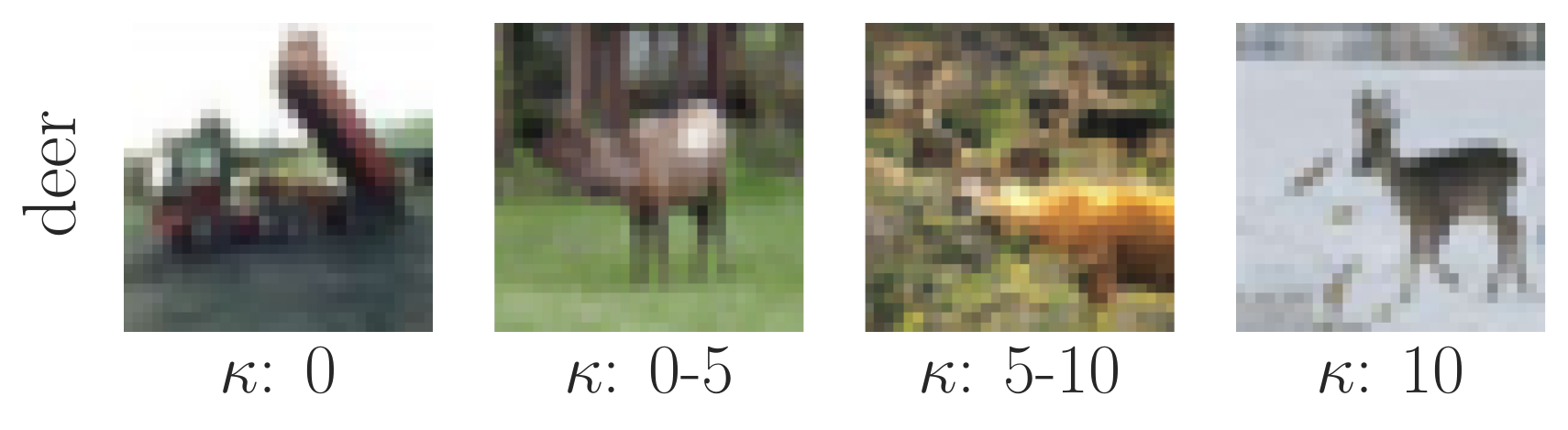}
    \hspace{2mm}
    \includegraphics[scale=0.43]{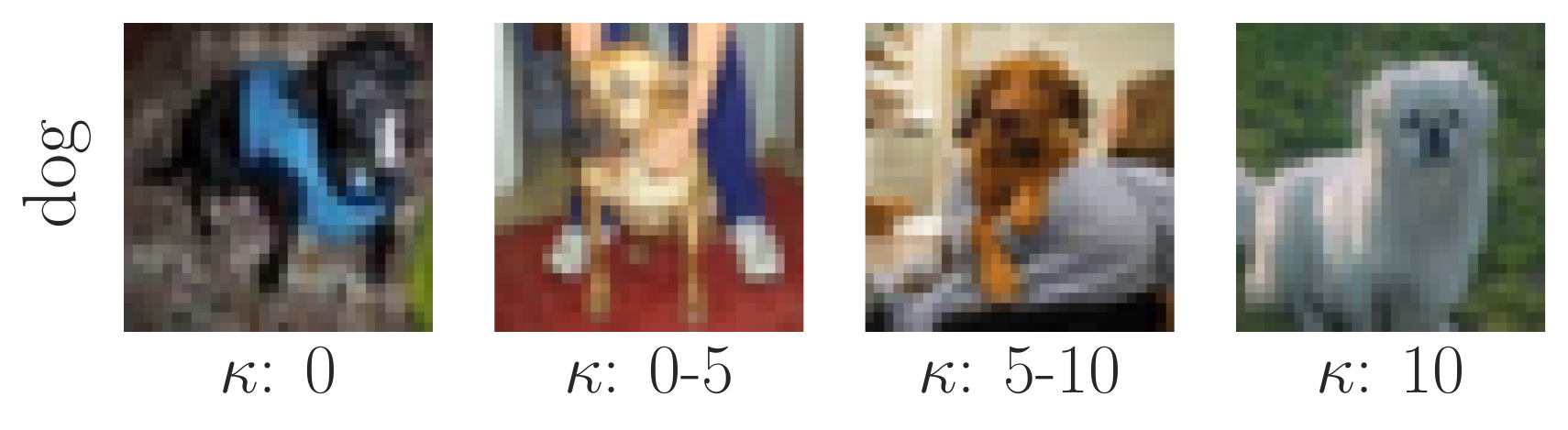}\\
    \includegraphics[scale=0.43]{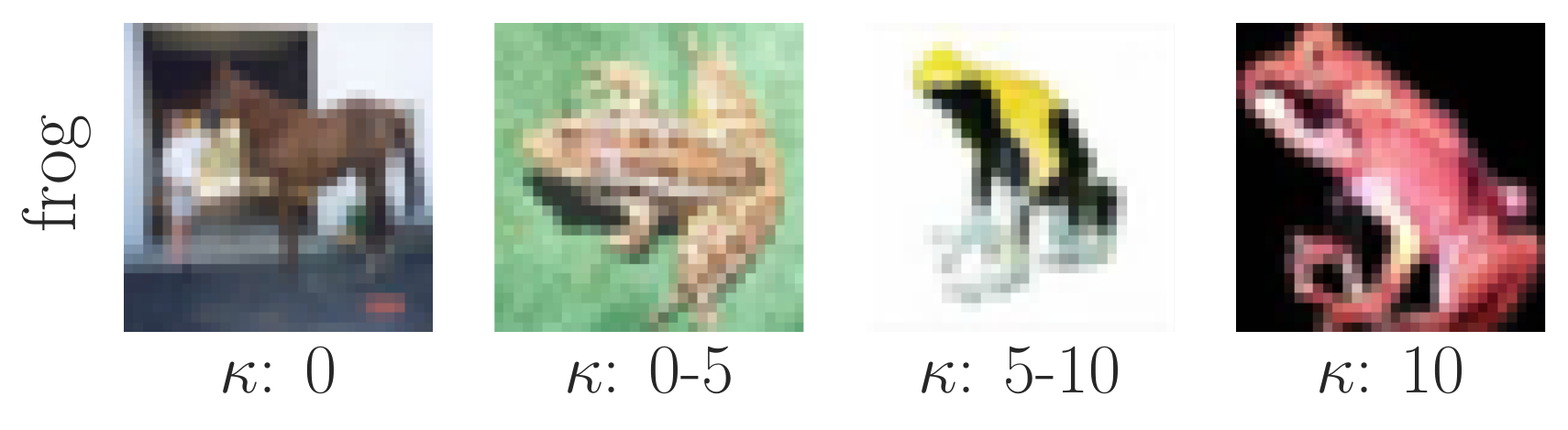}
    \hspace{2mm}
    \includegraphics[scale=0.43]{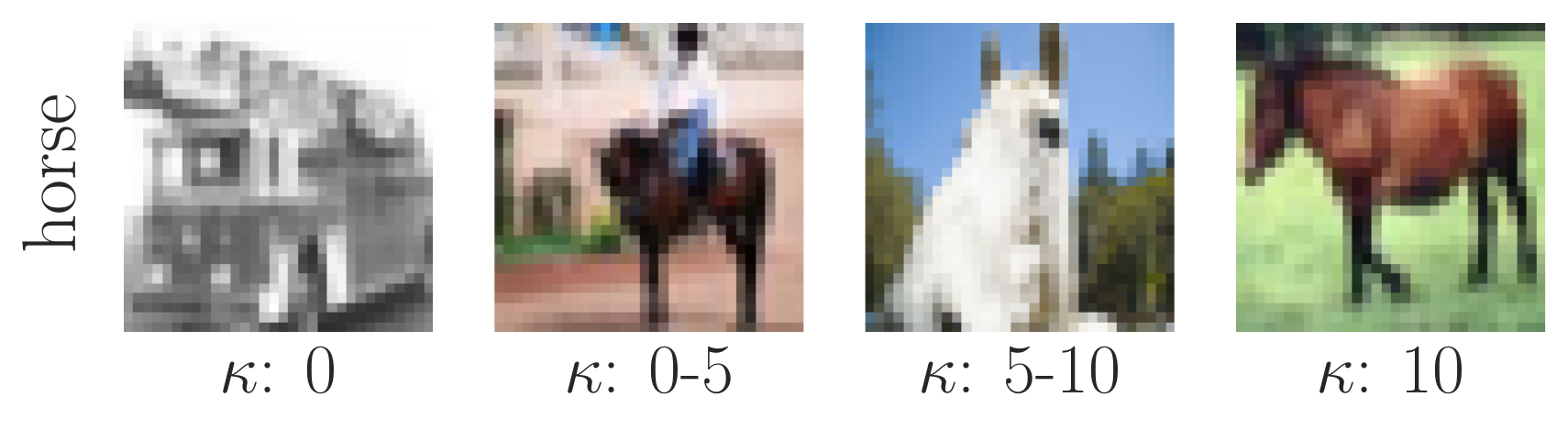}\\
    \includegraphics[scale=0.43]{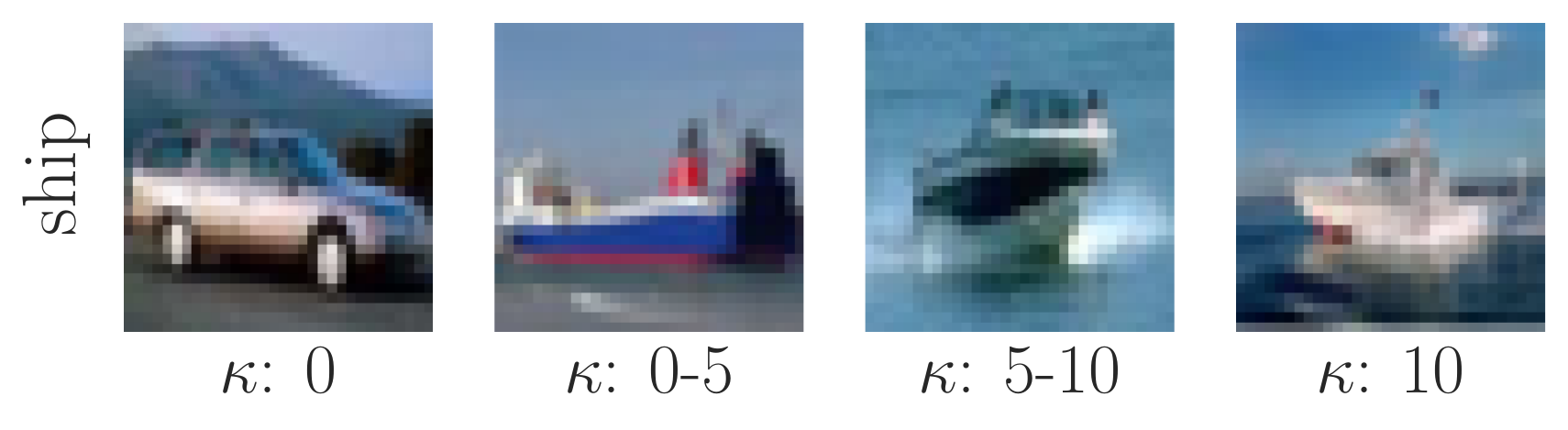}
    \hspace{2mm}
    \includegraphics[scale=0.43]{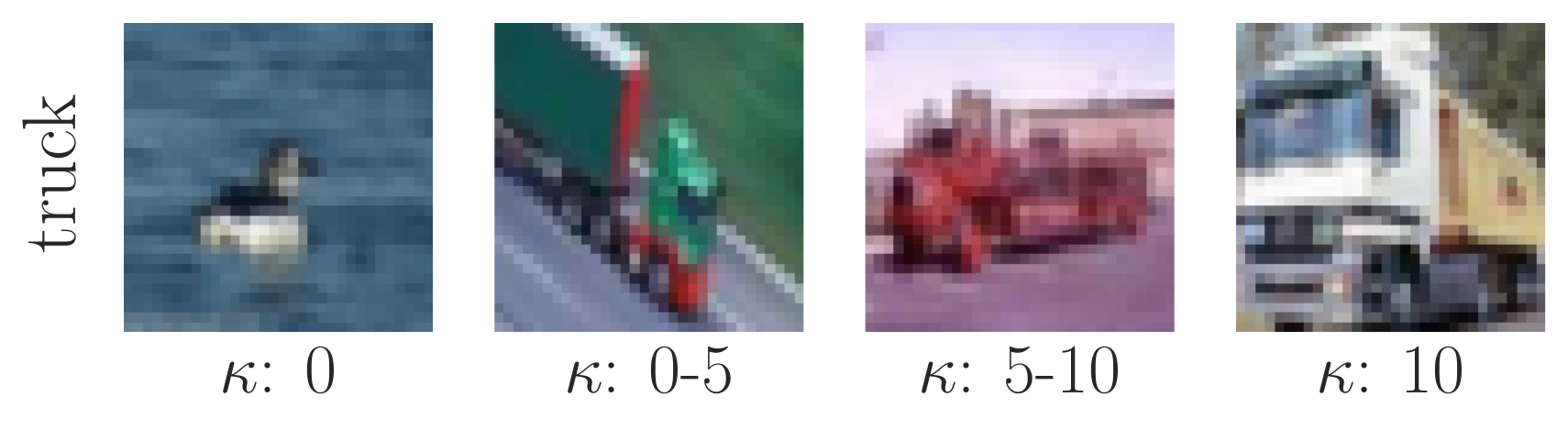}\\

    %\vspace{-3mm}
    \caption{The geometry value $\kappa$ w.r.t. images in \textit{CIFAR-10} with $20\%$ \textit{symmetric-flipping} noise. The leftmost is the given label of all images on the right. We randomly selected four examples with the different $\kappa$ ($\kappa=0$, $\kappa \in (0,5)$, $\kappa \in (5,10)$,  $\kappa=10$) in each class. As the geometric value $\kappa$ increases from left ($\kappa = 0$) to right ($\kappa = 10$), the semantic information of images is more typical and recognizable.}
    %\vspace{-4mm}
    \label{fig:appendix_cifar10_rare_pgd_steps}
\end{figure}

\begin{figure}[h!]
%\vspace{-2mm}
    \centering
    \includegraphics[scale=0.43]{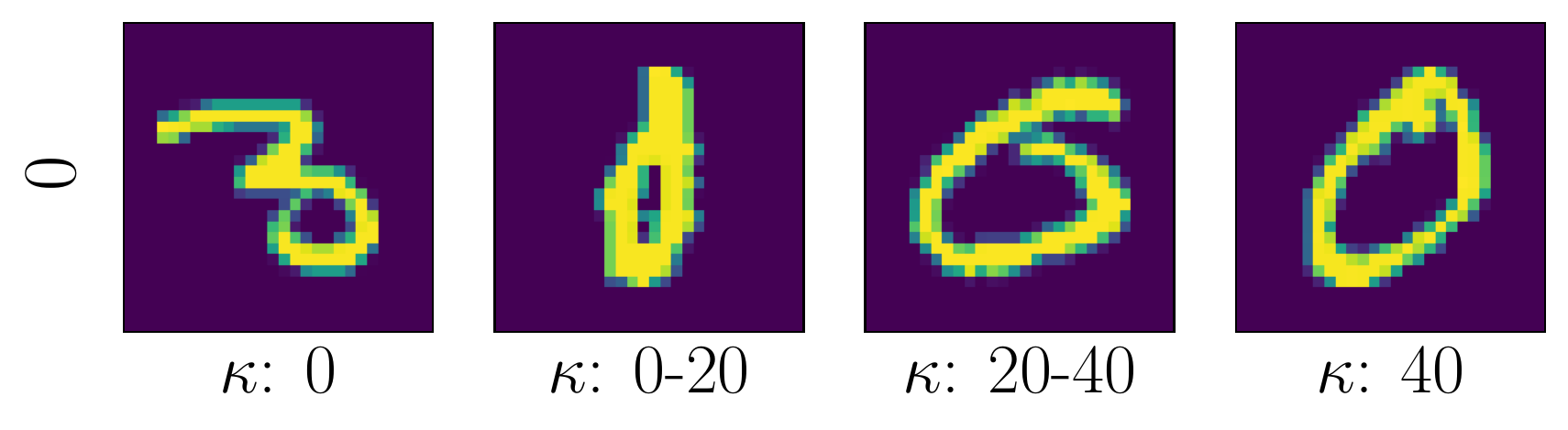}
    \hspace{2mm}
    \includegraphics[scale=0.43]{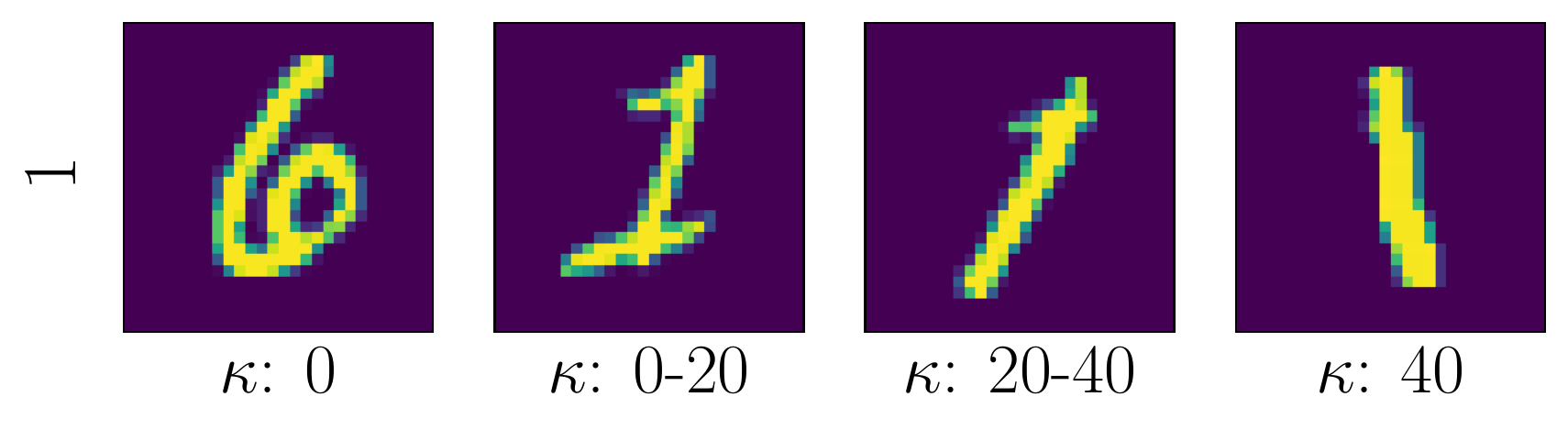}\\
    %\hspace{1mm}
    \includegraphics[scale=0.43]{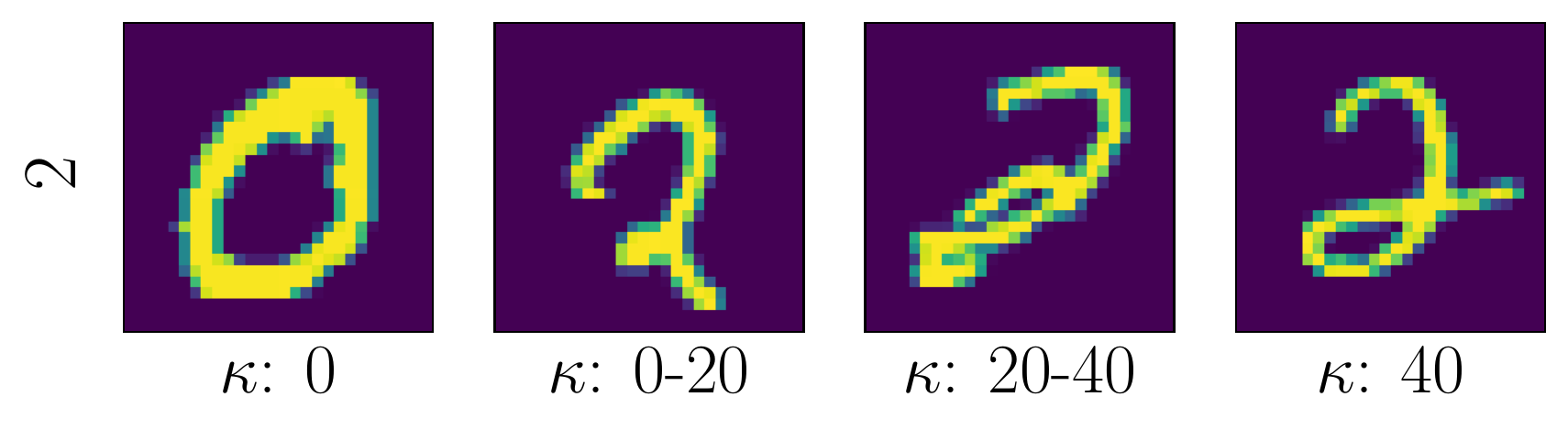}
    \hspace{2mm}
    \includegraphics[scale=0.43]{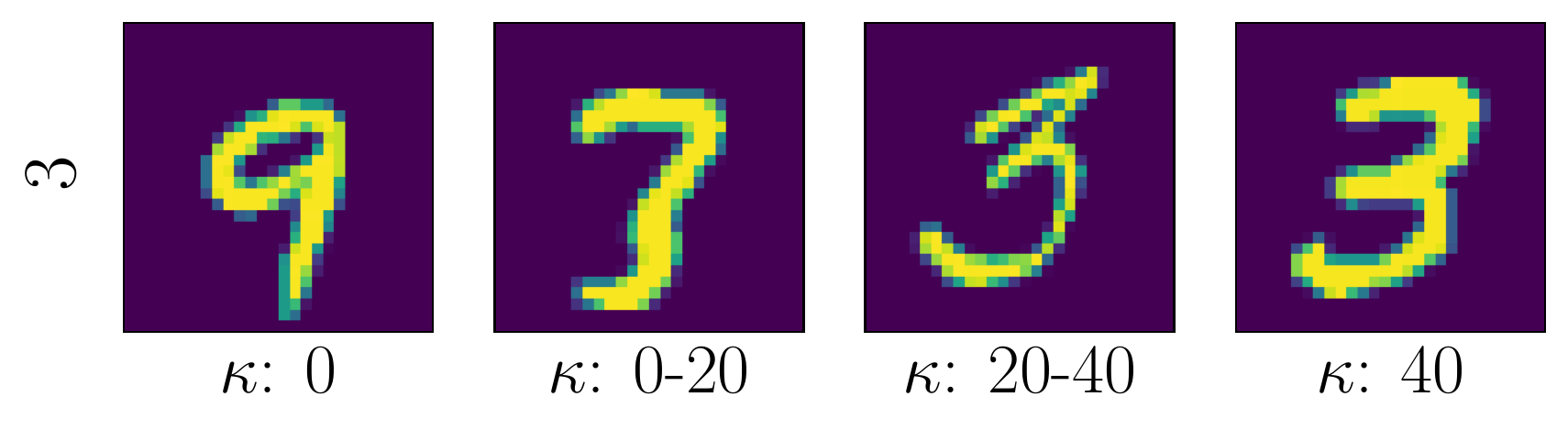}\\
    %\hspace{1mm}
    \includegraphics[scale=0.43]{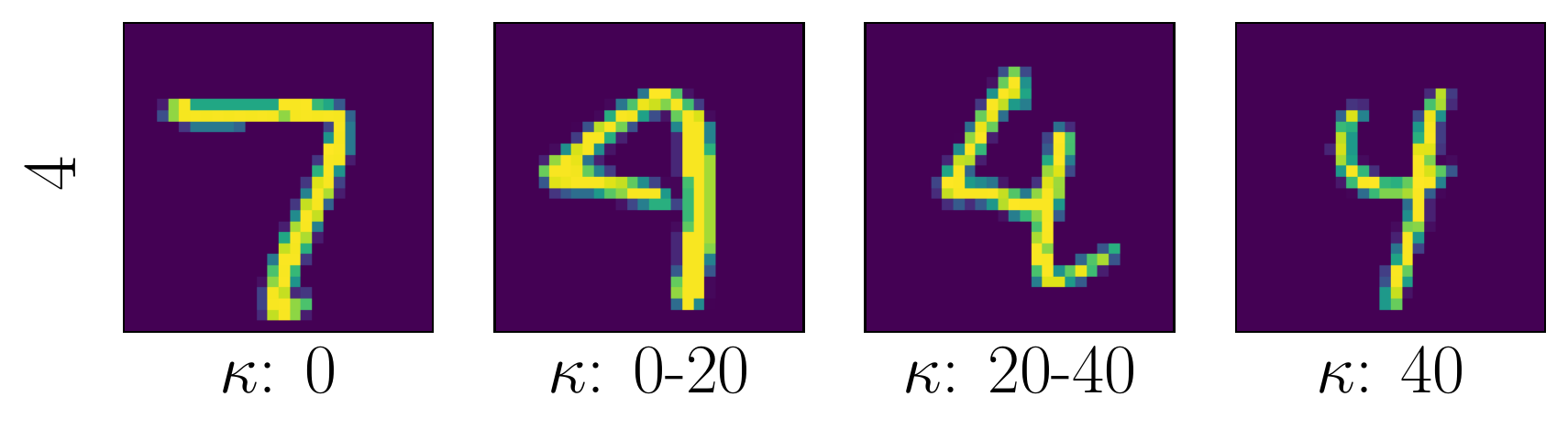}
    \hspace{2mm}
    \includegraphics[scale=0.43]{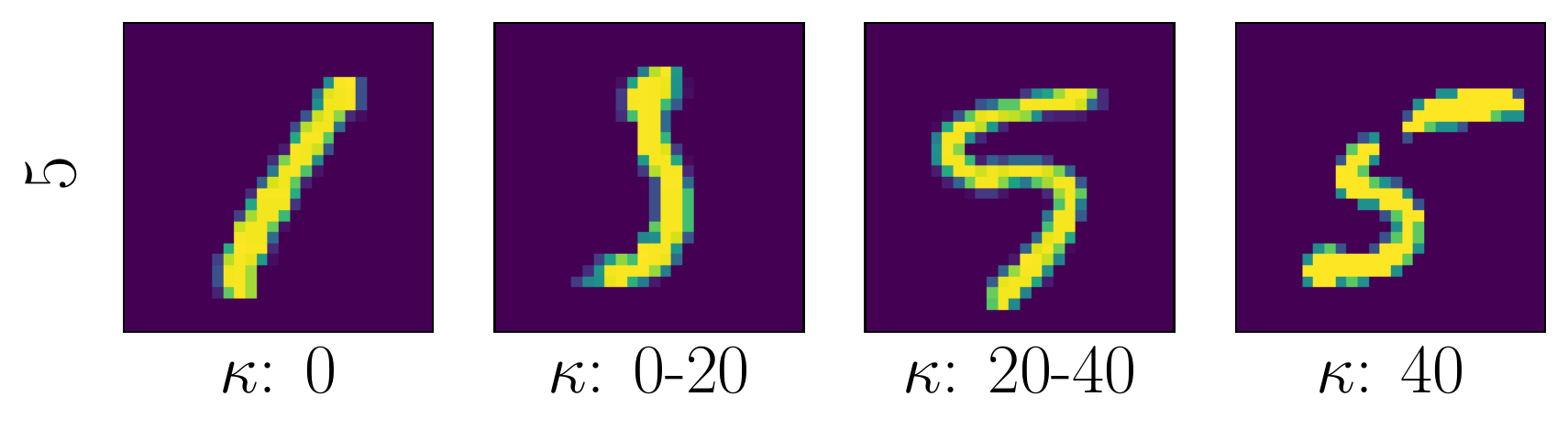}\\
    \includegraphics[scale=0.43]{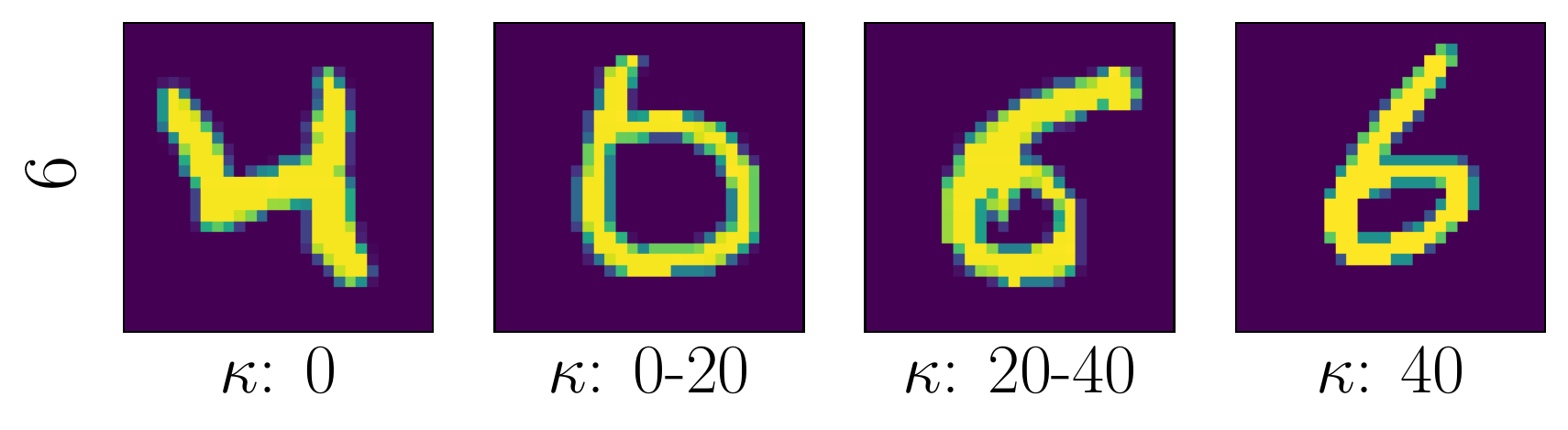}
    \hspace{2mm}
    \includegraphics[scale=0.43]{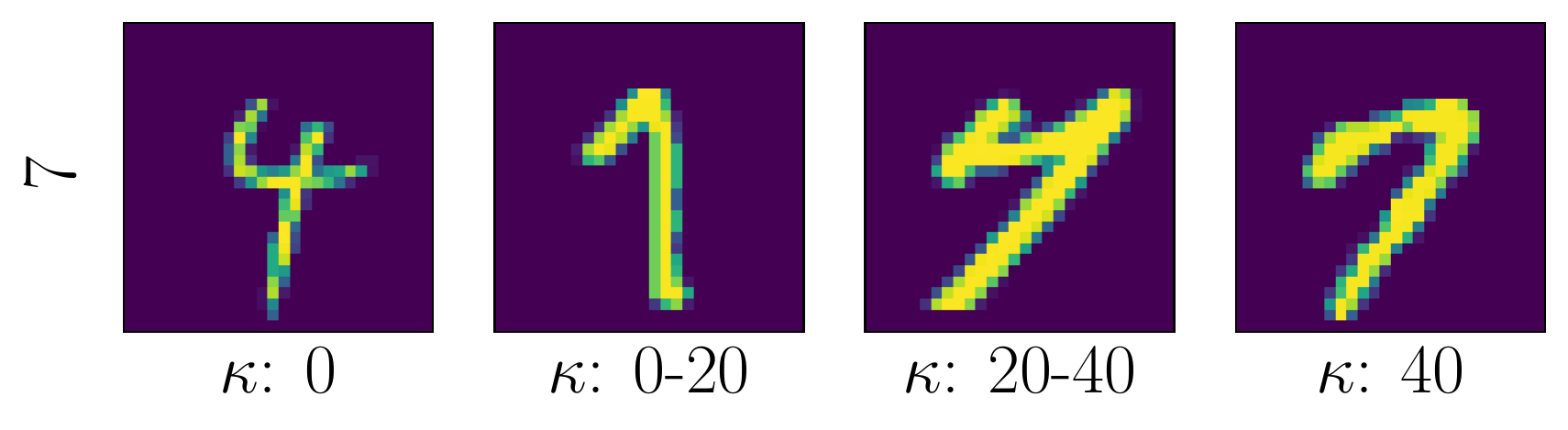}\\
    \includegraphics[scale=0.43]{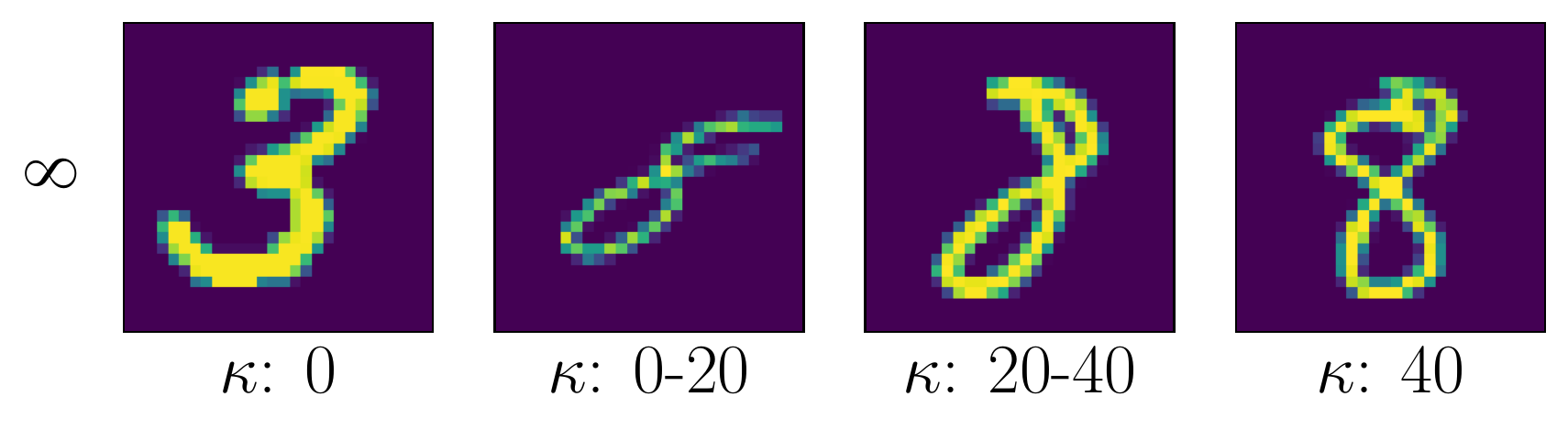}
    \hspace{2mm}
    \includegraphics[scale=0.43]{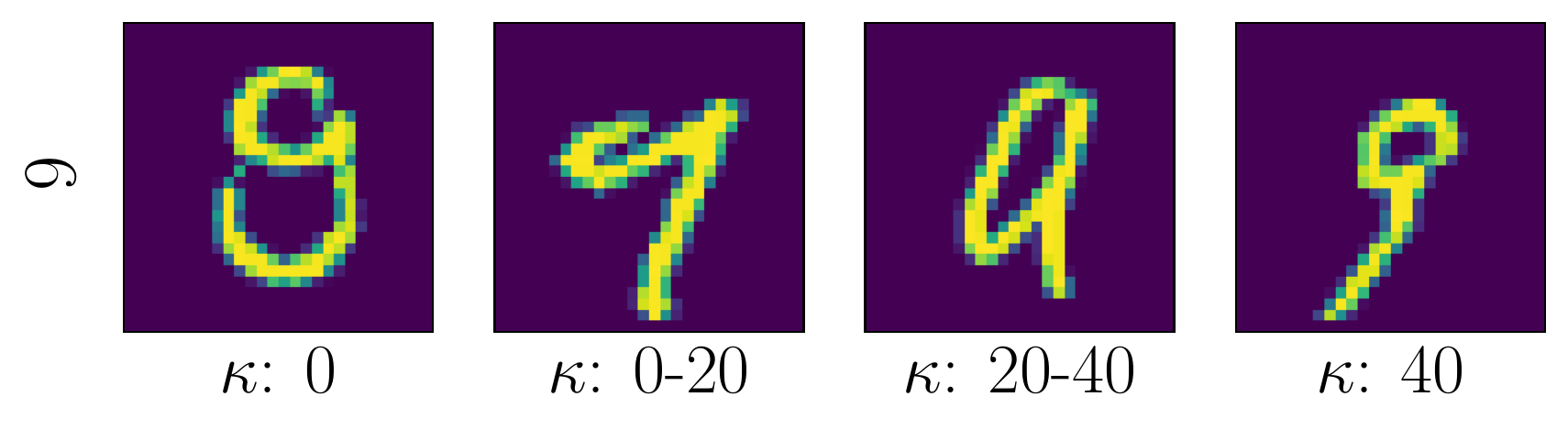}\\
    %\vspace{-3mm}
    \caption{The geometry value $\kappa$ w.r.t. images in \textit{MNIST} with $10\%$ \textit{symmetric-flipping} noise. The leftmost is the given label of all images on the right. We randomly selected four examples with the different $\kappa$ ($\kappa=0$, $\kappa \in (0,20)$, $\kappa \in (20,40)$,  $\kappa=40$) in each class. As the geometric value $\kappa$ increases from left ($\kappa = 0$) to right ($\kappa = 40$), the semantic information of images is more typical and recognizable.}
    %\vspace{-5mm}
    \label{fig:appendix_mnist_rare_pgd_steps}
\end{figure}

\paragraph{Result.} In Figures~\ref{fig:appendix_cifar10_rare_pgd_steps} and~\ref{fig:appendix_mnist_rare_pgd_steps}, we visualize more results about the semantic information of images corresponding to different $\kappa$ using \textit{CIFAR-10} and \textit{MNIST} datasets. For obtaining the geometry value $\kappa$, we use the GA-PGD method proposed by~\citet{zhang2020geometryaware}, which calculates the least number of iterations that PGD needs to find the mis-classified adversarial variants of input data. On \textit{CIFAR-10}, it is calculated by PGD-10 attack with the perturbation bound $\epsilon=0.031$ and the step size $\alpha=0.007$. On \textit{MNIST}, the geometry value $\kappa$ is calculated by PGD-40 attack with the perturbation bound $\epsilon=0.3$ and the step size $\alpha=0.01$. In general, the geometry value $\kappa$ can represent whether the data is relatively typical or rare.

\clearpage

\section{Applications of Geometry Value $\kappa$}
\label{sec:appendix_app_kappa}

In this section, we provide the detailed experimental setups for robust annotator and confidence scores. First, we provide the 
detailed version of Algorithm~\ref{alg:RA} (i.e., Algorithm~\ref{alg:detailed_RA}) and the details to implement the experiment in Figure~\ref{fig:annotator_effect} (Appendix~\ref{sec:appendix_app_robust_annotator}). Second, we provide the details to implement the experiment in Figure~\ref{fig:resnet18_confidence} (Appendix~\ref{sec:appendix_app_confidence_scores}).

\subsection{Robust Annotator}
\label{sec:appendix_app_robust_annotator}

In Figure~\ref{fig:annotator_effect}, we compare four methods on the \textit{CIFAR-10} dataset, namely, our robust annotator with $20\%$ symmetric-flipping noise, the PGD-based annotator with $20\%$ symmetric-flipping noise, the PGD-based annotator without noise, and the standard annotator without noise. 

\vspace{-1mm}
\begin{algorithm}[h]
   \caption{Robust Annotator Algorithm (in detail).}
   \label{alg:detailed_RA}
   \SetKwInOut{Input}{Input}\SetKwInOut{Output}{Output}

  \Input{network $f_{\mathbf{\theta}}$, training dataset $S = \{(\bx_i, y_i) \}^{n}_{i=1}$, learning rate $\eta$, number of epochs $T$, batch size $m$, number of batches $M$, threshold for geometry value $K$, threshold for loss value $L$.}
  
  \Output{robust annotator $f_{\mathbf{\theta}}$.}
  \For{\rm{epoch} $= 1$, $\dots$, $T$}{
    
    \For{\rm{mini-batch} $=1$, $\dots$, $M$}{
    \textbf{Sample:} a mini-batch $\{(\bx_i, y_i) \}^{m}_{i=1}$ from $S$.
    
    \For{i = 1,\ldots,m (\rm{in parallel})}{
     \textbf{Calculate:} $\kappa_i$ and $\ell_i$ of ($x_i$,$y_i$) by GA-PGD method~\citep{zhang2020geometryaware} and $\ell(f_{\mathbf{\theta}}(\bx_i), y_i)$, respectively.
    
    \If{$\kappa_i < K$ and $\ell_i > L$}{
        \textbf{Update:} $y_i \gets \arg\max_{i} f_{\theta} ({ {x} })$.
    }
    \textbf{Generate:} adversarial data $\bxtidle_i$ by PGD method~\citep{Madry_adversarial_training}.
    
    }
    
    \textbf{Update:} $\mathbf{\theta} \gets \mathbf{\theta} - \eta \nabla_{\mathbf{\theta}}  \{ \ell(f_{\mathbf{\theta}}(\bxtidle_i), y_i)\}.$
    }
  }
%\vspace{-3mm}
\end{algorithm}

\paragraph{Experimental setup.}
To generate the noisy training data, we randomly assign the wrong label for a part of correct training data using the method in~\citet{han2018co}. For the \textit{CIFAR-10} dataset, we normalize it into $[0,1]$: Each pixel is scaled by $1/255$. We perform the standard \textit{CIFAR-10} data augmentation: a random $4$ pixel crop followed by a random horizontal flip. For all annotators, we train WRN-32-10~\citep{zagoruyko2016WRN} for $120$ epochs using SGD with $0.9$ momentum. The initial learning rate is $0.1$ reduced to $0.01$, $0.001$ and $0.0005$ at epoch $60$, $90$ and $110$. The weight decay is $0.0002$. For standard annotator, we use natural data to update the model. For our robust annotator and the PGD-based annotator, we generate the adversarial data to update the model, the perturbation bound $\epsilon_{\mathrm{train}} = 0.031$, the PGD step is fixed to $10$, and the step size is fixed to $0.007$. All PGD generation have a random start, i.e, the uniformly random perturbation of $[-\epsilon,\epsilon]$ added to the natural data before PGD iterations. For our robust annotator, we use the same generation method with PGD-based annotator as previous mentioned before Epoch $40$. After that, we use our Algorithm~\ref{alg:detailed_RA} to train our robust annotator. We set the threshold for geometry value $K=2$ and the threshold for loss value $L$ to the loss value of $20\% \cdot m$ largest natural data in each mini-batch, where the $m=128$ is batch size. We use the model predictions of the selected natural data as their new label to generate the adversarial data. As for the evaluations, we select a part of natural test data on the test set of \textit{CIFAR-10} to add adversarial manipulations by PGD-20 attack. The perturbation bound $\epsilon_{test} = 0.031$, the step number is $20$, and the step size $\alpha = \epsilon_{test}/4$, which keeps the same as \cite{Wang_Xingjun_MA_FOSC_DAT}. We use the natural and adversarial test data to check the performance of annotators on assigning correct labels for the U data.

\subsection{Confidence Scores}
\label{sec:appendix_app_confidence_scores}

In Figure~\ref{fig:resnet18_confidence}, we plot the accuracy and number of correctly predicted U data w.r.t the geometry value $\kappa$. 

\paragraph{Experimental setup.}
We train ResNet-18 model in AT with $20\%$ symmetric-flipping noise on the \textit{CIFAR-10} dataset. The training settings keep the same as Appendix~\ref{sec:app_train_test}. We use the model checkpoint at Epoch $35$ for assigning labels and we randomly select $2000$ test data in \textit{CIFAR-10} as unlabeled data. We run the test with 5 repeated times with different random seeds for selecting different test data. In the left panel of Figure~\ref{fig:resnet18_confidence}, we calculate the mean and standard deviation value of accuracy. In the right panel of Figure~\ref{fig:resnet18_confidence}, we show the number of correctly/wrongly predicted data in one of the experiments.

%(To be completed)\\
%(To be completed)\\

%Some previous studies tentatively introduced both corrupted data and labels into their focused problems:\citet{teng1999correcting} selectively corrupted data attributions to handle noisy labels in training; \citet{huang2020self} designed a self-adaptive training to robustly learning from data corrupted by random noises and adversarial examples. In this paper, instead of designing new algorithms in adversarial training and label-noise training, our focus is to \textit{explore and understand} the interaction of adversarial training with label noise. Recently, \citet{sanyal2020benign} identified label noise as one of the causes for adversarial vulnerability and showed that adversarial training does not fit noise. Via studying the performance of adversarial training with correct/incorrect data in depth, we display some further findings (Sections~\ref{sec:geo_smoothing} and~\ref{sec:diff_st_at}) and provide a new measurement (Section~\ref{sec:geo_kappa}) and its applications (Section~\ref{sec:app_kappa}).
%\section{Robust annotator}
%\label{appendix:robust_annotator}
%\subsection{Ablation Study}

\clearpage

%%%%%%%%%%%%%%%%%%%%%%%%%%%%%%%%%%%%%%%%%%%%%%%%%%%%%%%%%%%%%%%%%%%%%%%%%%%%%%%
%%%%%%%%%%%%%%%%%%%%%%%%%%%%%%%%%%%%%%%%%%%%%%%%%%%%%%%%%%%%%%%%%%%%%%%%%%%%%%%

\end{document}